\pdfoutput=1
\documentclass[journal]{IEEEtran}

\usepackage{graphicx}
\usepackage{amsmath}
\allowdisplaybreaks[4]
\usepackage{array}
\usepackage{booktabs}
\usepackage{amssymb}
\usepackage{color}
\usepackage{array}

\usepackage{url}

\usepackage[CJKbookmarks=true, bookmarksnumbered=true, bookmarksopen=true,
colorlinks=true, 
pdfborder=001, 
citecolor=blue, linkcolor=blue, anchorcolor=red, urlcolor=black
]{hyperref}

\usepackage[tight,footnotesize]{subfigure}

\begin{document}\sloppy

\title{Sparse Graphical Representation based Discriminant Analysis for Heterogeneous Face Recognition}
%
%
%

\author{Chunlei Peng,
        Xinbo~Gao,~\IEEEmembership{Senior Member,~IEEE,}
        Nannan Wang,~\IEEEmembership{Member,~IEEE,}
        and~Jie~Li
\thanks{C. Peng and X. Gao are with the State Key Laboratory of Integrated Services Networks, School of Electronic Engineering, Xidian University, Xi'an 710071, Shaanxi, P. R. China (e-mail: clp.xidian@gmail.com; xbgao@mail.xidian.edu.cn).

N. Wang is with the State Key Laboratory of Integrated Services Networks, School of Telecommunications Engineering, Xidian University, Xi'an 710071, Shaanxi, P. R. China (e-mail: nnwang@xidian.edu.cn).

J. Li is with the Video and Image Processing System Laboratory, School of Electronic Engineering, Xidian University, Xi'an 710071, Shaanxi, P. R. China (leejie@mail.xidian.edu.cn).

}}
\maketitle

\begin{abstract}
Face images captured in heterogeneous environments, \textit{e.g.}, sketches generated by the artists or composite-generation software, photos taken by common cameras and infrared images captured by corresponding infrared imaging devices, usually subject to large texture (\textit{i.e.}, style) differences. This results in heavily degraded performance of conventional face recognition methods in comparison with the performance on images captured in homogeneous environments. In this paper, we propose a novel sparse graphical representation based discriminant analysis (SGR-DA) approach to address aforementioned face recognition in heterogeneous scenarios. An adaptive sparse graphical representation scheme is designed to represent heterogeneous face images, where a Markov networks model is constructed to generate adaptive sparse vectors. To handle the complex facial structure and further improve the discriminability, a spatial partition-based discriminant analysis framework is presented to refine the adaptive sparse vectors for face matching. We conducted experiments on six commonly used heterogeneous face datasets and experimental results illustrate that our proposed SGR-DA approach achieves superior performance in comparison with state-of-the-art methods.
\end{abstract}


%
\IEEEpeerreviewmaketitle
\section{Introduction}
\label{section I}

Face recognition is an important and challenging problem in computer vision. Despite great progress achieved in recent years, there are still many challenging face recognition scenarios, for example, face recognition problem on images captured in heterogeneous environments. Conventional face recognition methods generally perform poor because of large texture (or style) differences between heterogeneous face images. Matching heterogeneous face images, \textit{i.e.}, heterogeneous face recognition (HFR), is now attracting growing attentions on account of its large theoretical challenges and great potential applications. For example, in the law enforcement agency, when no face image of the suspect is available or there are only poor quality images in video surveillance, face sketches created by forensic artists\footnote{http://www.askaforensicartist.com/composite-sketch-leads-to-arrest-in-virginia-highland-robbery/.}$^,$\footnote{http://www.askaforensicartist.com/phoenix-police-sketch-leads-to-arrest-of-kidnapper/.}, or composite-generation software \cite{Ref1} are commonly used to perform matching with mug shot photos. In complex illumination environment, near infrared images (NIR) \cite{Ref2} or thermal infrared images (TIR) \cite{Ref3} are preferred for authentication by matching with controlled indoor visible light (VIS) face images that have been enrolled before. These scenarios introduce a great challenge to face recognition systems and in this paper we present a sparse graphical representation based discriminant analysis (SGR-DA) approach for aforementioned scenarios.

Up to now, many HFR approaches have been proposed, which can be broadly classified into three categories: image synthesis-based methods, common subspace projection-based methods, and modality invariant feature descriptor-based methods. Image synthesis-based methods \cite{RefRR3,Ref5,Ref6,Ref10,Ref11,Ref12,Ref13,Ref14,Ref15,Ref16} transform face images from one modality into another such that they become homogeneous and the conventional face recognition methods can then be applied directly. However, the image synthesis process is a complex problem itself, even more difficult than the recognition task. Furthermore, these synthesis-based methods are designed for fixed modalities respectively and cannot be well generalized to different HFR scenarios. The common subspace projection-based methods \cite{Ref2,Ref17,Ref18,Ref19,Ref20,Ref21,Ref22,Ref23} usually learn modality-specific mappings to project heterogeneous face images into a common latent space, where they can be matched directly. However, since the projection procedure reduces the discriminability, it degrades the performance of HFR methods. The modality invariant feature descriptor based methods \cite{Ref1,Ref3,Ref7,Ref8,Ref24,Ref27,Ref28,Ref29,Ref30,Ref32,Ref33} first represent face images by extracting modality invariant features which are then measured for matching. Yet most of these methods extract feature descriptors ignoring the facial spatial information and thus these methods have limited discriminability.

Recently, a graphical representation-based approach \cite{Ref4} has been proposed to deal with HFR problem. The spatial information of facial structures is taken into consideration by jointly modeling heterogeneous face patches through Markov networks. A new similarity metric is developed for matching and the method is effective and efficient on multiple HFR scenarios. Different from the feature representations directly learnt from raw pixels \cite{RefR1,RefR2}, the graphical representations are generated through state-of-the-art face synthesis model (Markov networks). The basic assumption is that the heterogeneous faces of the same person tend to have similar weight matrixes during the synthesis process. Based on this assumption, a representation dataset containing some heterogeneous face pairs is constructed to encode the faces. The graphical representations are generated with spatial information taken into consideration. Furthermore, the graphical representation-based approach can be easily and effectively generalized to multiple HFR scenarios. However, the Markov networks employed in \cite{Ref4} suffer from the same shortcoming with those used in synthesis scenarios \cite{Ref5,Ref6}: fixed $K$ nearest neighbors of the probe image patch are selected when constructing Markov networks. The performance of these methods is heavily affected by the number of nearest neighbors $K$, which is manually determined. In addition, in the method \cite{Ref4}, the whole face image is matched through a single classifier without considering the complex facial structure. Although there are several methods \cite{Ref7,Ref8,Ref9,RefTC2} that divide face images into local regions and perform discriminant analysis on each region respectively, it remains an unresolved problem to improve the discriminability to the complex facial structure.

In this paper, we propose a novel sparse graphical representation based discriminant analysis (SGR-DA) method for heterogeneous face recognition. Firstly, a new Markov networks model is deployed to generate an adaptive sparse graphical representation. Unlike selecting $K$ nearest neighbors as employed in \cite{Ref5,Ref6,Ref4}, the proposed method skips the $K$ nearest neighbor searching process and all related image patches are considered when the Markov networks model is constructed. The non-negative sparse regularization in the Markov networks model results in adaptive sparse vectors. Secondly, a spatial partition-based discriminant analysis framework is proposed to handle the complex facial structure and improve the discriminability. Three spatial partition strategies are developed and discriminant analysis is performed separately on each spatial partition region. The proposed discriminant analysis framework is simple yet effective and it results in high recognition accuracy. Experimental results on six commonly used heterogeneous face datasets demonstrate the effectiveness of the proposed method.

The main contributions of this paper are summarized as follows.
\begin{enumerate}
    \item We propose an adaptive sparse graphical representation scheme to represent heterogeneous face images. By skipping the $K$ nearest neighbor selection process, adaptive sparse vectors can be generated from the Markov networks model;
    \item We develop a spatial partition-based discriminant analysis framework for heterogeneous face matching. With the proposed spatial partition strategies, the discriminability of heterogeneous face images is improved.
\end{enumerate}

The rest of the paper is organized as follows. In section \ref{section II} we briefly review existing HFR methods. Section \ref{section III} details the proposed SGR-DA approach. Experimental results are presented in section \ref{section IV}, and section \ref{section V} concludes the paper.

\section{Related Work}
\label{section II}

In this section, we give a brief review on representative HFR methods of aforementioned three categories: image synthesis-based methods, common subspace projection based methods, and modality invariant feature descriptor-based methods.

Image synthesis-based methods transform heterogeneous face images into the same modality. \cite{Ref10} first proposed an eigen-transformation algorithm for face sketch-photo synthesis. Considering the drawback of performing synthesis on the whole face \cite{Ref10}, \cite{Ref11} and \cite{Ref12} employed local linear embedding (LLE) to perform face sketch-photo synthesis and NIR-VIS synthesis respectively. In order to take the relationship between a face image patch and its neighboring patches into consideration, the Markov random field (MRF) model was introduced by \cite{Ref13} (for face sketch-photo synthesis) and \cite{Ref14} (for TIR-VIS synthesis). In the aforementioned MRF model-based methods only the ``best'' candidate patch for representing the probe patch was selected, and it would cause facial deformations. Thus, \cite{Ref5} proposed a Markov weight field (MWF) model, by selecting a number of candidates to construct the Markov networks model, which is capable of synthesizing new patches without existing in the training set. \cite{Ref6} further incorporated the test image into the learning process through a transductive face sketch-photo synthesis (TFSPS) framework. A multiple representations based approach was proposed by Peng \emph{et al.} \cite{RefTNN1}. Recently \cite{Ref15} proposed a real-time face sketch synthesis method by considering the synthesis procedure as a denoising issue. Inspired by the wide applications of convolutional neural network (CNN) in computer vision, \cite{Ref16} developed a CNN-based sketch-photo synthesis method by taking the whole face photo as inputs and generating the corresponding whole face sketch.

Common subspace projection-based methods attempt to project heterogeneous face images into a latent subspace where the heterogeneity is minimized. It began with \cite{Ref2} through a common discriminant feature extraction (CDFE) approach. \cite{Ref17} proposed to use the correlational regression method (canonical correlation analysis) to map NIR and VIS images into a common feature space. \cite{Ref18} proposed a coupled spectral regression (CSR) based method for NIR-VIS matching, which was later improved by learning mappings from both modalities \cite{Ref19}. The partial least squares (PLS) algorithm was exploited by \cite{Ref20} to learn the linear mapping transformations between face images in different modalities. \cite{Ref21} took both the positive and negative constraints during metric learning process into consideration, and proposed a cross modal metric learning (CMML) method for heterogeneous face matching. A multi-view discriminant analysis (MvDA) method was proposed by \cite{Ref22}, which exploited both inter-view and intra-view correlations of heterogeneous face images. Inspired by the unsupervised deep learning algorithms, \cite{Ref23} utilized Restricted Boltzmann Machines to learn a shared representation for HFR.

The modality invariant feature descriptor-based methods encode face images with local feature descriptors, which can then be utilized for recognition. \cite{Ref24} first proposed to use a difference of Gaussian (DoG) filter and multiblock local binary patterns (MB-LBP) for matching NIR and VIS. Later \cite{Ref7} explored the scale invariant feature transform feature (SIFT) \cite{Ref25} and multiscale local binary pattern feature (MLBP) \cite{Ref26} and proposed a local feature-based discriminant analysis (LFDA) framework for forensic sketch recognition with a populated gallery. \cite{Ref8} designed a learning-based feature by coupled information-theoretic encoding (CITE) for matching viewed sketches with photos. Two other binary pattern features, \textit{i.e.}, local radon binary pattern (LRBP) \cite{Ref27} and local difference of Gaussian binary pattern (LDoGBP) \cite{Ref28}, were also designed for viewed sketch recognition. In order to mimic the gap between viewed sketch recognition and forensic sketch recognition, \cite{Ref29} proposed a semi-forensic sketch dataset and deployed the multi-scale circular Weber's local descriptor (MCWLD) for matching. \cite{Ref3} utilized nonlinear kernel similarities to represent face image and evaluated their prototype random subspace (P-RS) approach on four HFR scenarios. Recently, a number of composite sketch recognition methods \cite{Ref1,Ref30,Ref32,Ref33} were proposed. Mittal \textit{et al.} \cite{Ref33} presented a transfer learning-based deep learning representation for composite sketch recognition. Considering the insufficient usage of facial spatial information, a graphical representation based HFR approach was proposed recently \cite{Ref4}. The graphical representation was extracted by Markov networks and a coupled representation similarity metric was designed to cater for the obtained representations. However, the graphical representation in \cite{Ref4} suffered the same shortcomings with existing methods \cite{Ref5,Ref6} that the performance was seriously affected by the $K$ nearest neighbor searching process with manually defined $K$. In this paper, we skip this process and propose an adaptive sparse graphical representation. Additionally, we propose a new spatial partition-based discriminant analysis framework to improve the discriminability and test our method on six different heterogeneous face datasets.

\section{Sparse Graphical Representation based Discriminant Analysis for Heterogeneous Face Recognition}
\label{section III}

In this section, we present the proposed sparse graphical representation based discriminant analysis method for heterogeneous face recognition. We first give the formulation and analysis of the adaptive sparse graphical representation. Then, we introduce the spatial partition-based discriminant analysis framework. Finally, the whole SGR-DA approach is developed. Without loss of generality and for ease of presentation, we take viewed sketch recognition as an example here to introduce the proposed method. It can be seen from the experimental section that the proposed approach can be generalized to other HFR scenarios.

\subsection{Adaptive Sparse Graphical Representation}
\label{subsection III-A}

\begin{figure*}[t]
\begin{center}
\hspace{-10mm}
\subfigure{
\begin{minipage}[b]{0.18\linewidth}
    \centering
   \includegraphics[width=1.15\linewidth]{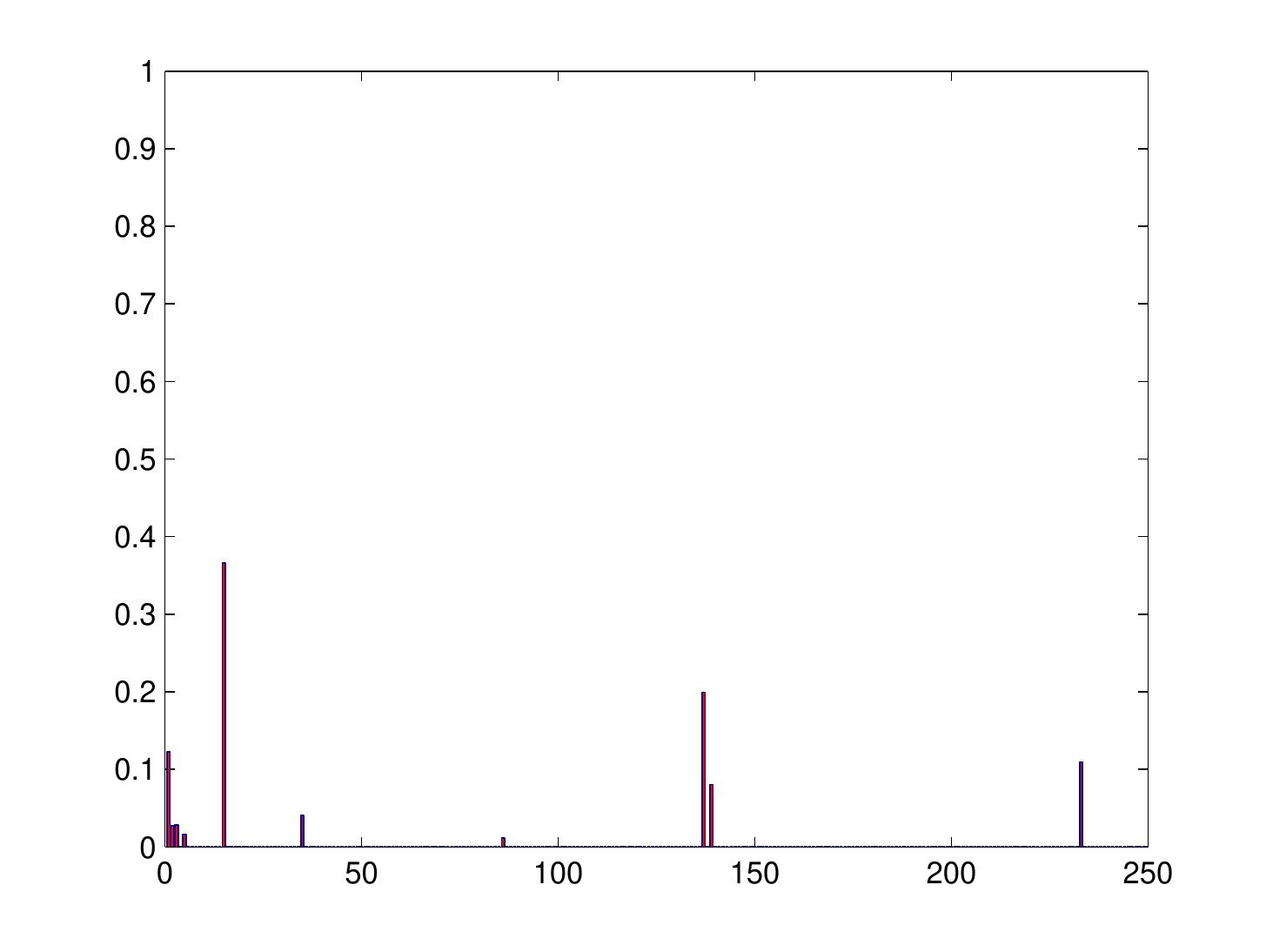}
\end{minipage}}
\subfigure{
\begin{minipage}[b]{0.18\linewidth}
    \centering
   \includegraphics[width=1.15\linewidth]{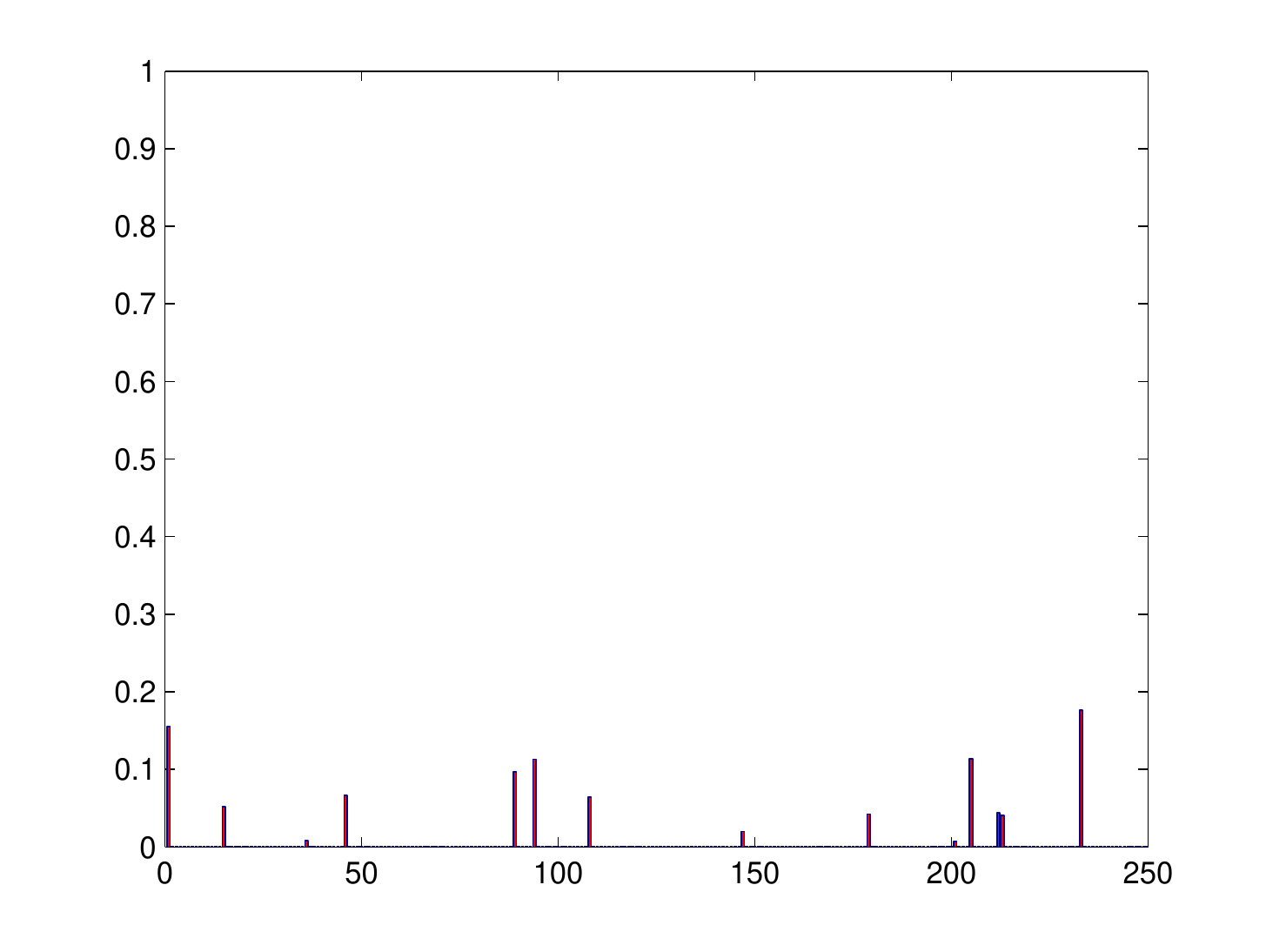}
\end{minipage}}
\subfigure{
\begin{minipage}[b]{0.18\linewidth}
    \centering
   \includegraphics[width=1.15\linewidth]{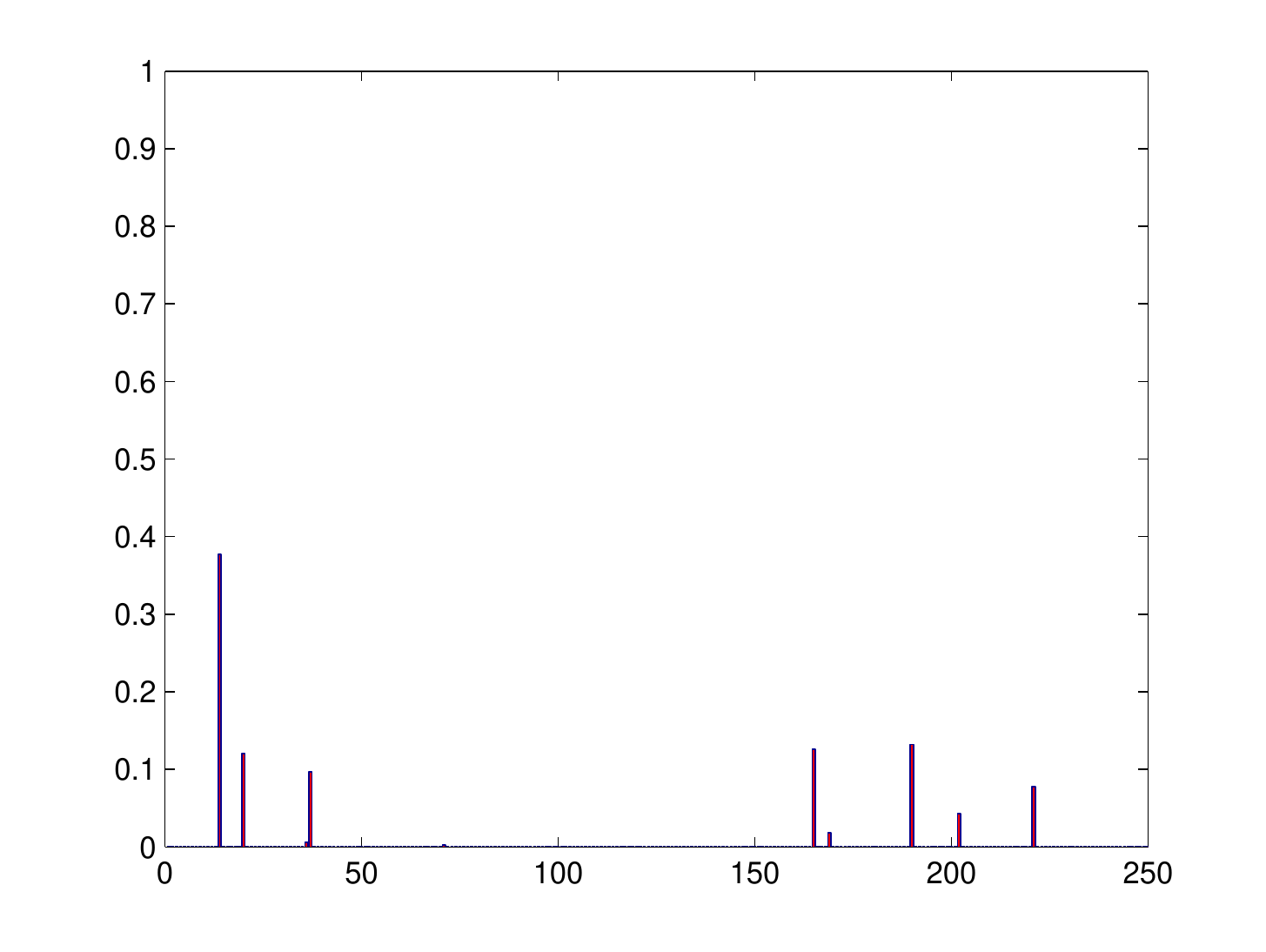}
\end{minipage}}
\subfigure{
\begin{minipage}[b]{0.18\linewidth}
    \centering
   \includegraphics[width=1.15\linewidth]{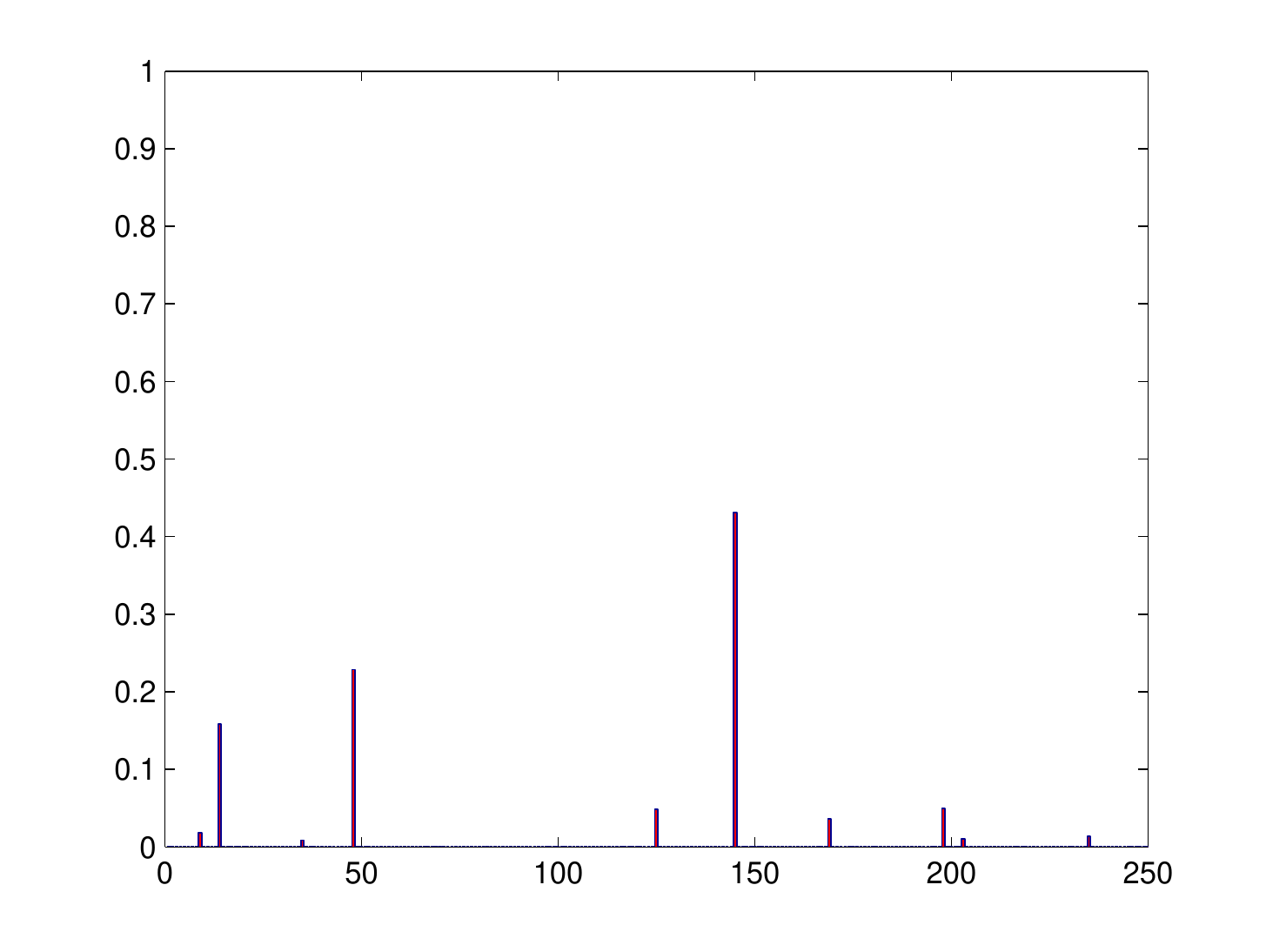}
\end{minipage}}
\subfigure{
\begin{minipage}[b]{0.18\linewidth}
    \centering
   \includegraphics[width=1.15\linewidth]{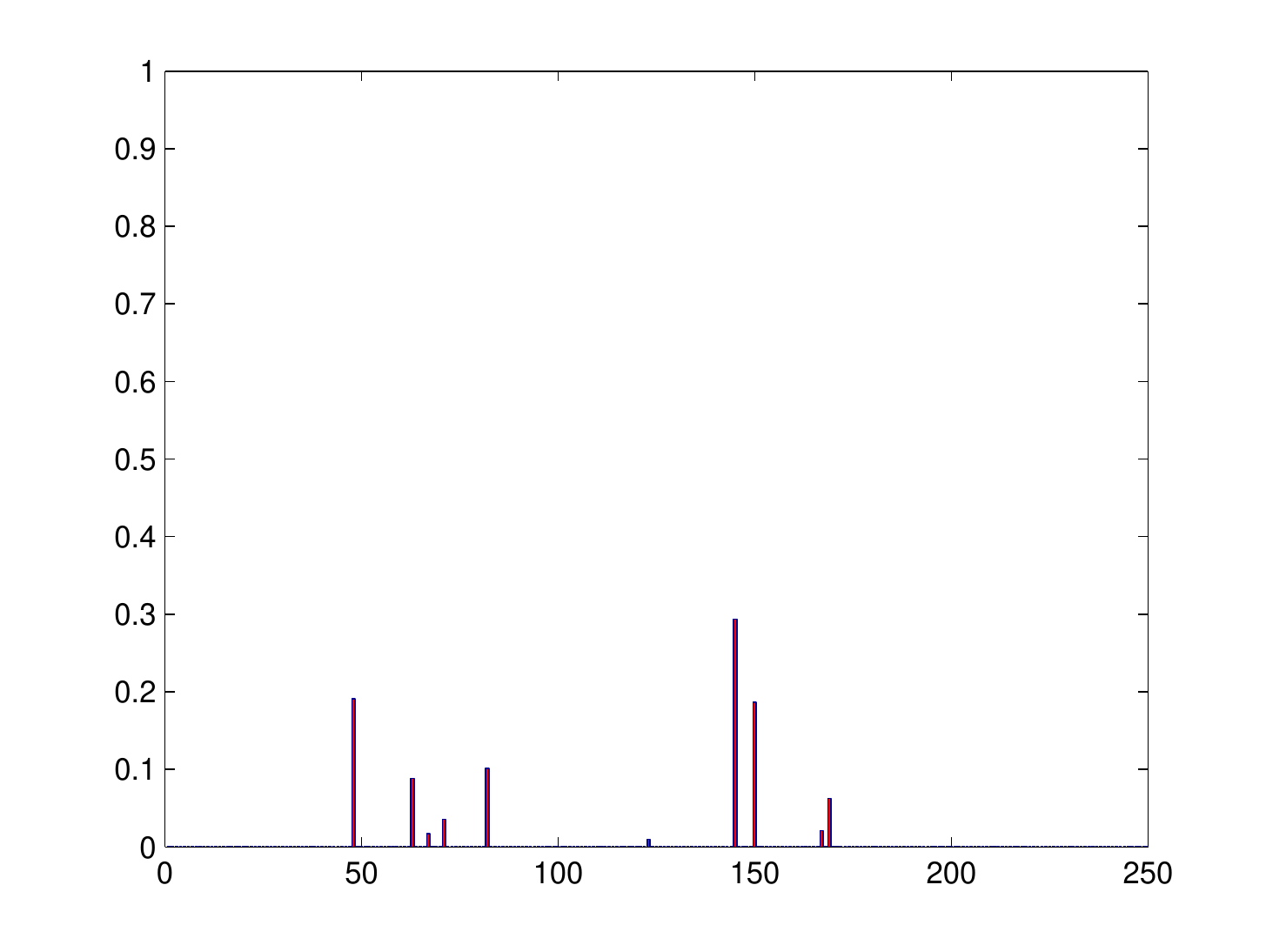}
\end{minipage}}\\
\hspace{-10mm}
\subfigure{
\begin{minipage}[b]{0.18\linewidth}
    \centering
   \includegraphics[width=1.15\linewidth]{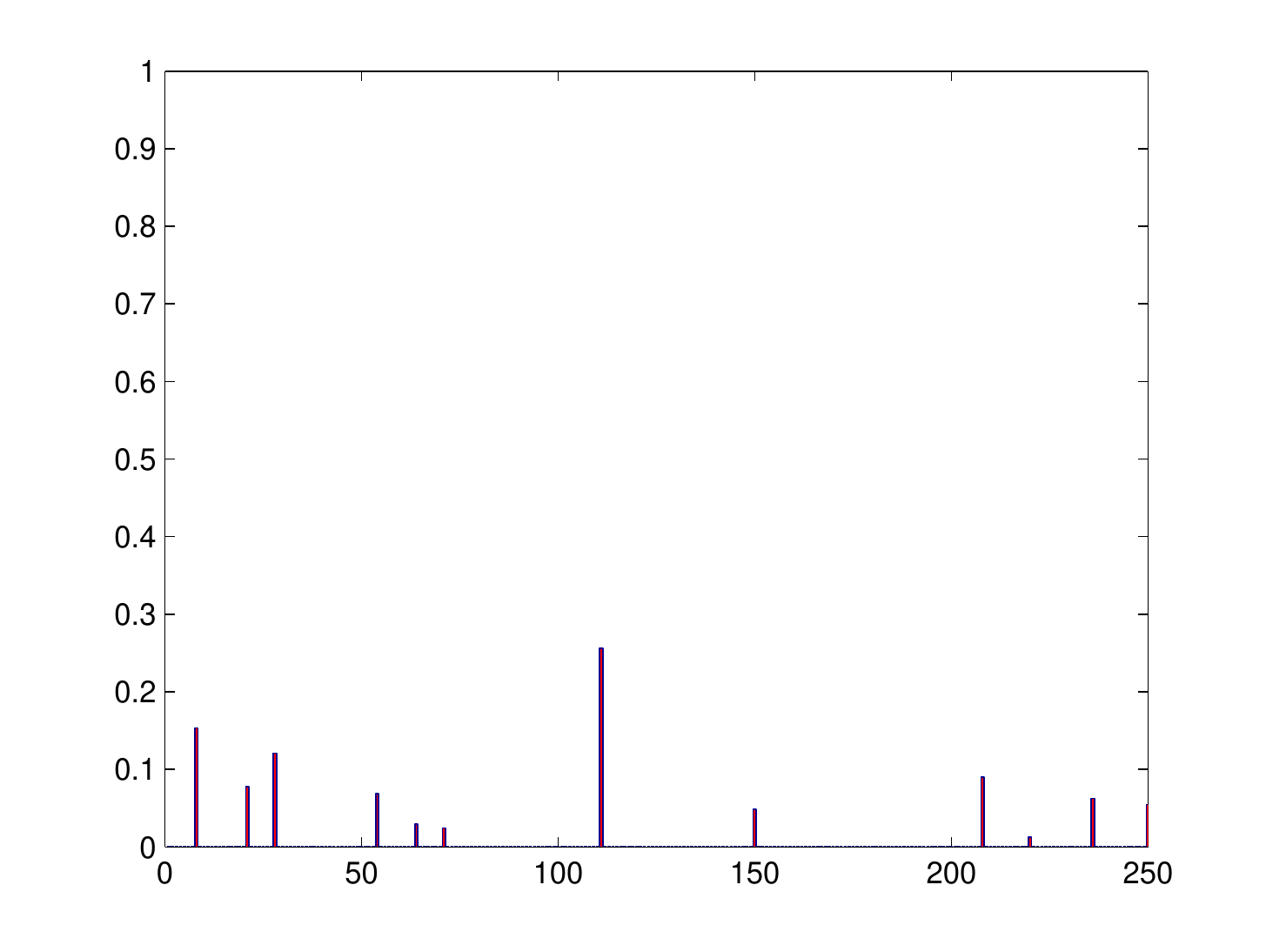}
\end{minipage}}
\subfigure{
\begin{minipage}[b]{0.18\linewidth}
    \centering
   \includegraphics[width=1.15\linewidth]{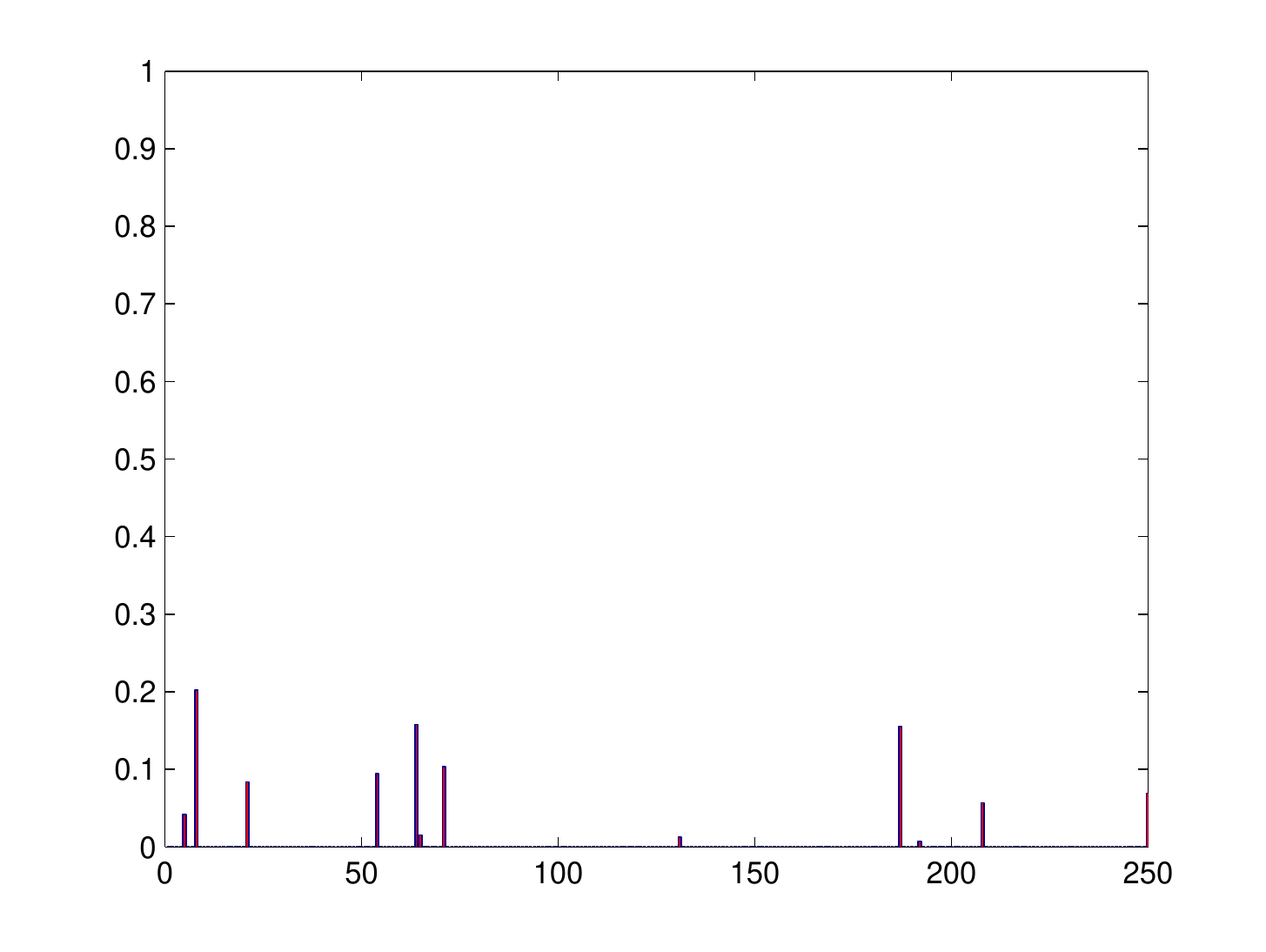}
\end{minipage}}
\subfigure{
\begin{minipage}[b]{0.18\linewidth}
    \centering
   \includegraphics[width=1.15\linewidth]{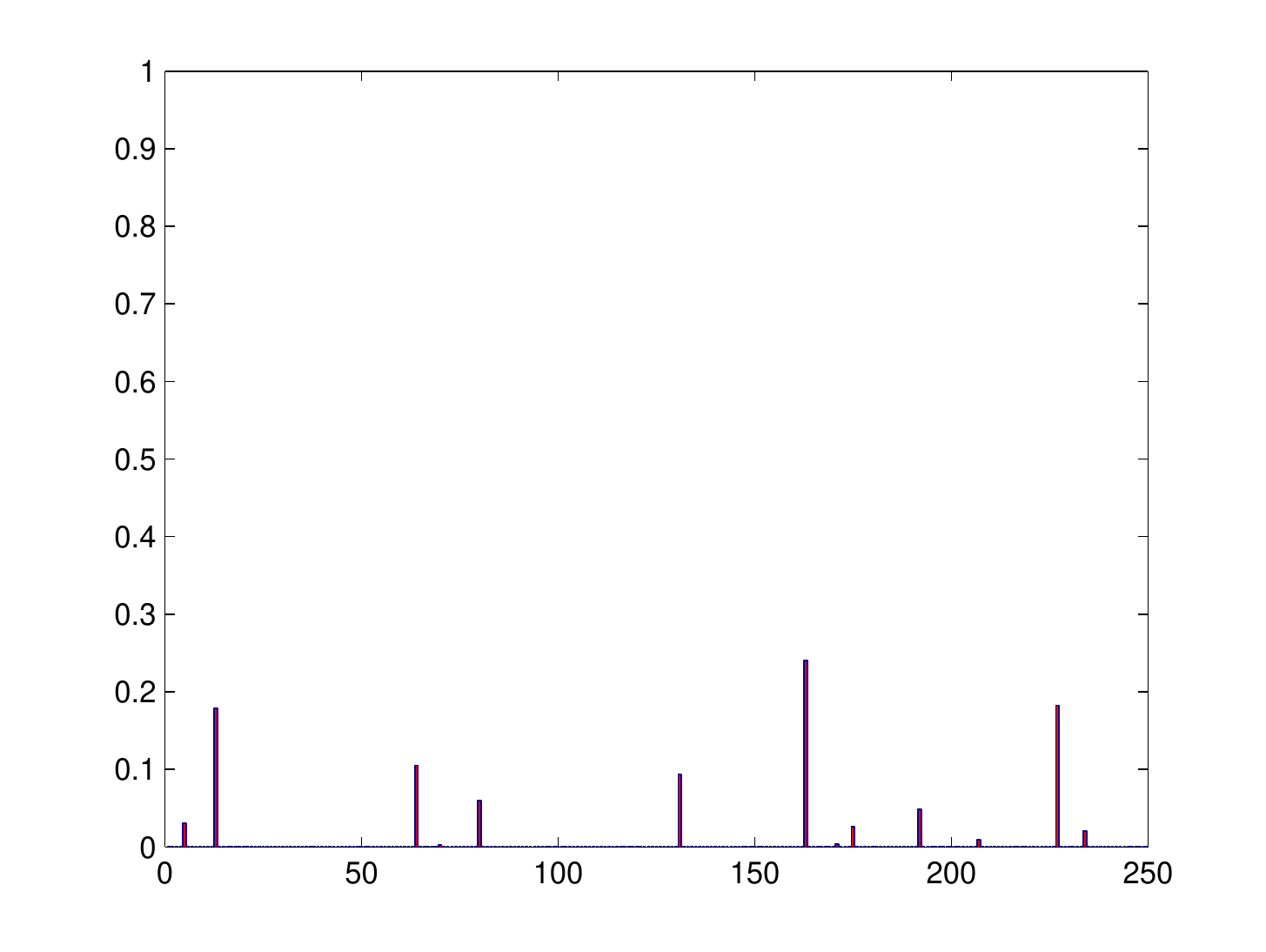}
\end{minipage}}
\subfigure{
\begin{minipage}[b]{0.18\linewidth}
    \centering
   \includegraphics[width=1.15\linewidth]{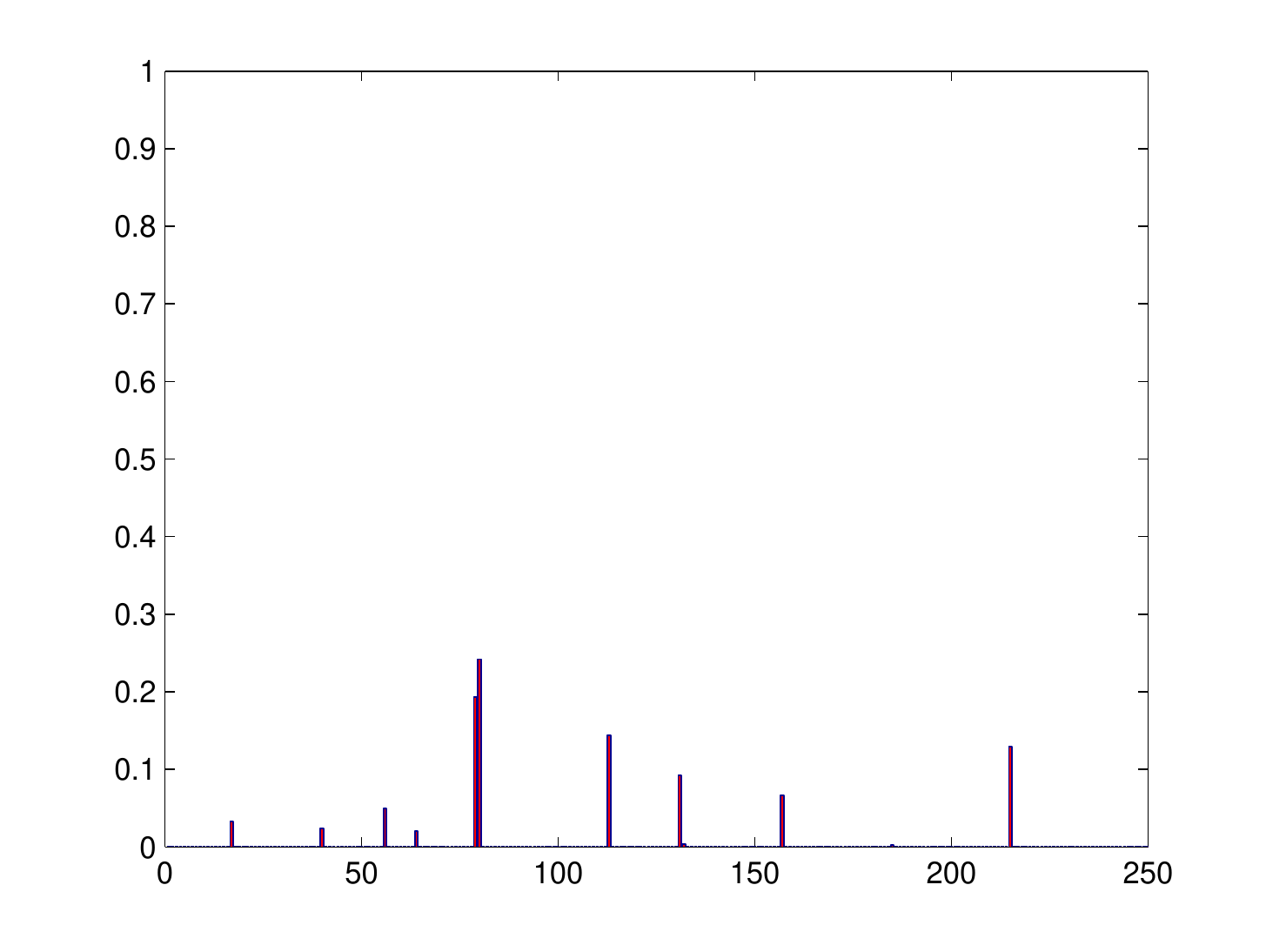}
\end{minipage}}
\subfigure{
\begin{minipage}[b]{0.18\linewidth}
    \centering
   \includegraphics[width=1.15\linewidth]{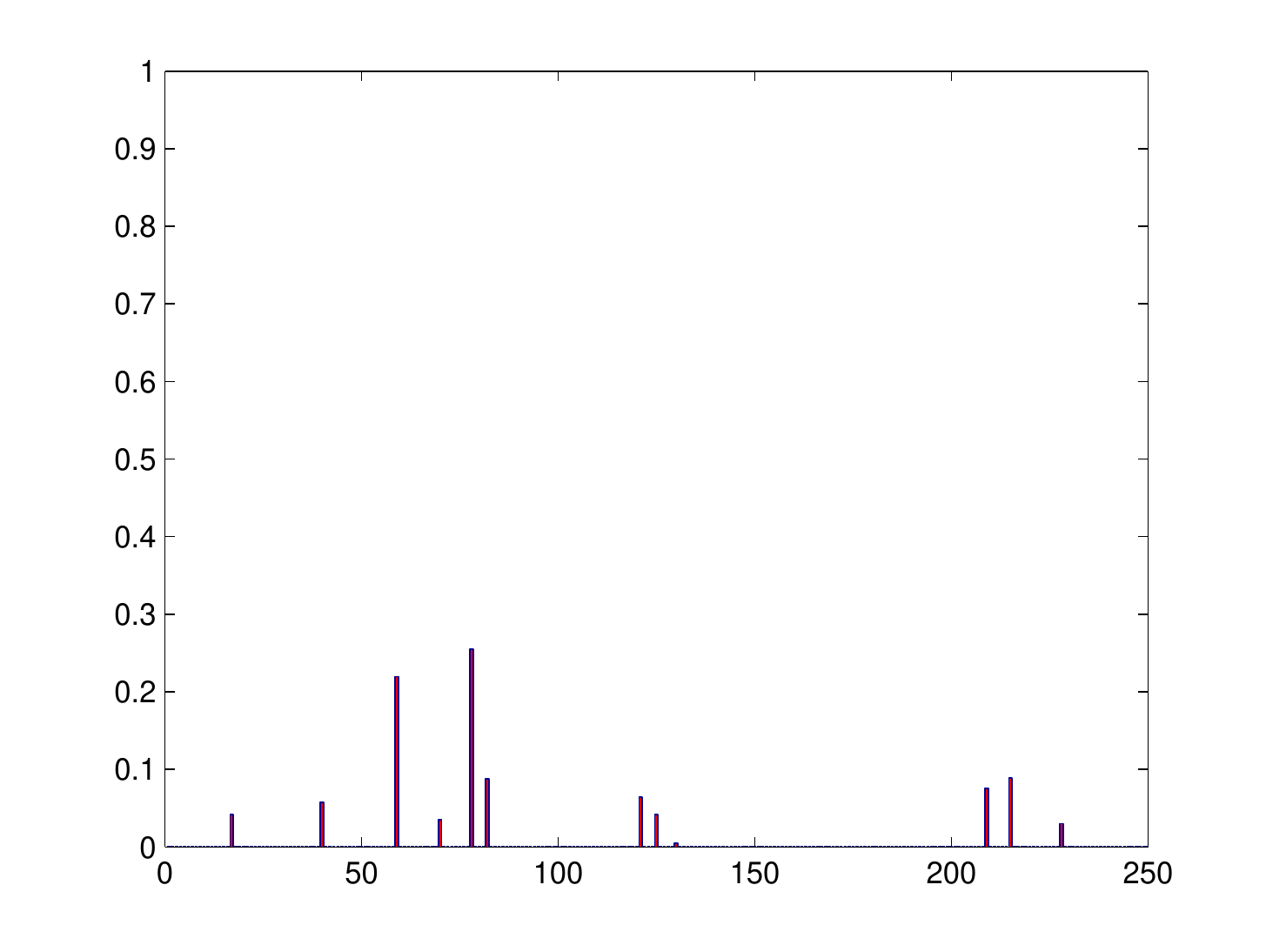}
\end{minipage}}
\end{center}
   \caption{Examples of several adaptive sparse graphical representations obtained on the CUHK face sketch FERET database. 250 face sketch-photo pairs is incorporated into the representation dataset and SIFT feature is used. Each subfigure shows an adaptive sparse vector, with horizontal axis and vertical axis representing the length of the vector and values of each element respectively.}
\label{Figure1}
\end{figure*}
A representation dataset, consisting of $M$ face sketch-photo pairs, is constructed firstly. We divide each face image into $N$ overlapping patches and represent each patch by a feature descriptor (SIFT, for example). Given a probe sketch $\mathbf{t}$ and a gallery photo $\mathbf{g}$, we divide them into patches and represent each patch by the feature descriptor in exactly the same way we have done before for the representation dataset. Let $\mathbf{y}_i$ ($i=$ 1$,$2$,\cdots,N$) denote a probe sketch patch, and $\mathbf{f}(\mathbf{y}_i)$ be the feature descriptor corresponding to $\mathbf{y}_i$. The nearest sketch patch on each face sketch in the representation dataset within the search region $R\times$$R$ around the location of $\mathbf{y}_i$ is selected based on the Euclidean distance of the feature descriptors. Therefore, we can find $M$ related sketch patches for a probe sketch $\mathbf{y}_i$. Likewise, $M$ related photo patches for a gallery photo $\mathbf{x}_i$ can be found.

In existing methods \cite{Ref5,Ref6,Ref4}, $K$ nearest related neighbors are selected from these related patches and the probe sketch patch $\mathbf{y}_i$ can be regarded as a linear combination of the $K$ nearest related neighbors weighted by a column vector $\mathbf{w}_{\mathbf{y}_i}=(w_{\mathbf{y}_{i,1}},\cdots,w_{\mathbf{y}_{i,K}})^T$. A Markov networks model can then be built by jointly modeling all probe sketch patches and their nearest neighbors:
\begin{equation}
\label{Eq:eq 1}
\begin{aligned}
&p(\mathbf{w}_{\mathbf{y}_1},\cdots,\mathbf{w}_{\mathbf{y}_N},\mathbf{y}_1,\cdots,\mathbf{y}_N)\\
=\quad&\prod_i\Phi(\mathbf{f}(\mathbf{y}_i),\mathbf{f}(\mathbf{w}_{\mathbf{y}_i}))\prod_{(i,j)\in\Xi}\Psi(\mathbf{w}_{\mathbf{y}_i},\mathbf{w}_{\mathbf{y}_j})\\
\end{aligned}
\end{equation}
where $(i,j)\in\Xi$ denotes the $i$th probe sketch patch and the $j$th probe sketch patch are adjacent. $\mathbf{f}(\mathbf{w}_{\mathbf{y}_i})$ denotes the linear combination of feature descriptors on the $K$ nearest related neighbors, \textit{i.e.}, $\mathbf{f}(\mathbf{w}_{\mathbf{y}_i})=\sum_{k=1}^K{w_{\mathbf{y}_{i,k}}\mathbf{f}(\mathbf{y}_{i,k})}$. $\Phi(\mathbf{f}(\mathbf{y}_i),\mathbf{f}(\mathbf{w}_{\mathbf{y}_i}))$ and $\Psi(\mathbf{w}_{\mathbf{y}_i},\mathbf{w}_{\mathbf{y}_j})$ are the local evidence function and the neighboring compatibility function respectively. Maximizing the problem in (\ref{Eq:eq 1}) can be formulated as the following problem (\ref{Eq:eq 2}). The detailed proof can be found in the Appendix.
\begin{equation}
\label{Eq:eq 2}
\begin{aligned}
\min_\mathbf{w}\quad&\mathbf{w}^T\mathbf{Q}\mathbf{w}+\mathbf{w}^T\mathbf{c}\\
s.t.\quad&\sum_{k=1}^K{w_{\mathbf{y}_{i,k}}}=1,\ 0\leq{w_{\mathbf{y}_{i,k}}}\leq1,\\
&i=1,2,\cdots,N,\ k=1,2,\cdots,K\\
\end{aligned}
\end{equation}
where $\mathbf{w}$ is the concatenation of $\{\mathbf{w}_{\mathbf{y}_1},\cdots,\mathbf{w}_{\mathbf{y}_N}\}$. The matrix $\mathbf{Q}$ and $\mathbf{c}$ are quadratic parameters which are also explained in the Appendix. The problem (\ref{Eq:eq 2}) can be solved by the cascade decomposition method \cite{Ref5}.

The shortcoming of the above procedure is that the parameter $K$ (\textit{i.e.}, the number of nearest related neighbors) is always defined manually. For example, $K$ was set to 10 in \cite{Ref5,Ref6} and 15-40 in \cite{Ref4}. However, the performance of these methods is heavily affected by $K$. Therefore, in this paper we propose to skip the $K$ nearest neighbor searching process and all the $M$ related image patches are considered. Now, the problem (\ref{Eq:eq 2}) becomes the following optimization issue:
\begin{equation}
\label{Eq:eq 3}
\begin{aligned}
\min_\mathbf{w}\quad&\mathbf{w}^T\mathbf{Q}\mathbf{w}+\mathbf{w}^T\mathbf{c}\\
s.t.\quad&\sum_{m=1}^M{w_{\mathbf{y}_{i,m}}}=1,\ 0\leq{w_{\mathbf{y}_{i,m}}}\leq1,\\
&i=1,2,\cdots,N,\ m=1,2,\cdots,M\\
\end{aligned}
\end{equation}
The constraint in function (\ref{Eq:eq 3}), \textit{i.e.}, $\sum_{m=1}^M{w_{\mathbf{y}_{i,m}}}=1$, is identical to the following constraint when $0\leq{w_{\mathbf{y}_{i,m}}}\leq1$:
\begin{equation}
\label{Eq:eq 4}
\begin{aligned}
\|\mathbf{w}_{\mathbf{y}_i}\|_1=1,\ 0\leq{w_{\mathbf{y}_{i,m}}}\leq1\\
\end{aligned}
\end{equation}
which is a non-negative sparse regularization. The non-negative constraint here prevents subtraction from occurring in the linear combination of the $M$ related image patches, which is contrary to the intuitive notion of combining parts to form a whole \cite{Ref34}. It has been shown that the non-negativity property is advantageous \cite{Ref34}. Different from existing sparse graphical representation applied in several computer vision applications \cite{RefTC1}, the proposed adaptive sparse graphical representation is generated based on state-of-the-art face synthesis model (Markov networks) which can take spatial information into consideration.

The non-negative sparse regularization in our Markov networks model produces an adaptive sparse representation of the data. In our experiments, statistics show that above 90\% elements of the adaptive sparse graphical representation are near zero ($<$10$^{-6}$). Examples of several representation vectors are shown in Fig. \ref{Figure1}. It should be noticed that the size of the representation dataset $M$ is far larger than $K$. Instead of manually selecting $K$ related neighbors for each probe sketch patch at the beginning, the proposed method can adaptively utilize different numbers of related neighbors for different probe sketch patches. This adaptive sparse property makes the face images of different identities to have maximum discriminability. We will validate the effectiveness of it in the experiment section. The obtained adaptive sparse vectors are regarded as the adaptive sparse graphical representation of the probe sketch $\mathbf{t}$, \textit{i.e.}, $\mathbf{W}_{\mathbf{t}}=[\mathbf{w}_{\mathbf{y}_1},\cdots,\mathbf{w}_{\mathbf{y}_N}]$. The adaptive sparse graphical representation of the gallery photo $\mathbf{g}$ can be obtained in a similar way: $\mathbf{W}_{\mathbf{g}}=[\mathbf{w}_{\mathbf{x}_1},\cdots,\mathbf{w}_{\mathbf{x}_N}]$.

\begin{figure*}[t]
\begin{center}
\vspace{-1mm}
\hspace{-15mm}
\subfigure{
\begin{minipage}[b]{0.18\linewidth}
    \centering
   \includegraphics[width=1.4\linewidth]{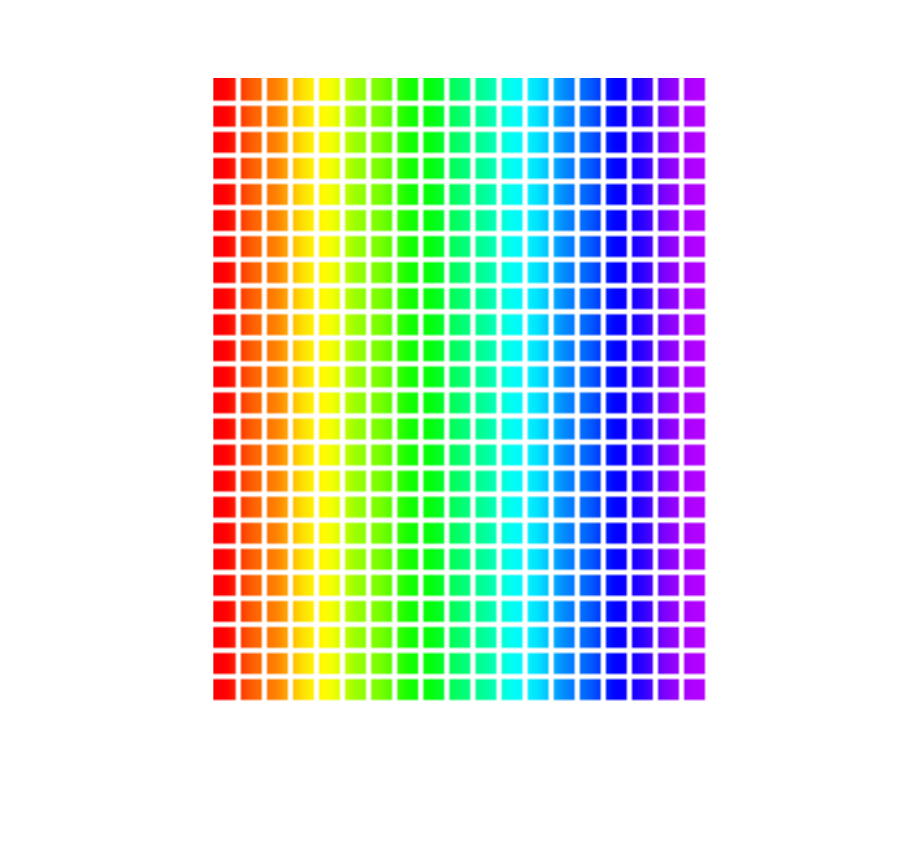}
\end{minipage}}
\subfigure{
\begin{minipage}[b]{0.18\linewidth}
    \centering
   \includegraphics[width=1.4\linewidth]{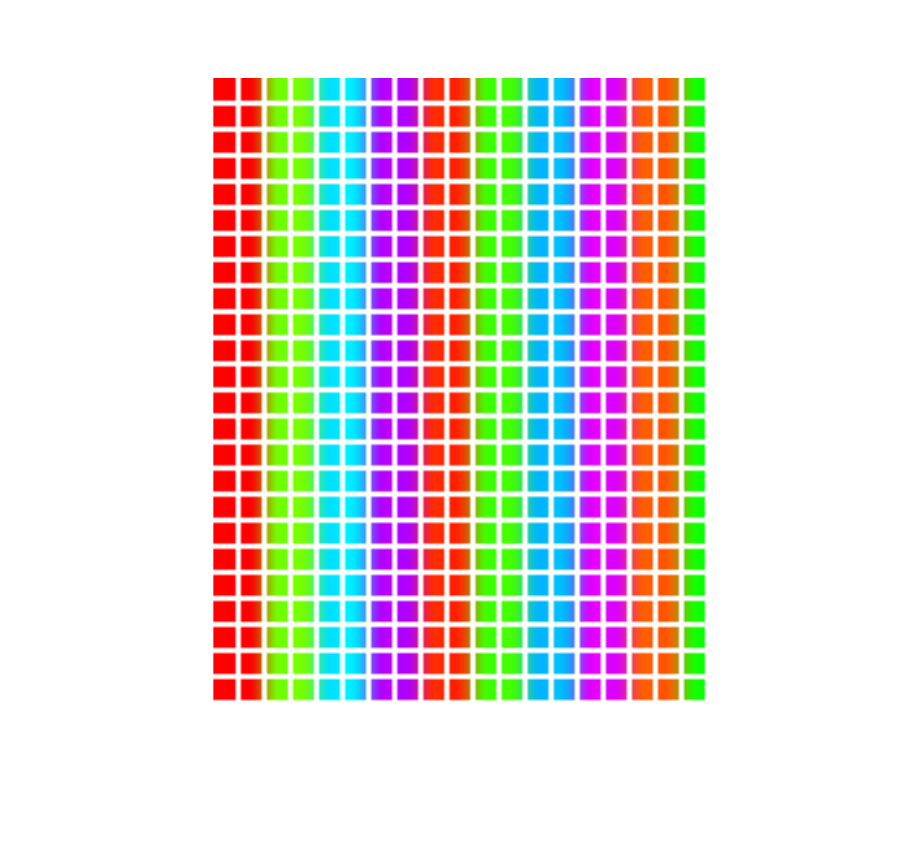}
\end{minipage}}
\subfigure{
\begin{minipage}[b]{0.18\linewidth}
    \centering
   \includegraphics[width=1.4\linewidth]{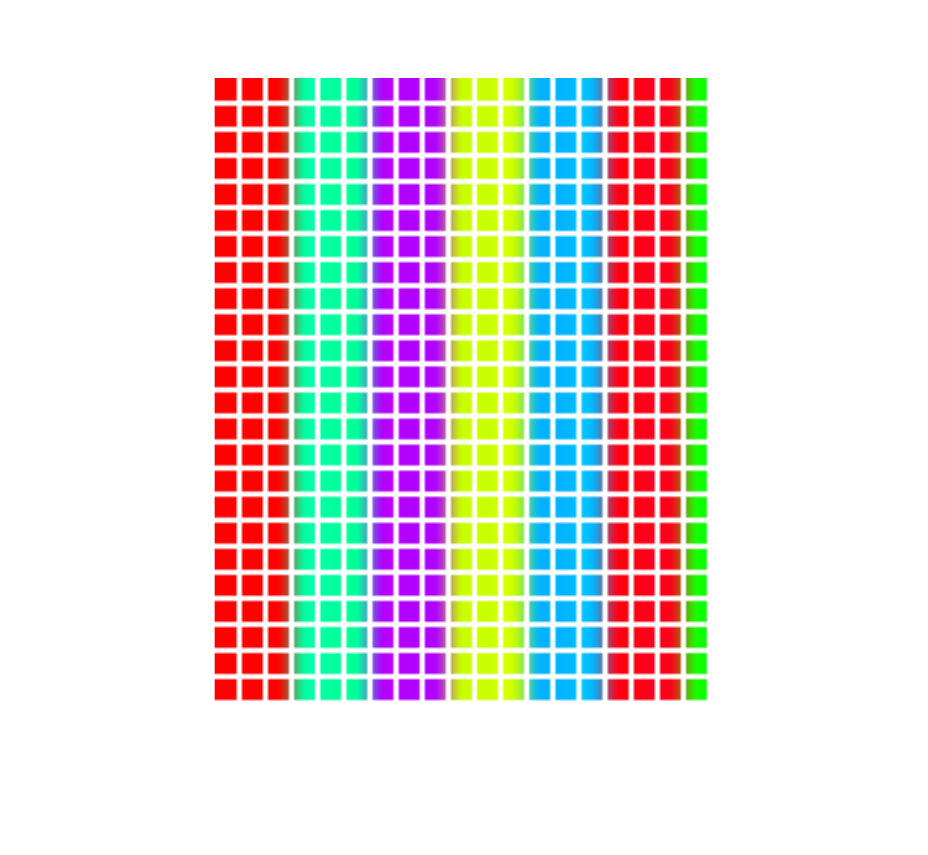}
\end{minipage}}
\subfigure{
\begin{minipage}[b]{0.18\linewidth}
    \centering
   \includegraphics[width=1.4\linewidth]{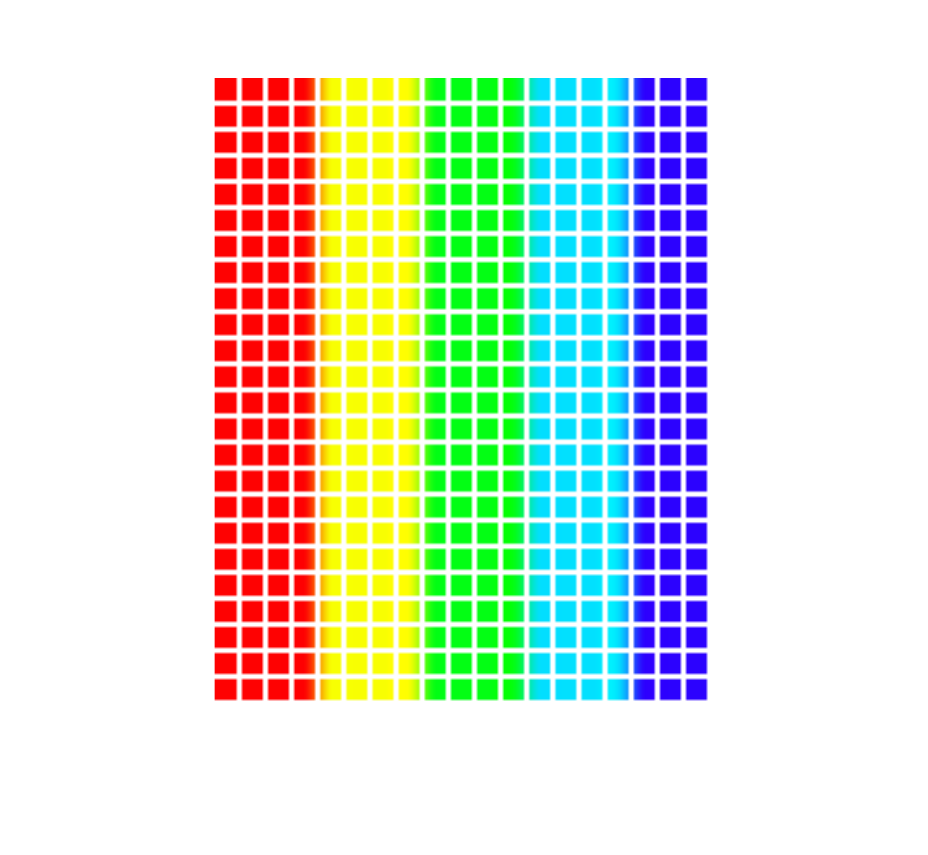}
\end{minipage}}
\subfigure{
\begin{minipage}[b]{0.18\linewidth}
    \centering
   \includegraphics[width=1.4\linewidth]{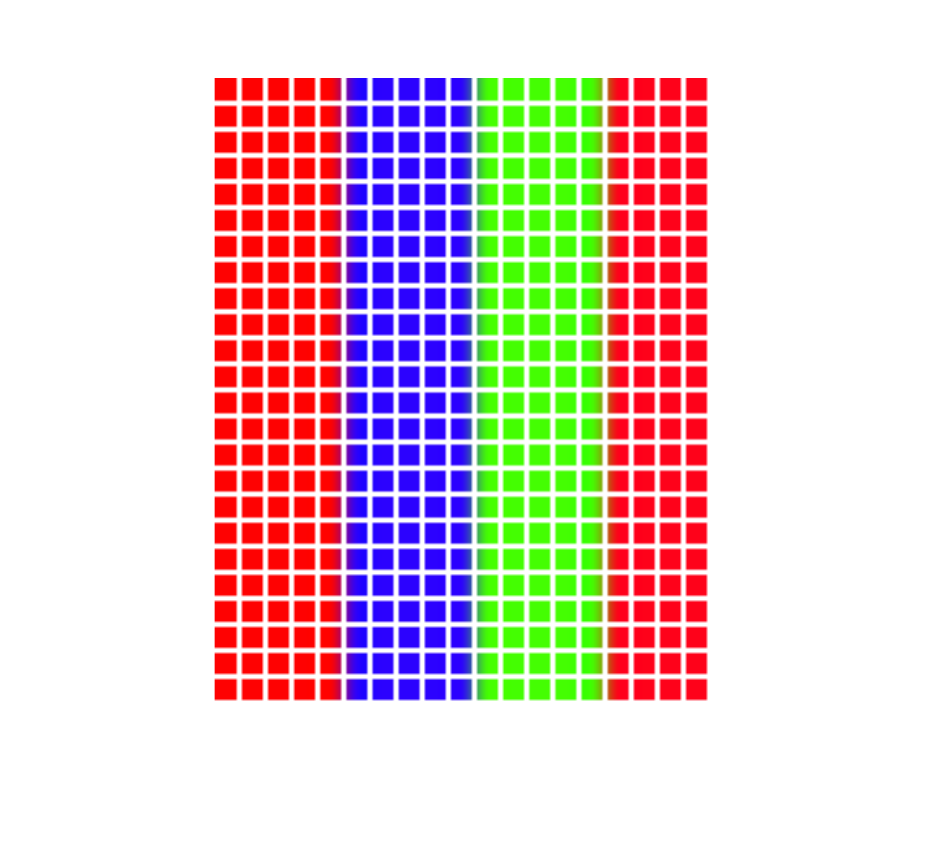}
\end{minipage}}\\
\vspace{-8mm}
\hspace{-15mm}
\subfigure{
\begin{minipage}[b]{0.18\linewidth}
    \centering
   \includegraphics[width=1.4\linewidth]{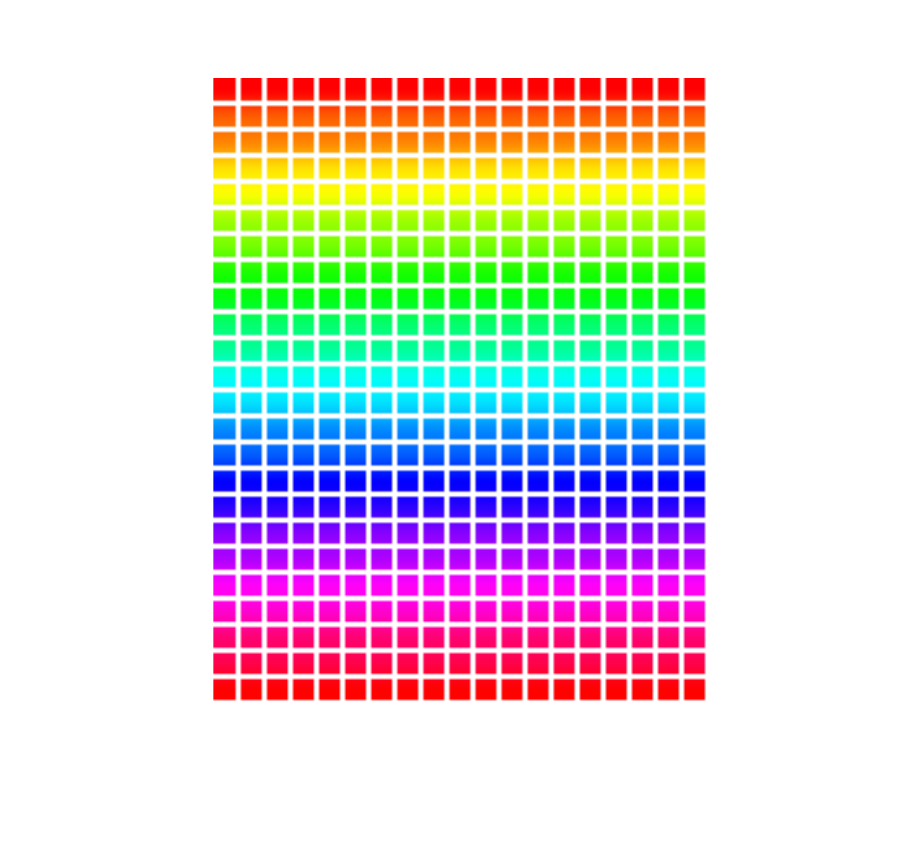}
\end{minipage}}
\subfigure{
\begin{minipage}[b]{0.18\linewidth}
    \centering
   \includegraphics[width=1.4\linewidth]{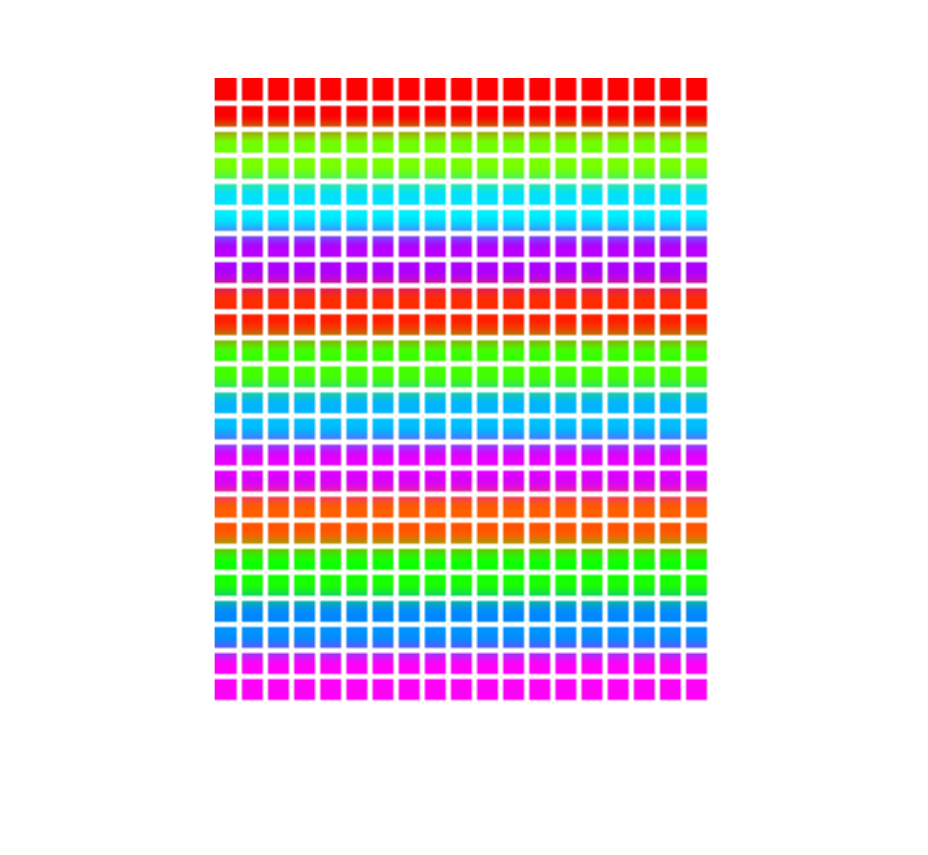}
\end{minipage}}
\subfigure{
\begin{minipage}[b]{0.18\linewidth}
    \centering
   \includegraphics[width=1.4\linewidth]{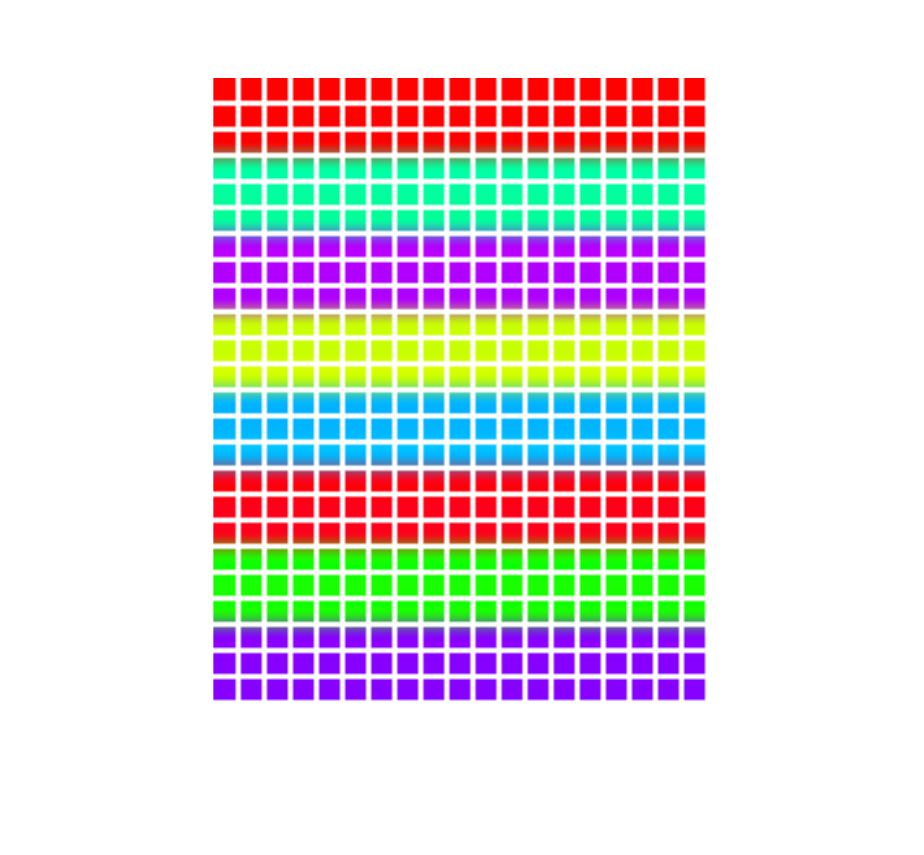}
\end{minipage}}
\subfigure{
\begin{minipage}[b]{0.18\linewidth}
    \centering
   \includegraphics[width=1.4\linewidth]{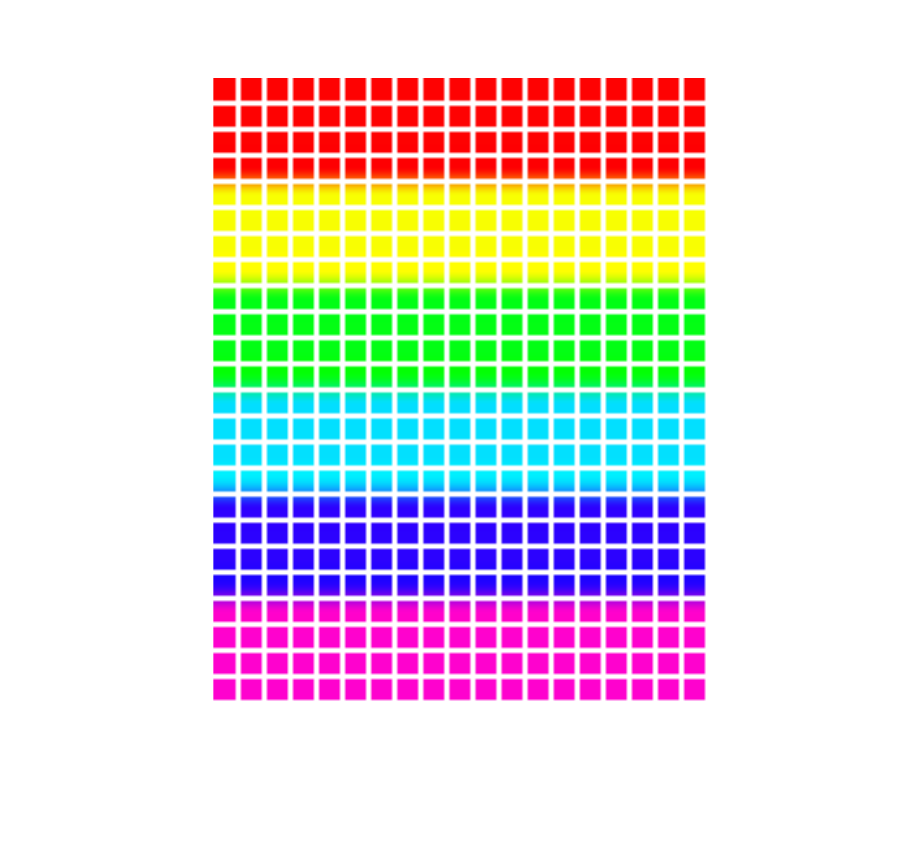}
\end{minipage}}
\subfigure{
\begin{minipage}[b]{0.18\linewidth}
    \centering
   \includegraphics[width=1.4\linewidth]{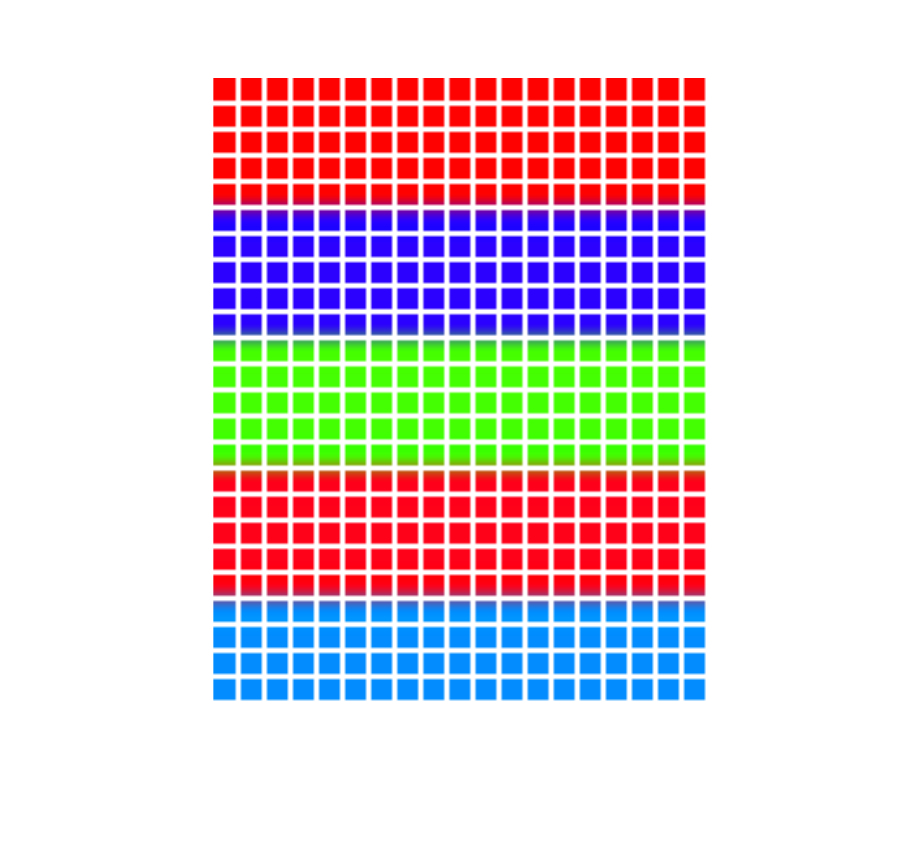}
\end{minipage}}\\
\vspace{-8mm}
\hspace{-13.5mm}
\subfigure{
\begin{minipage}[b]{0.18\linewidth}
    \centering
   \includegraphics[width=1.4\linewidth]{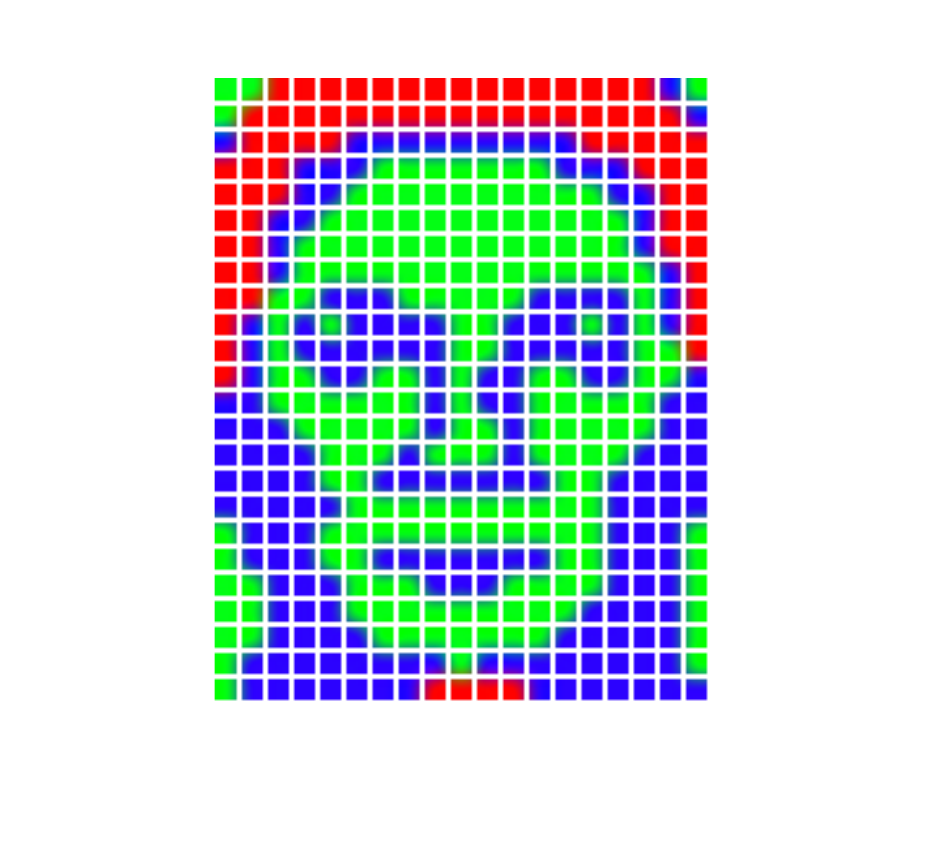}
\end{minipage}}
\subfigure{
\begin{minipage}[b]{0.18\linewidth}
    \centering
   \includegraphics[width=1.4\linewidth]{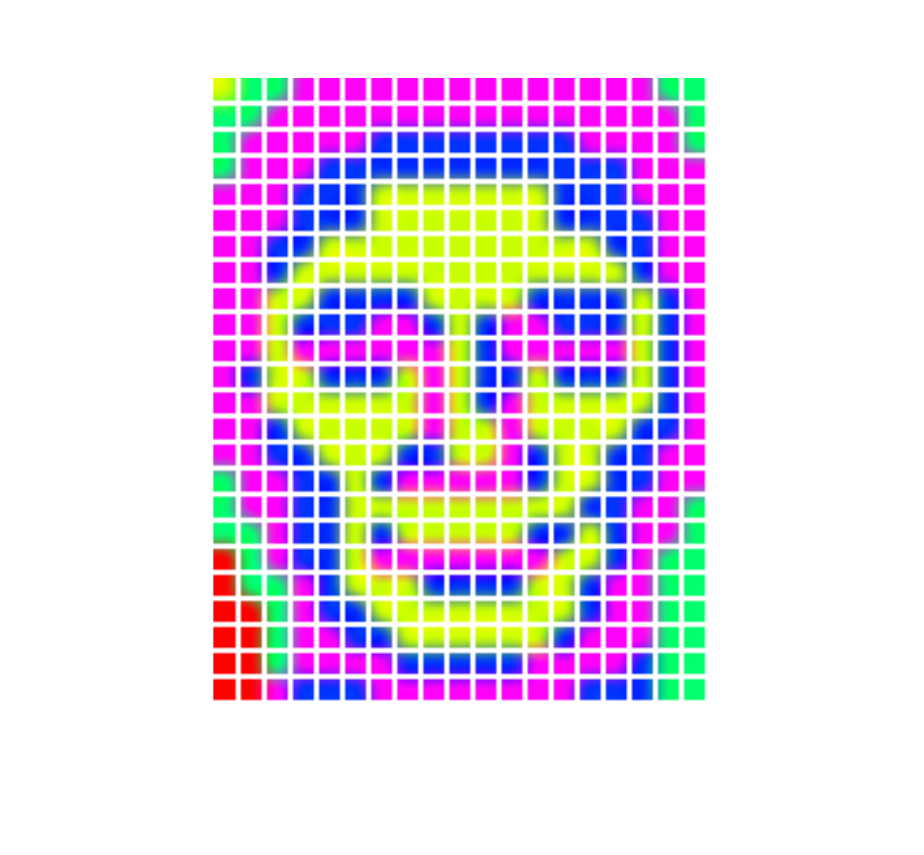}
\end{minipage}}
\subfigure{
\begin{minipage}[b]{0.18\linewidth}
    \centering
   \includegraphics[width=1.4\linewidth]{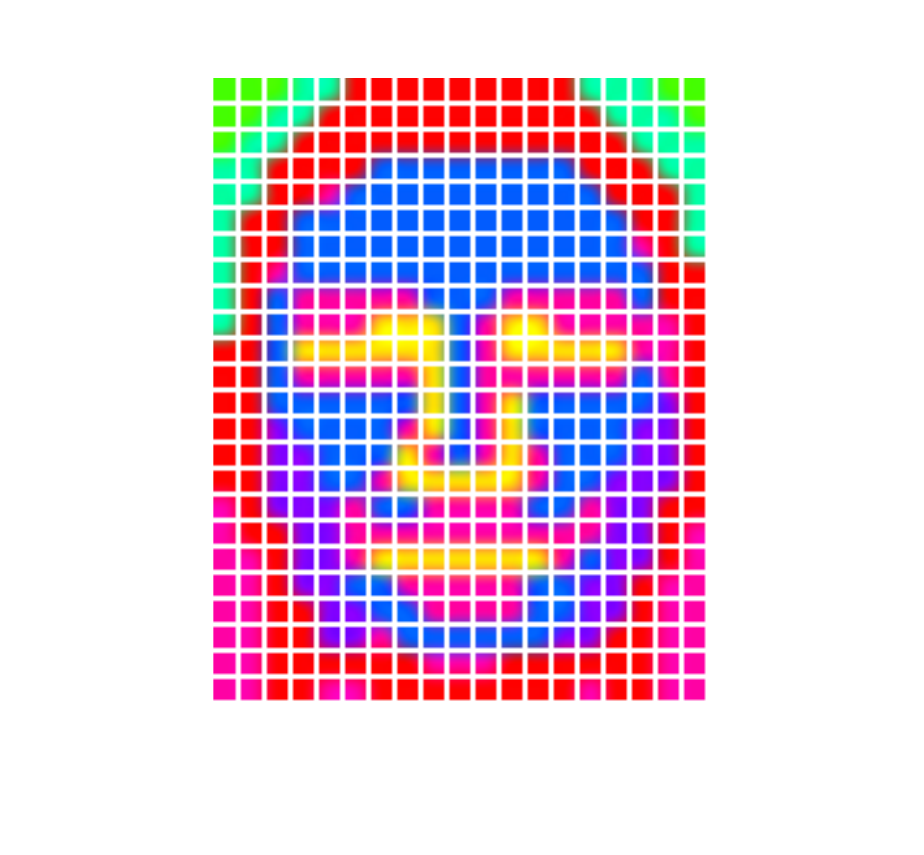}
\end{minipage}}
\subfigure{
\begin{minipage}[b]{0.18\linewidth}
    \centering
   \includegraphics[width=1.4\linewidth]{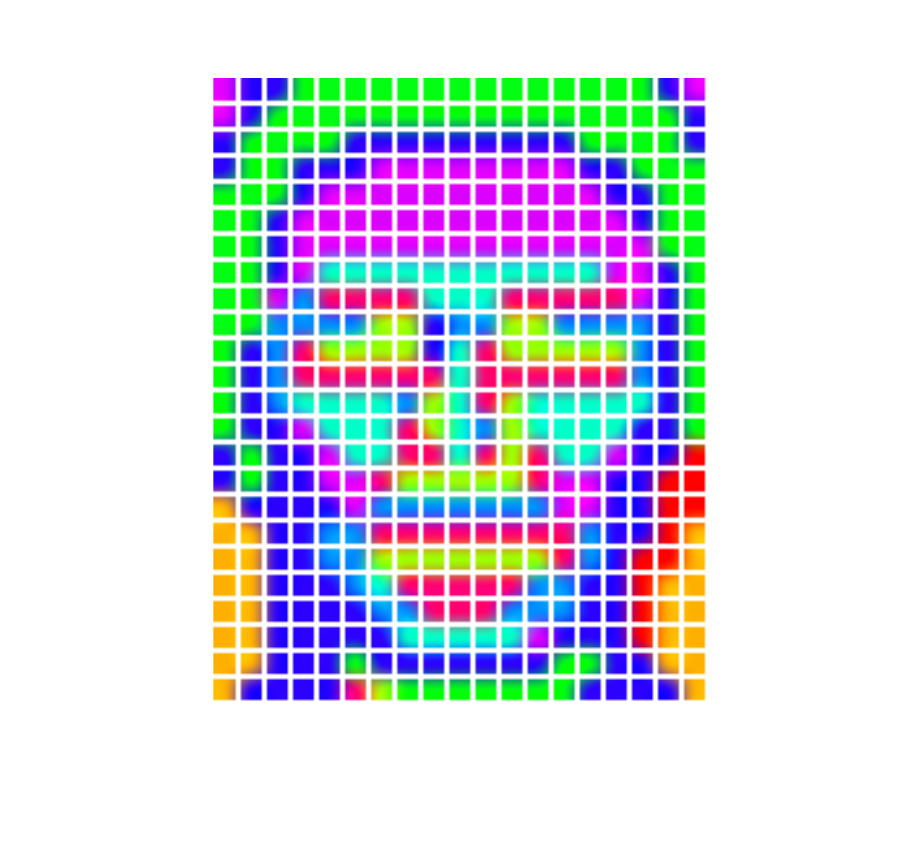}
\end{minipage}}
\subfigure{
\begin{minipage}[b]{0.18\linewidth}
    \centering
   \includegraphics[width=1.4\linewidth]{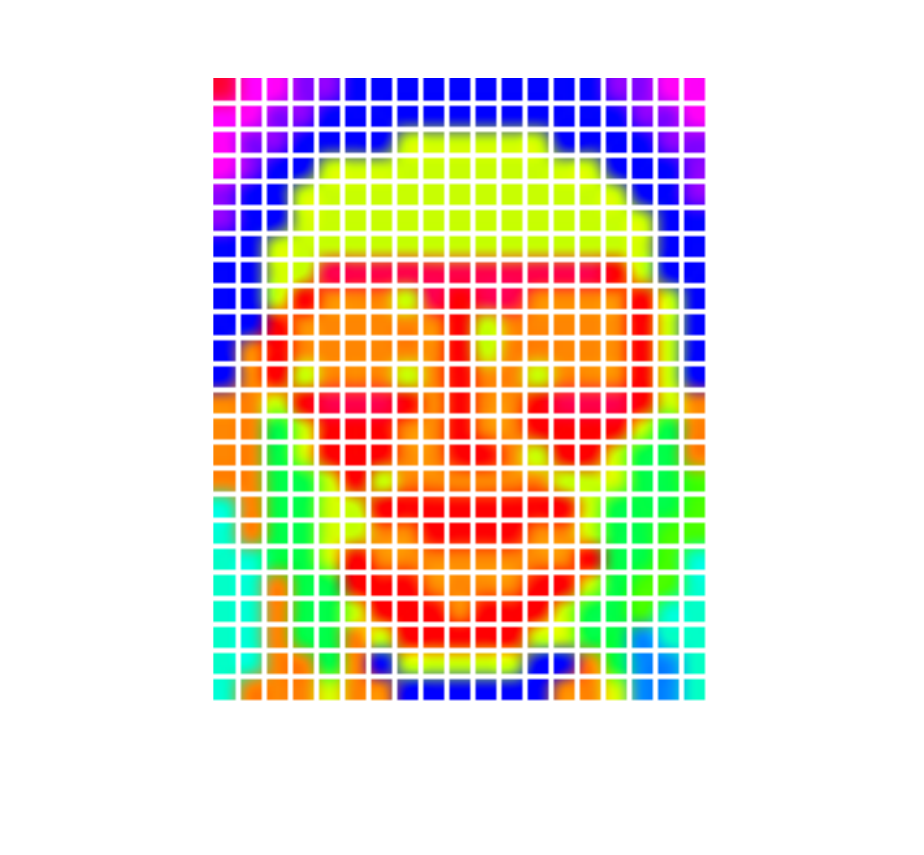}
\end{minipage}}
\vspace{-3mm}
\end{center}
\vspace{-3mm}
   \caption{Illustrations of the three proposed spatial partition strategies in section \ref{subsection III-B}. In this paper the face images are divided into patches and the representations of the patches in the same spatial partition region are concatenated together for matching. In the images of the first two rows above, it should be noticed that we demonstrate the results of taking $K_c$ columns (or $K_r$ rows) of patches as the same spatial partition region and represent these patches by the same color. The first row shows the column-based spatial partition strategy when $K_c=$ 1,2,3,4,5. The second row shows the row-based spatial partition strategy when $K_r=$ 1,2,3,4,5. The last row shows the learning-based spatial partition strategy when $K_l=$ 3,5,7,9,11. These are results on the CUHK face sketch FERET database with SIFT as the feature descriptor.}
\label{Figure2}
\end{figure*}
\subsection{Spatial Partition-based Discriminant Analysis}
\label{subsection III-B}

\begin{figure*}
\begin{center}
    \includegraphics[width=0.9\linewidth]{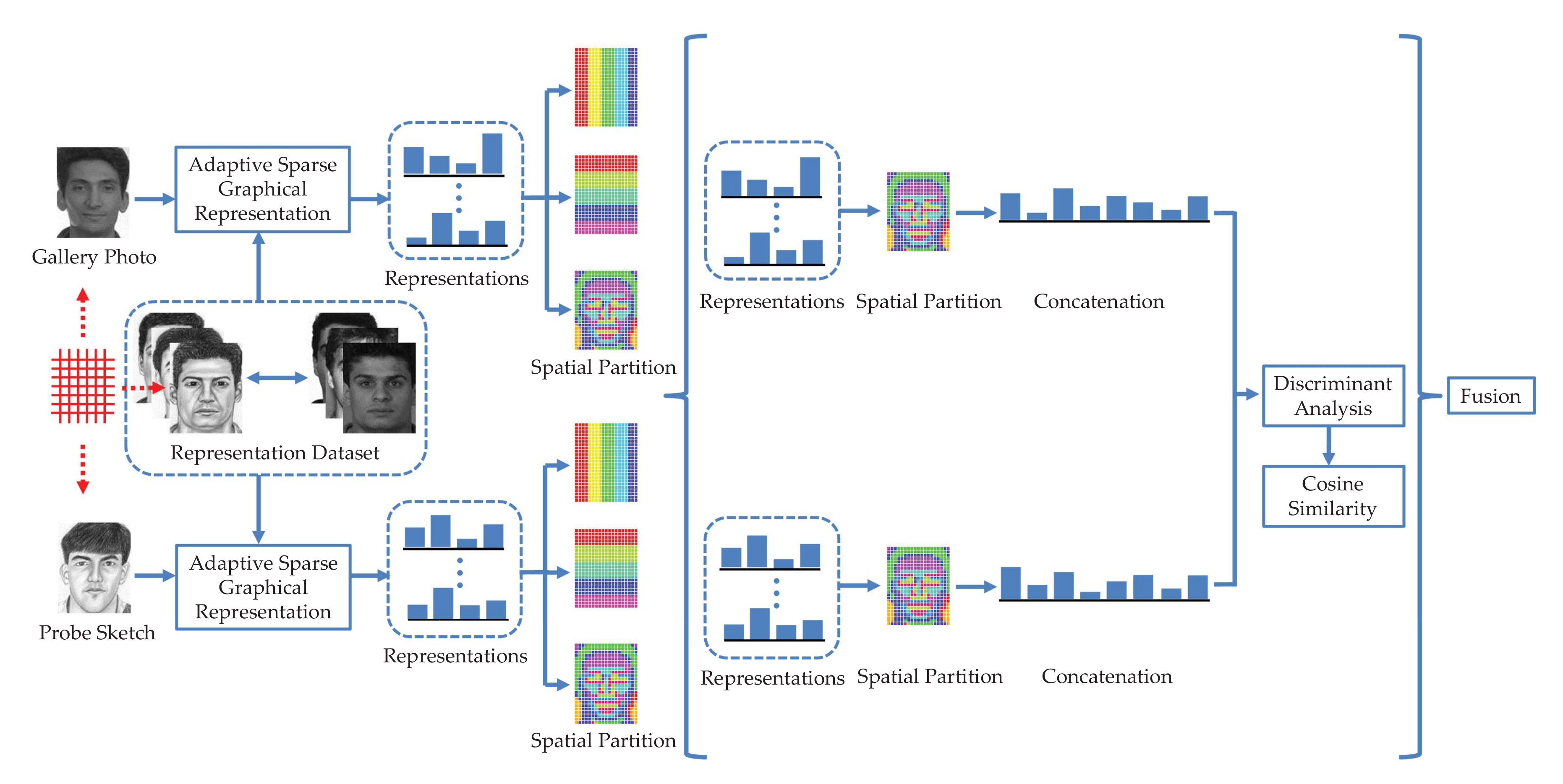}
\end{center}
   \caption{Overview of the proposed SGR-DA method for HFR.}
\label{Figure3}
\end{figure*}

After obtaining the adaptive sparse graphical representations of both probe sketches and gallery photos, we refine these representations through discriminant analysis for face matching. The representations of all face image patches can be simply concatenated together and then apply classical subspace analysis, such as principal component analysis (PCA) \cite{Ref35} and linear discriminant analysis (LDA) \cite{Ref36} to extract discriminative information for matching. However, the facial structure is complex and this direct concatenation approach neglects the spatial facial structure. Furthermore, it is likely to be overfitting due to the small sample size \cite{Ref7}.

In order to handle the complex facial structure and improve the discriminability, many discriminant analysis strategies have been proposed. The LFDA approach \cite{Ref7} divided face image patches into ``slices'', where ``slices'' correspond to the concatenation of features from each column of image patches. However, the drawback of LFDA is that combining only one column of image patches together may not be the most discriminative strategy. On the other hand, it is also practical to concatenate features from several rows of images patches, which was not exploited in \cite{Ref7}. There are other methods that divide face image into local regions manually. For example, the CITE approach \cite{Ref8} divided a face image into 7$\times$5 local regions with equal size. The method \cite{Ref9} manually divided a face image into five local regions corresponding to five facial components (eye, eyebrow, checks, nose, and mouth). These approaches suffer the same problem that the local regions are designed manually, without the consideration of the characteristics of face image data. The semantic pixel sets based method \cite{RefTC2} exploited the semantic pixel relation by intensity distribution and clustered face regions by the pixel intensity values. However, as illustrated below, this clustering-based strategy is complementary to the column-based and row-based strategy. The fusion of combining different spatial partition strategies to further improve the discriminability has not been investigated yet.

Three spatial partition strategies are developed in this paper to address the drawbacks of aforementioned approaches and improve the discriminability:

(1) Considering the shortcoming of combining only one column of image patches together, we propose to combine $K_c$ columns of image patches as a spatial partition region. Discriminant analysis can then be performed separately on each spatial partition region. The extracted features are then concatenated together for matching. The column-based spatial partition strategy is shown in the first row of Fig. \ref{Figure2}, when $K_c=$ 1,2,3,4,5. It should be noticed that we demonstrate the results of taking $K_c$ columns of patches as the same spatial partition region and represent these patches by the same color in the first row of Fig. \ref{Figure2}.

(2) In order to exploit the row-based spatial partition strategy, we combine $K_r$ rows of image patches as a spatial partition region. Discriminant analysis can be performed on each region similarly. The illustration when $K_r=$ 1,2,3,4,5 is shown in the second row of Fig. \ref{Figure2}.

(3) Instead of manually dividing face image into local regions, we further exploit learning-based spatial partition strategy. The image patches can be clustered together through machine learning techniques. We use $K$-means clustering \cite{Ref37} here. The features of image patches at the same location of each face image are concatenated as a vector. The purpose is to cluster the locations of face images in the dataset. In our experiments, the clusters are determined by the long feature vectors created through concatenating the feature descriptors from coupled heterogeneous face images across the training set. Therefore, different heterogeneous face datasets (\emph{e.g.}, face sketch-photo dataset and NIR-VIS dataset) may result in different clustering results. Illustrations of learning-based spatial partition strategy on the CUHK face sketch FERET database is shown in the last row of Fig. \ref{Figure2} when the cluster number $K_l=$ 3,5,7,9,11. Discriminant analysis can be performed on each clustered region.

The effects of different $K_c$, $K_r$, and $K_l$ will be discussed in the experiment section, and the best $K_c$, $K_r$, and $K_l$ are used. In the discriminant analysis process performed on each spatial partition region, PCA is firstly applied with 99 percent of the variance preserved. Subsequently, LDA is performed to further reduce the dimensionality and improve the discriminability. Finally, all the projected vectors of the same face image are concatenated and the cosine similarity measure is used to calculate the similarity score between a probe sketch and a gallery photo.

We further investigate that the proposed column-based, row-based, and learning-based spatial partition strategies are complementary. The fusion of these three spatial partition strategies can naturally enhance the recognition performance. Details are given in the experiment section. In our work, we simply sum the similarity scores of different spatial partition strategies after a min-max score normalization.

\subsection{SGR-DA Method for HFR}
\label{subsection III-C}

In order to better illustrate the proposed approach, the whole approach is outlined in Fig. \ref{Figure3}. Firstly, the face images are divided into patches, and common feature descriptors (SIFT, for example) are used to represent each image patch. Secondly, for a probe sketch (or a gallery photo), a Markov networks model is constructed on the features of probe sketch patches (or gallery photo patches) and sketch patches (or photo patches) in representation dataset. The adaptive sparse graphical representations of the input image can then be generated by solving (\ref{Eq:eq 3}). Thirdly, the column-based, row-based, and learning-based spatial partition strategies are applied to refine the adaptive sparse graphical representations and improve its discriminability. Finally, the cosine similarity measure is used to calculate the similarity score of the three refined vectors, which are then fused. A nearest neighbor matcher is used for recognition in the end.

\section{Experiments}
\label{section IV}

In this section, we evaluate our SGR-DA through extensive experiments on six commonly used heterogeneous face datasets: the CUHK Face Sketch FERET Database (CUFSF) \cite{Ref8}, PRIP Viewed Software-Generated Composite Database (PRIP-VSGC) \cite{Ref1}, IIIT-D Semi-Forensic Sketch Database \cite{Ref38}, Forensic Sketch Database \cite{Ref4}, CASIA NIR-VIS 2.0 Face Database \cite{Ref39}, and Natural Visible and Infrared facial Expression Database (USTC-NVIE) \cite{Ref40}. We first introduce the experimental settings of our method and evaluate the effectiveness of our contributions. Then we illustrate that our approach achieves superior performance in comparison with state-of-the-art methods on these six datasets.

Three baseline results are provided in this section: Fisherface algorithm \cite{Ref36}, the open source face recognition algorithm OpenBR \cite{Ref31}, and the state-of-the-art HFR algorithm P-RS \cite{Ref3}. For the Fisherface algorithm, we combine the heterogeneous face images together to train the projection matrix and the Euclidean distance is used for matching. For the OpenBR algorithm, we use the public source which is freely available online\footnote{http://openbiometrics.org/.}. For the P-RS algorithm, we implemented the prototype random subspace framework and the direct random subspace (D-RS) framework. The results of fusing P-RS and D-RS are reported. Note that the results in this paper are reported as 10-fold cross validation by randomly splitting the training and testing sets.

Three features are utilized in this section: SIFT feature, speeded up robust features (SURF) \cite{Ref42} feature, and histograms of oriented gradients (HOG) \cite{Ref43} feature. The SURF feature is extracted by exploiting the implementation embedded in the MATLAB R 2012b software. The center of the image patch is set as the interest point and the standard SURF-64 version is utilized. The SIFT feature and HOG feature are extracted through an open source library\footnote{http://www.vlfeat.org/.}. For the SIFT feature the center of the image patch is set as the interest point. A 128-dimensional vector and a 124-dimensional vector are generated for SIFT and HOG respectively.

\subsection{Experimental Settings}
\label{subsection IV-A}

All the heterogeneous face images are aligned based on five facial points (centers of two eyes, nose tip, left and right mouth corner), which are automatically detected by \cite{Ref41}. Because the facial point detection method \cite{Ref41} failed on the TIR images in the USTC-NVIE database, we manually located the five points on the TIR images. After the facial points are located, each face image is cropped to 100$\times$125 based on the five points. The image patch size is 10$\times$10, and 50\% overlapping ratio is kept. The size of the search region $R$ is 16. We further conduct adjustment experiments on the CUFSF database to determine other parameters as well as evaluate the effectiveness of our contributions. 250 sketch-photo pairs in CUFSF consist of the representation dataset and other 250 pairs are used for training. The rest 694 pairs are used for testing (There are 1194 face sketch-photo pairs in this database in total). In our experiments, it is evaluated that little influence will be introduced when different sources of the representation dataset are chosen. The only principle is that the images in the representation dataset should not appear in the training set or the testing set again.

The most time-consuming part of the proposed approach lies in the extraction process of the adaptive sparse graphical representation. Although in the proposed method the $K$ nearest neighbor searching process is skipped, we still need to find the best image patch on each face image in the representation dataset within the search region. Therefore, the complexity of this process is $O(P_cP_MMP_f)$. Here $P_c$ is the number of candidates in the search region around one patch. $P_M$ is the number of patches per image. $M$ is the number of face image pairs in the representation dataset and $P_f$ is the dimension of the local descriptor. During the discriminant analysis process, standard PCA and LDA are applied before matching. In our experiments, it takes approximately five minutes to encode one face image through the proposed adaptive sparse graphical representation. This may hinder the usage of the proposed representation in a part of real applications. However, in real law enforcement scenarios it is quite acceptable since several days even months are usually taken to search a suspect by human beings. In such applications, the recognition performance is more important than the speed of the encoding process. All the experiments in this paper are conducted on an Intel Core i7-4790 3.6GHz PC under MATLAB R 2012b environment.

\begin{figure}[t]
\begin{center}
    \includegraphics[width=0.9\linewidth]{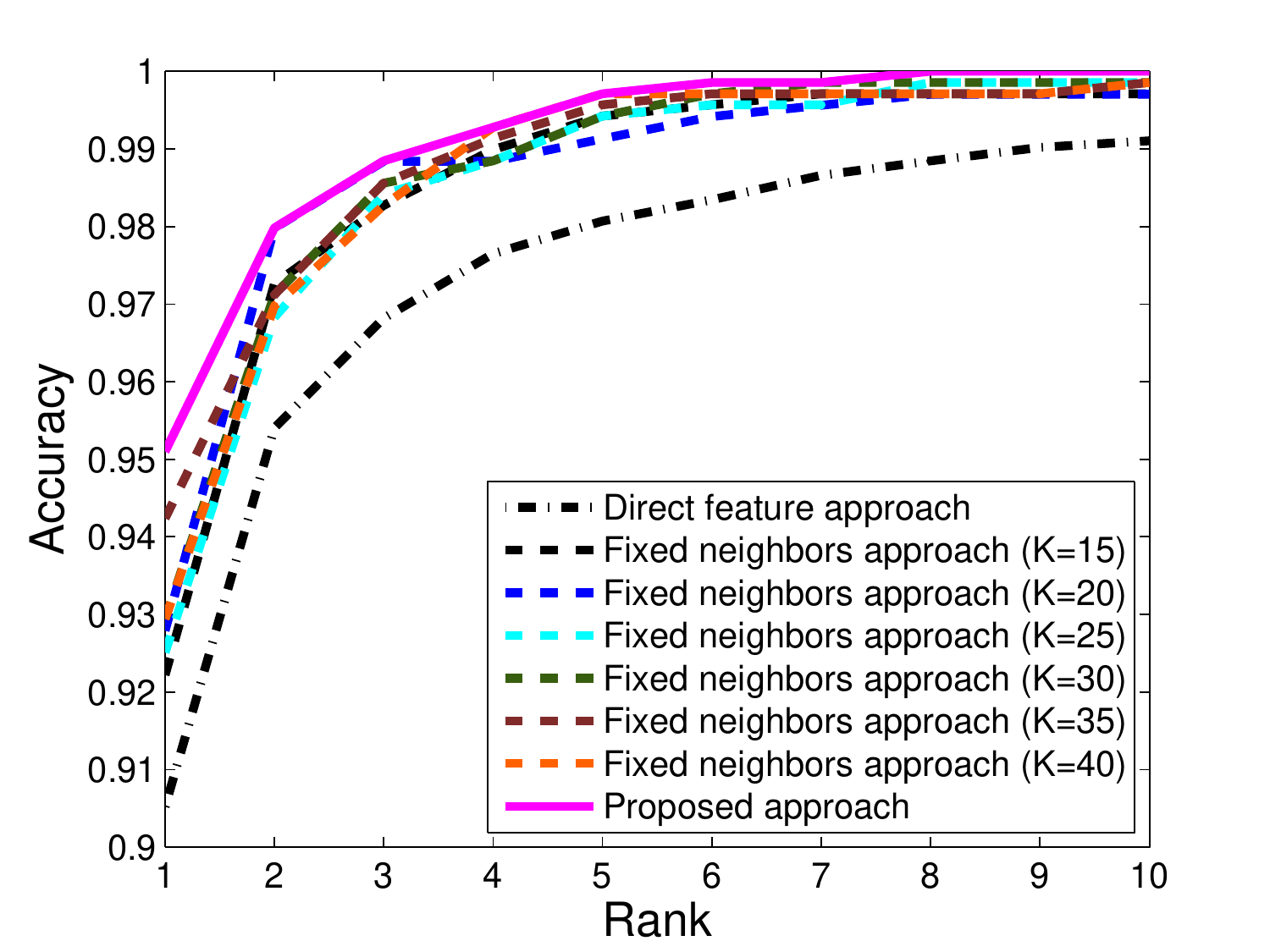}
\end{center}
   \caption{Evaluation of the proposed adaptive sparse graphical representation.}
\label{Figure4}
\end{figure}

We first evaluate the effectiveness of the proposed adaptive sparse graphical representation. The SURF feature is utilized as the feature descriptor to represent image patches. Because there are 250 pairs in the representation dataset, the size of related neighbors $M$ in our method is 250. To compare with existing methods \cite{Ref5,Ref6,Ref4} which manually selected $K$ nearest neighbors, we set the number of nearest neighbors $K$ to be 15, 20, 25, 30, 35, 40 and the accuracies of different fixed number of related neighbors are shown in Fig. \ref{Figure4}, denoted as ``Fixed neighbors approach''. We further implemented a direct feature based approach, by replacing the adaptive sparse graphical representation with the original SURF feature, denoted as ``Direct feature approach''. As shown in Fig. \ref{Figure4}, the proposed adaptive sparse graphical representation is superior to existing $K$ nearest related neighbor selection strategy.

\begin{figure}[t]
\begin{center}
\subfigure{
\label{Figure5_1}
\begin{minipage}[b]{0.47\linewidth}
    \centering
   \includegraphics[width=1.1\linewidth]{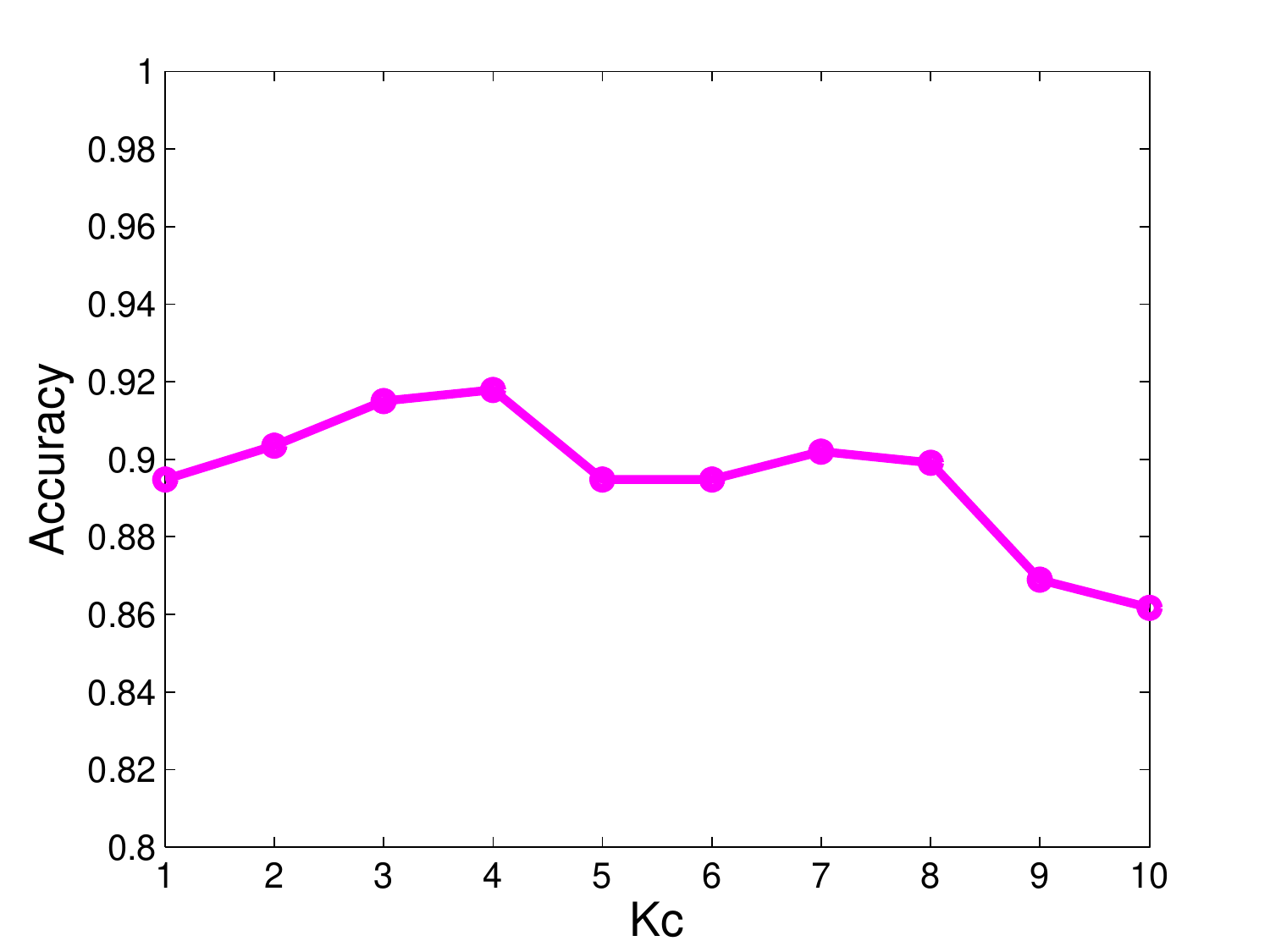}
\end{minipage}}%
\subfigure{
\label{Figure5_2}
\begin{minipage}[b]{0.47\linewidth}
    \centering
   \includegraphics[width=1.1\linewidth]{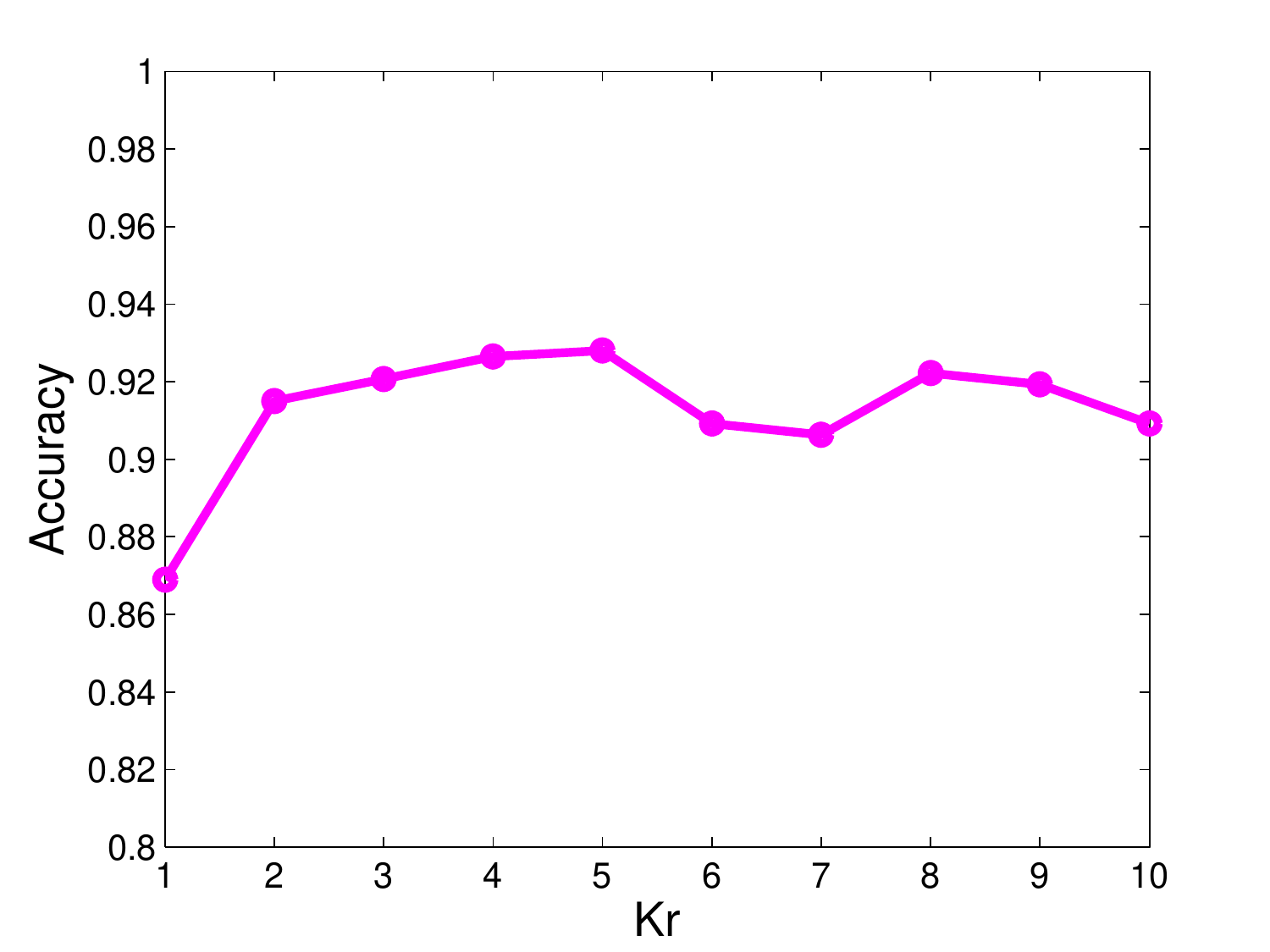}
\end{minipage}}
\subfigure{
\label{Figure5_3}
\begin{minipage}[b]{0.47\linewidth}
    \centering
   \includegraphics[width=1.1\linewidth]{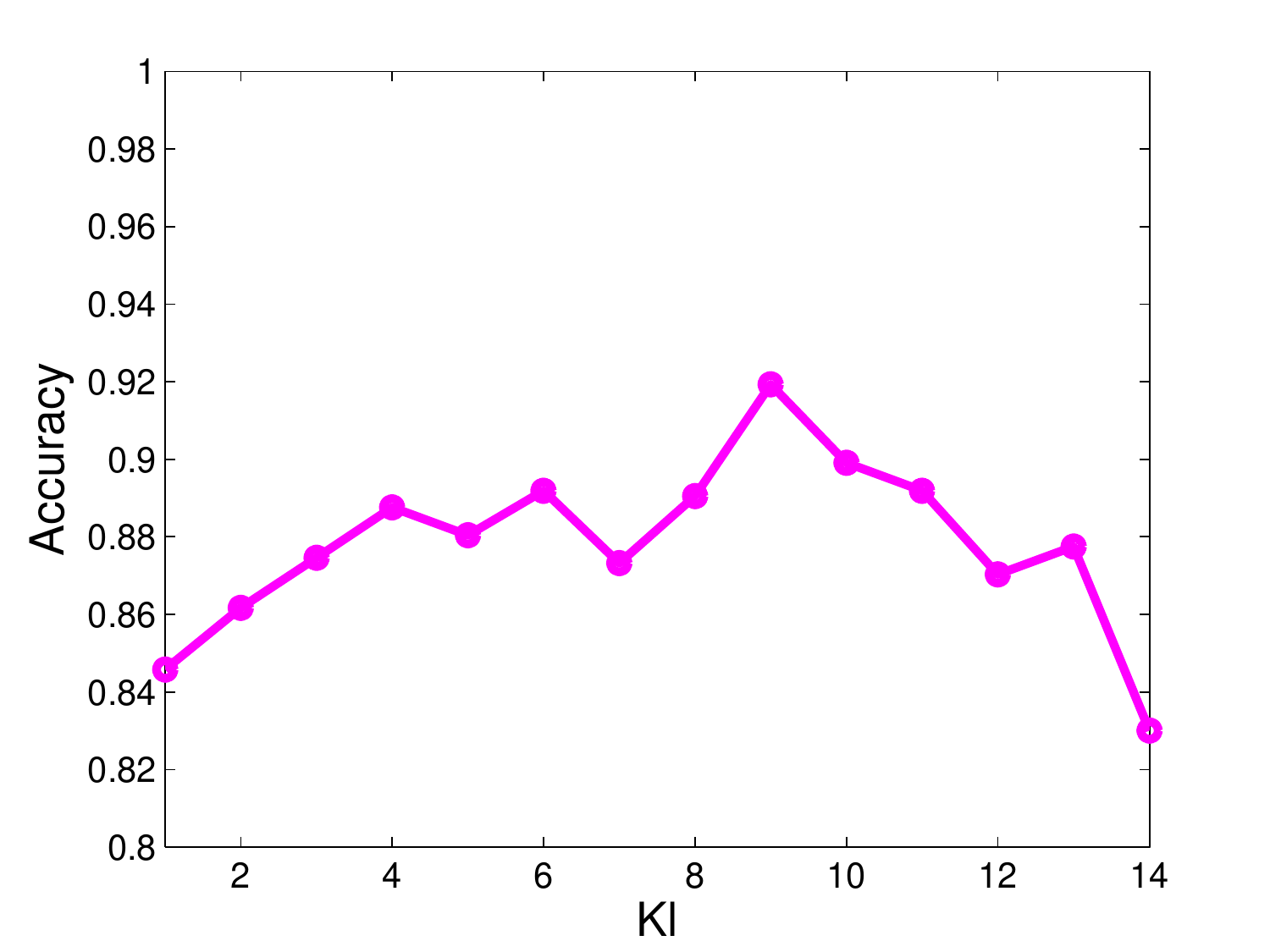}
\end{minipage}}%
\subfigure{
\label{Figure5_4}
\begin{minipage}[b]{0.47\linewidth}
    \centering
   \includegraphics[width=1.1\linewidth]{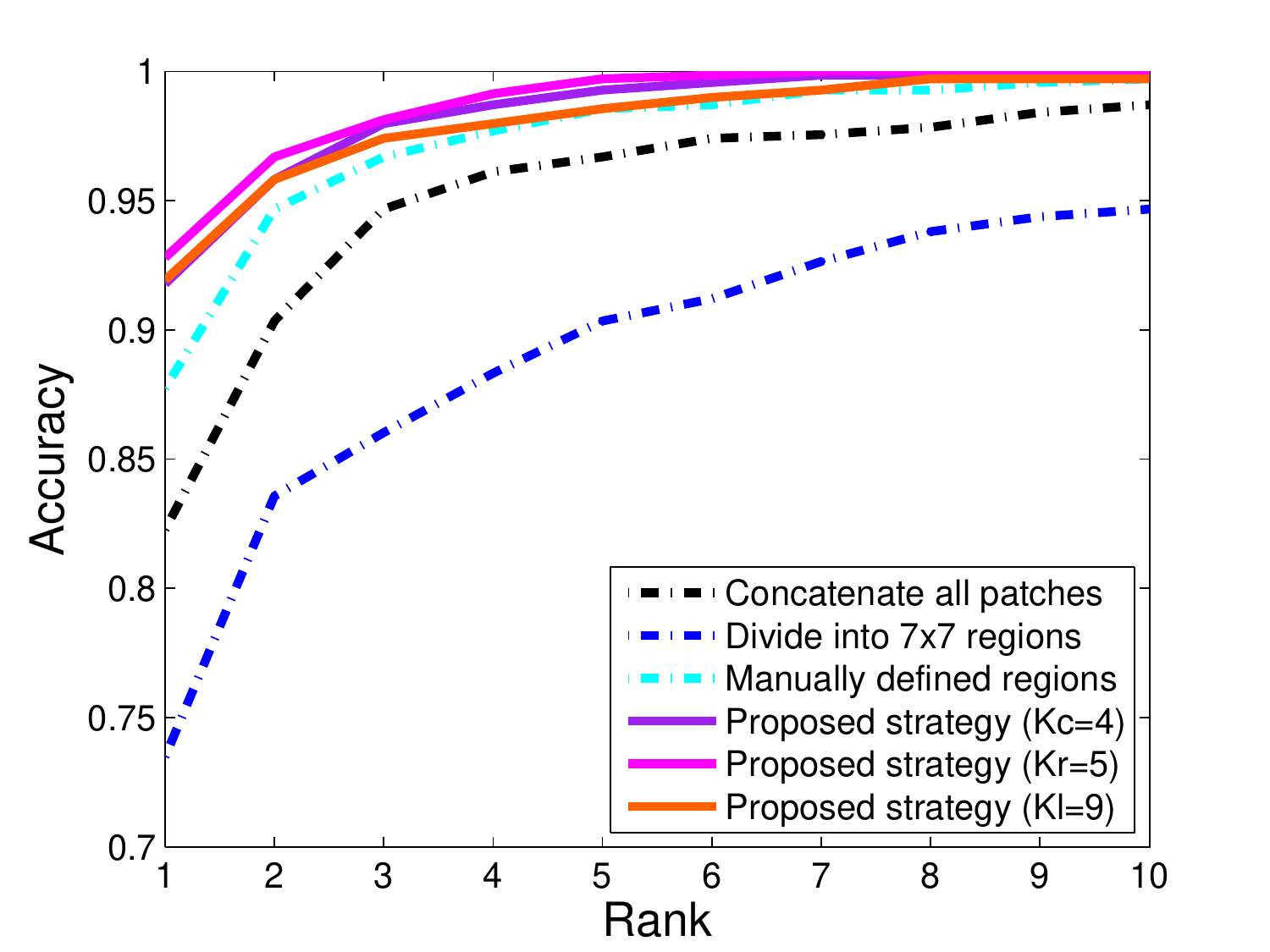}
\end{minipage}}
\end{center}
   \caption{Evaluation of the proposed spatial partition-based discriminant analysis on CUFSF database.}
\label{Figure5}
\end{figure}

We then demonstrate the effects of different $K_c$, $K_r$, and $K_l$ in the proposed spatial partition-based discriminant analysis framework. The left top subfigure of Fig. \ref{Figure5} shows the rank-1 accuracy when different $K_c$ is set. The right top subfigure shows the results when different $K_r$ is set. The left bottom subfigure shows the results when different $K_l$ is set. It can be seen that $K_c=$ 4, $K_r=$ 5, and $K_l=$ 9 achieve the best accuracies respectively on CUFSF database under our experimental settings (image size and patch size). In order to further illustrate the effectiveness of the proposed spatial partition-based discriminant analysis, we compare it with three conventional strategies without the proposed strategy (concatenating all patches, dividing a face image into 7$\times$7 local regions, and manually defined regions). As shown in right bottom subfigure of Fig. \ref{Figure5}, the proposed spatial partition-based discriminant analysis strategy exploits the characteristics of the data and achieves better performance than conventional strategies. We assume that the best parameters here can be generalized to other datasets, and $K_c$, $K_r$, and $K_l$ are fixed to 4, 5, and 9 respectively in the following experiments.

\begin{figure}[t]
\begin{center}
\subfigure{
\label{Figure6_1}
\begin{minipage}[b]{0.47\linewidth}
    \centering
   \includegraphics[width=1.1\linewidth]{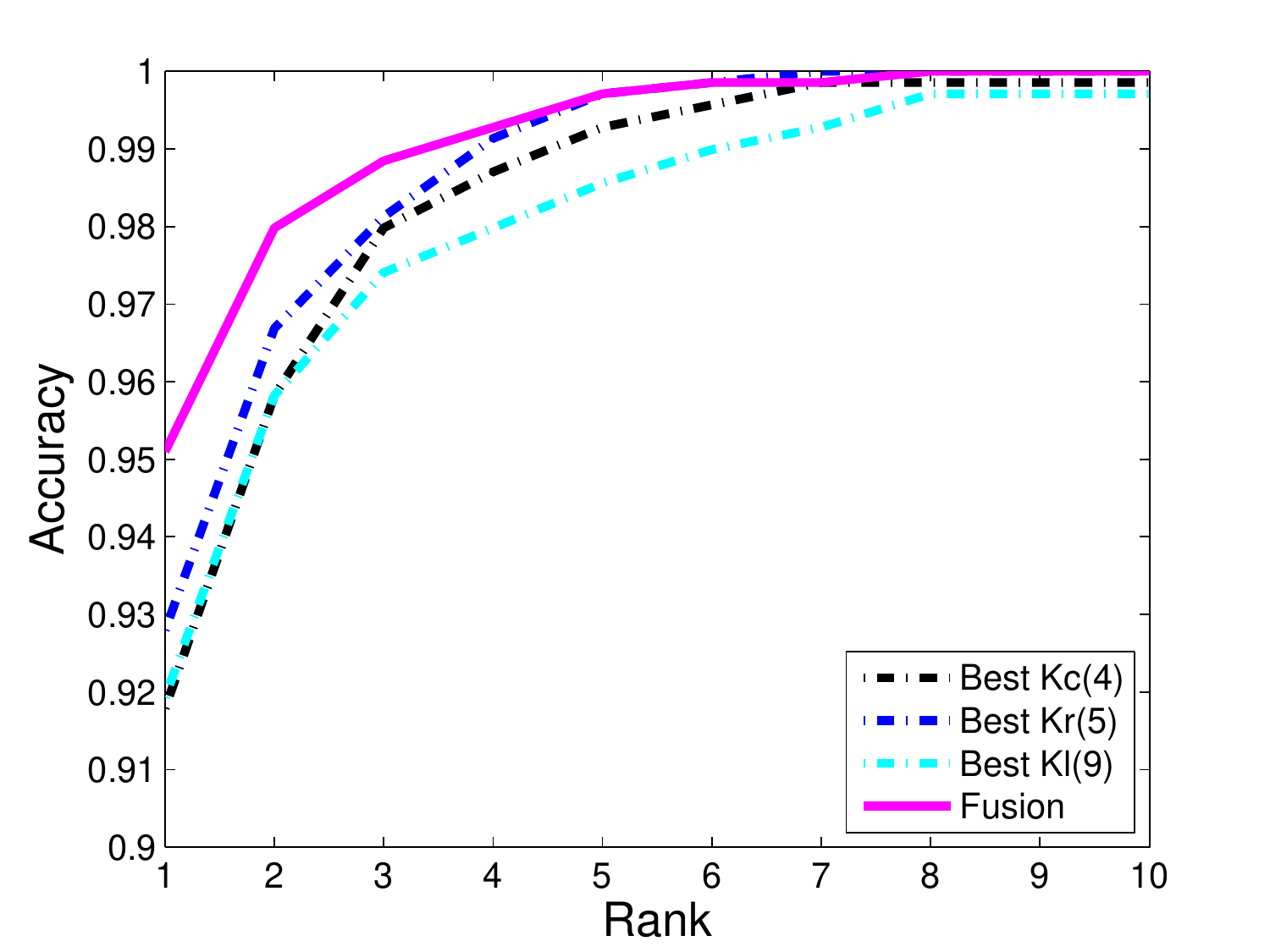}
\end{minipage}}%
\subfigure{
\label{Figure6_2}
\begin{minipage}[b]{0.47\linewidth}
    \centering
   \includegraphics[width=1.1\linewidth]{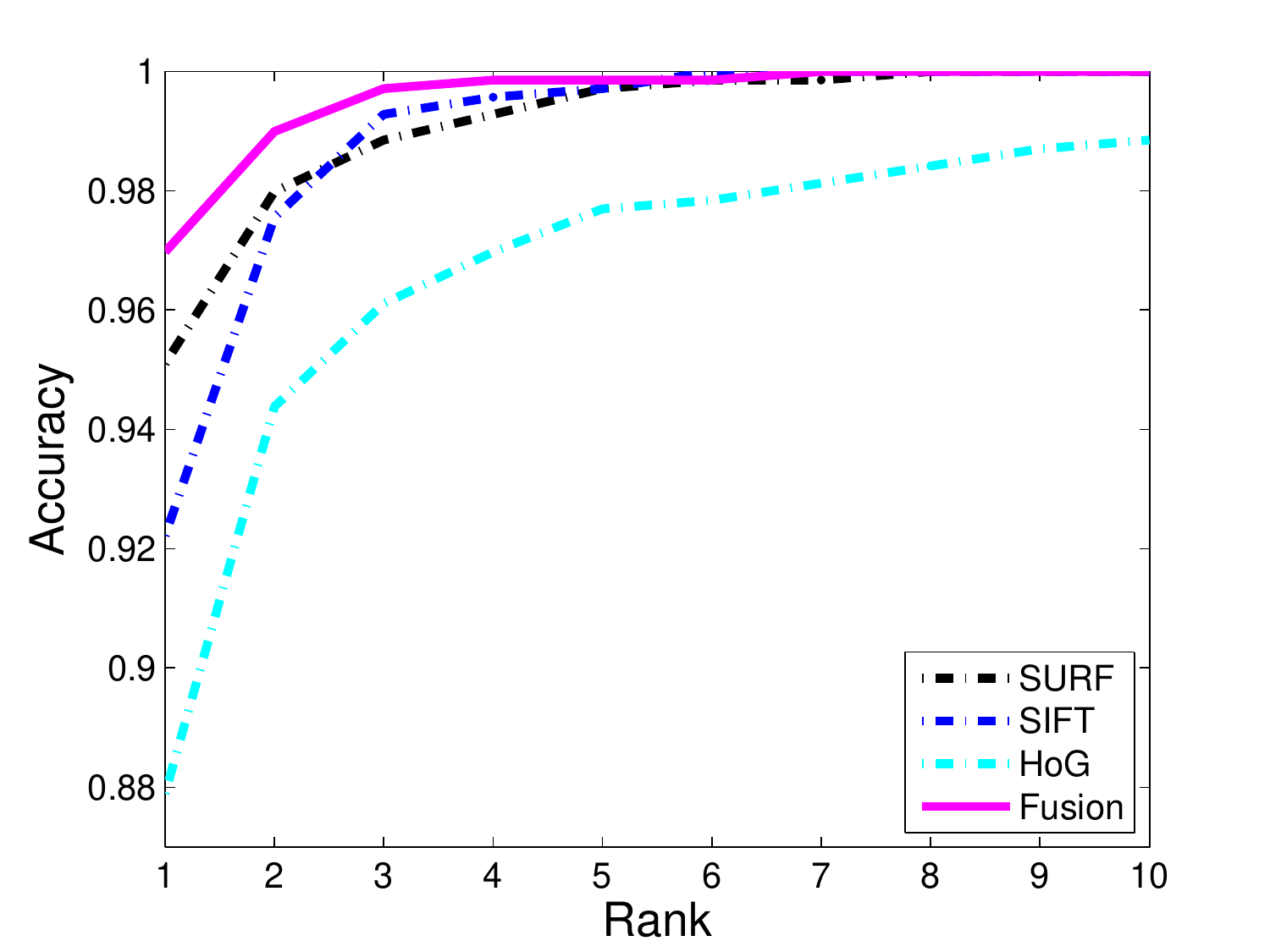}
\end{minipage}}
\end{center}
   \caption{Experiments on fusion of different spatial partition strategies (left subfigure) and different features (right subfigure).}
\label{Figure6}
\end{figure}

The proposed column-based, row-based, and learning-based spatial partition strategies are complementary. Therefore, the fusion of these three spatial partition strategies can naturally enhance the recognition performance, as shown in the left subfigure of Fig. \ref{Figure6}. Because the proposed method represents heterogeneous face images in each modality separately, common features used in conventional face recognition tasks can be used to represent image patches in our method. We utilize SURF feature, SIFT feature, and HOG feature in this paper. We further investigate that fusion of these three features results in improved recognition accuracy, as shown in the right subfigure of Fig. \ref{Figure6}.

\subsection{CUFSF Viewed Sketch Database}
\label{subsection IV-B}

\begin{figure}[t]
\begin{center}
    \includegraphics[width=0.9\linewidth]{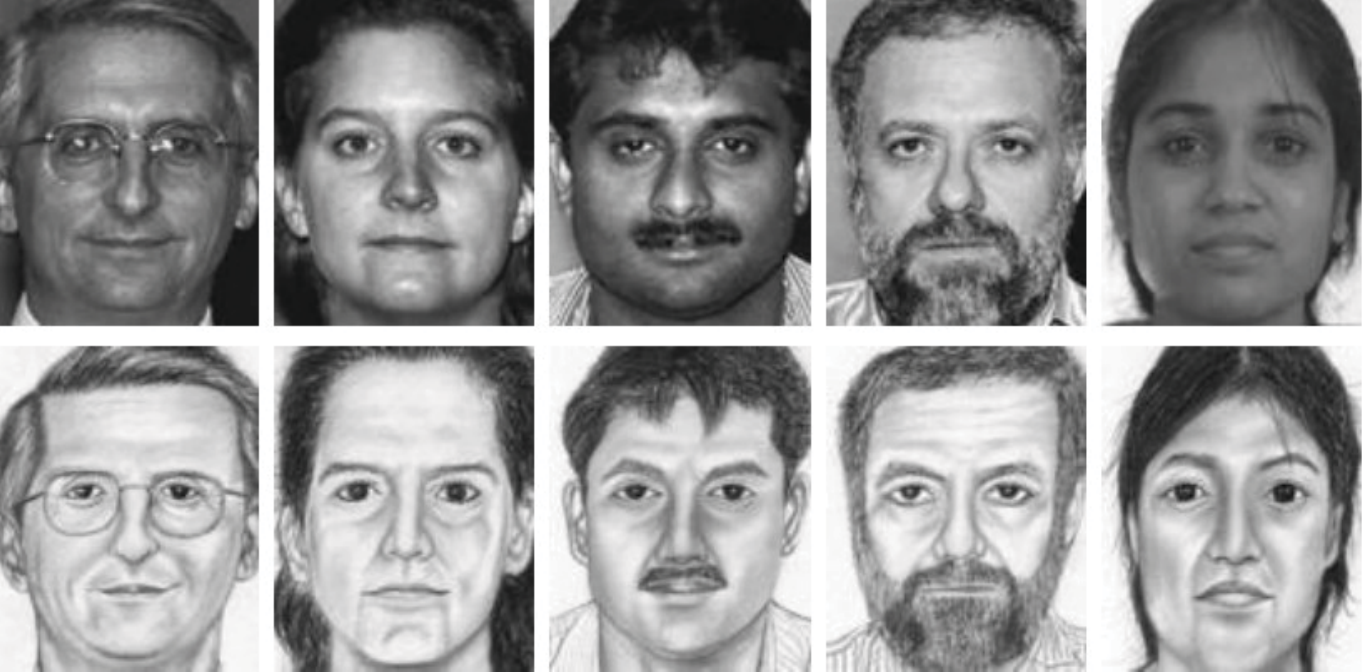}
\end{center}
   \caption{Example photos (first row) and sketches (second row) of the CUFSF database.}
\label{Figure7}
\end{figure}

\begin{table*}[t]
\small
\caption{Rank-1 recognition accuracies on the CUFSF database.}
\begin{center}
\begin{tabular}{cc|cc}
\hline
    Method &Accuracy &Method &Accuracy \\
\hline
Fisherface \cite{Ref36} &28.82\% & OpenBR \cite{Ref31} &10.80\% \\
TFSPS \cite{Ref6} &72.62\% & MrFSPS \cite{RefTNN1} &75.36\% \\
PLS \cite{Ref20} &51.00\% & MvDA \cite{Ref22} &55.50\% \\
LRBP \cite{Ref27} &91.12\% & P-RS \cite{Ref3} &83.95\% \\
G-HFR \cite{Ref4} &96.04\% & SGR-DA  &\textbf{96.97\%} \\
\hline
\end{tabular}
\end{center}
\label{tab:table1}
\end{table*}

\begin{table*}[t]
\small
\caption{Rank-10 recognition accuracies on the PRIP-VSGC database under protocol I.}
\begin{center}
\begin{tabular}{cc|cc}
\hline
    Method &Accuracy &Method &Accuracy \\
\hline
Fisherface \cite{Ref36} &21.87\% &OpenBR \cite{Ref31} &16.65\% \\
MCWLD \cite{Ref29} &15.40\% &\cite{Ref46} &27.60\% \\
SSD-based \cite{Ref32} &45.30\% &Deep Network \cite{Ref33} &52.00\% \\
P-RS \cite{Ref3} &53.73\% &SGR-DA &\textbf{70.00\%} \\
\hline
\end{tabular}
\end{center}
\label{tab:table2}
\end{table*}
The CUFSF database is used to evaluate the proposed method on matching viewed sketches with photos. There are totally 1194 persons in this database. Each person has one photo and corresponding one sketch drawn by the artist. There are illumination variations in photos and shape exaggerations in sketches of the database. The viewed sketches are drawn by professional forensic artist while viewing the photos. Some examples used in this paper are shown in Fig. \ref{Figure7}. In our experiment, 250 sketch-photo pairs are randomly selected to construct the representation dataset. Other 250 pairs are randomly selected for training and the rest 694 pairs are used for testing.

We compare the proposed SGR-DA method with three aforementioned baseline approaches, \textit{i.e.}, Fisherface, OpenBR, and P-RS, as well as several state-of-the-art methods. The rank-1 recognition accuracies of different methods are reported in Table \ref{tab:table1}. The two baseline face recognition methods (Fisherface and OpenBR) performed poorly on the HFR scenario. The image synthesis based method TFSPS first transformed face sketches and photos into the same modality and utilized random sampling LDA method \cite{Ref44} for recognition. The two common subspace projection-based methods (PLS and MvDA) only achieved rank-1 accuracies below 60\%. The two modality invariant feature descriptor-based methods (LRBP and P-RS) achieved good performance with 91.12\% and 83.94\% respectively. The graphical representation based method (G-HFR) achieved a rank-1 accuracy of 96.04\% on CUFSF. The proposed method represents heterogeneous faces with adaptive sparse graphical representations, in which the adaptive sparse property makes face images of different persons discriminative. The proposed spatial partition-based discriminant analysis framework further improves the discriminability and finally achieves a rank-1 accuracy of 96.97\%, which is superior to state-of-the-art methods.

\subsection{PRIP-VSGC Composite Sketch Database}
\label{subsection IV-C}

\begin{table}[t]
\small
\caption{Recognition accuracies on the PRIP-VSGC database under protocol II.}
\begin{center}
\begin{tabular}{ccc}
\hline
    Method &Rank-20 &Rank-40 \\
\hline
Fisherface \cite{Ref36} &12.50\% &15.36\% \\
OpenBR \cite{Ref31} &3.75\% &4.42\% \\
Deep Network \cite{Ref33} &15.60\% &48.30\% \\
P-RS \cite{Ref3} &42.27\% &55.47\% \\
SGR-DA &\textbf{54.93\%} &\textbf{67.60\%} \\
\hline
\end{tabular}
\end{center}
\label{tab:table3}
\end{table}

\begin{figure}[t]
\begin{center}
    \includegraphics[width=0.9\linewidth]{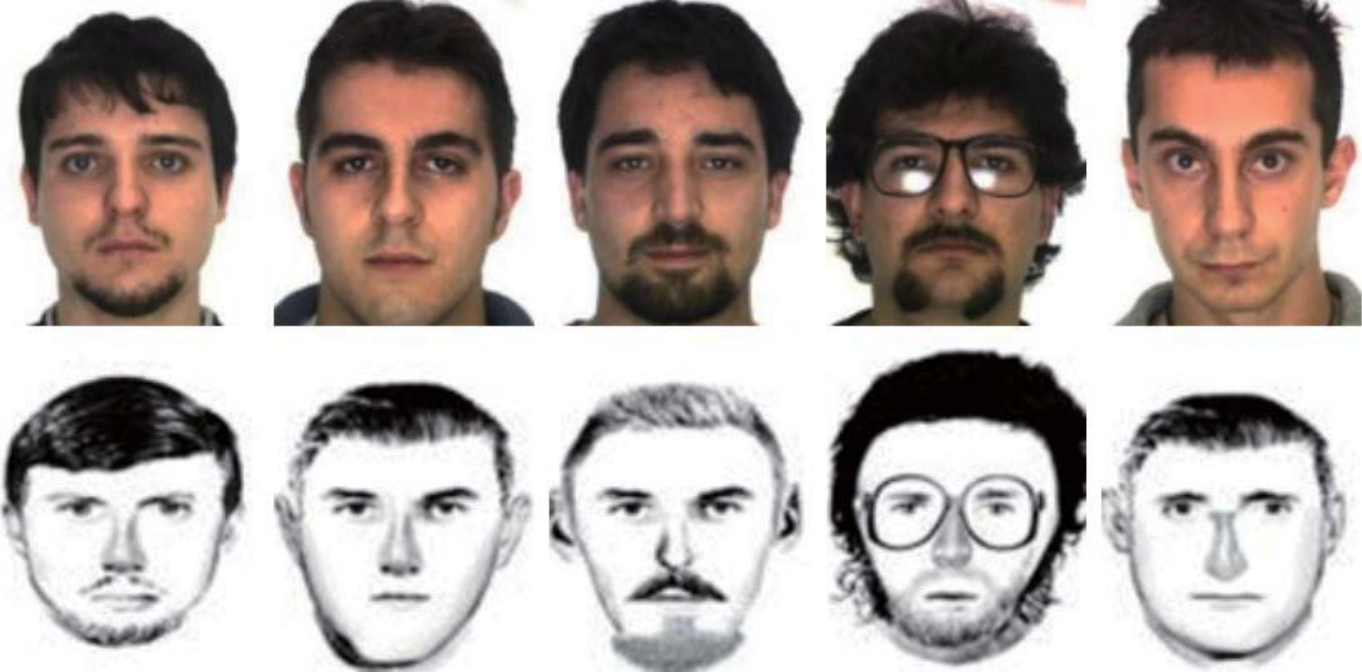}
\end{center}
   \caption{Example photos (first row) and composite sketches (second row) of the PRIP-VSGC database.}
\label{Figure8}
\end{figure}

The PRIP-VSGC database contains 123 photos from the AR database \cite{Ref45} and corresponding sketches created using composite generation software (FACES\footnote{http://www.iqbiometrix.com.} and Identi-Kit\footnote{http://www.identikit.net.}). The composite sketches are more easily available than hand drawn sketches because it is more affordable to create sketches by composite generation software than training a professional forensic artist. In our experiment, 123 sketch-photo pairs from the CUHK Student database \cite{Ref13} are used to form the representation dataset, and the 123 composite sketches generated by Identi-Kit are used for testing\footnote{Currently only the 123 composites generated using Identi-Kit are released on http://biometrics.cse.msu.edu/pubs/databases.html.}. Some examples of the composite sketch-photo pairs used are shown in Fig. \ref{Figure8}.

We first follow the baseline experiment protocol in \cite{Ref32,Ref33}, denoted as protocol I. 48 composite sketch-photo pairs are randomly selected for training. The rest 75 pairs form the testing set. The rank-10 accuracies of different methods under protocol I are reported in Table \ref{tab:table2}. The deep network-based transfer learning approach \cite{Ref33} and the state-of-the-art P-RS method achieved good performance of 52\% and 53.73\% respectively. It can be seen that the proposed method outperforms existing methods under protocol I and reached rank-10 accuracy of 70\%.

In order to evaluate the performance of the proposed method on large database, we then follow the extended experiment protocol in \cite{Ref32,Ref33}, denoted as protocol II. The size of the training set remains to be 48, and the gallery size is extended to be 2400 while the probe size is 75. It should be noticed that the methods \cite{Ref3,Ref32} used images obtained from law enforcement agencies to extend the gallery, and \cite{Ref33} selected images from multiple face databases which are not clearly introduced. Considering the images used to extend the gallery in existing methods are not available to the community, we randomly selected face images from a publicly available dataset, i.e. the labelled faces in the wild-a (LFW-a) \cite{Ref47}, to extend the gallery. It can help increase the diversity of the gallery set and mimic the real-world face recognition scenarios. We acknowledge that there may be bias of similarity scores between the LFW-a images and the gallery images \cite{RefTC3}. This bias is due to the fact that the face images in LFW-a are collected from Internet while the gallery photos are captured under controlled conditions. The usage of 10000 photos from LFW-a to extend the gallery set aims to make the face recognition problem more challenging. The rank-20 and rank-40 accuracies of different methods under protocol II are reported in Table \ref{tab:table3}. The proposed method achieves a rank-20 accuracy of 54.93\% and a rank-40 accuracy of 67.60\%, which outperform state-of-the-art methods of at least 12\%.

\begin{figure}[t]
\begin{center}
    \includegraphics[width=0.9\linewidth]{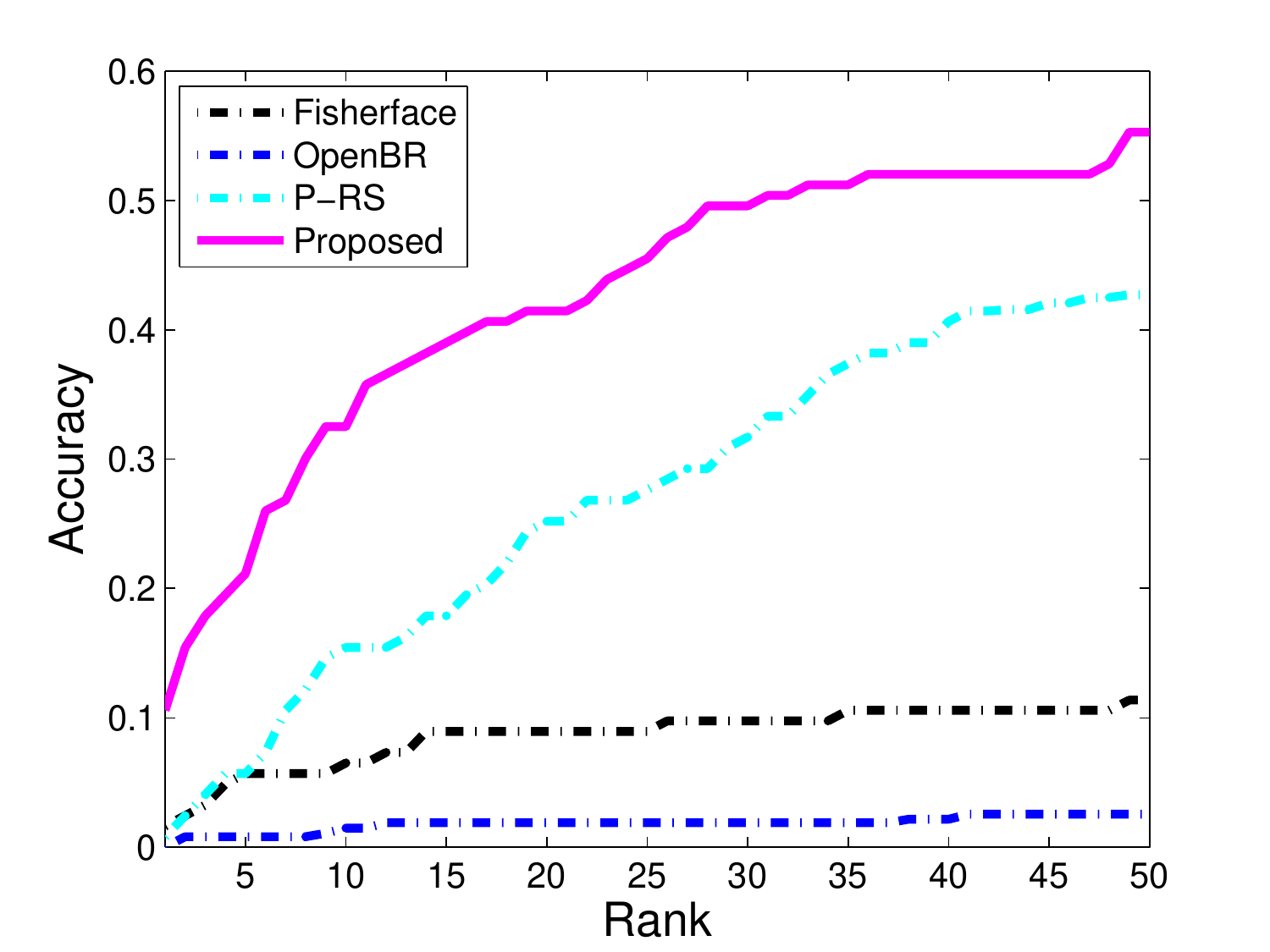}
\end{center}
   \caption{The cumulative match score results on the PRIP-VSGC database under protocol III.}
\label{Figure9}
\end{figure}

Finally, we further extend the gallery size to 10000 to mimic the real-world face retrieval scenarios in law enforcement agencies, denoted as protocol III. Face images from LFW-a are used to extend the gallery size. We randomly select 100 sketch-photo pairs from the CUFSF database for training and the whole 123 composite sketch-photo pairs are used for testing. The cumulative match scores comparison with baseline methods under protocol III is demonstrated in Fig. \ref{Figure9}. The Fisherface method achieves a rank-50 accuracy of 11.38\%. The open source face recognition algorithm OpenBR is developed for general face recognition and it performed poorly on the composite sketches (a rank-50 accuracy of 2.55\%). The kernel prototype similarities based P-RS approach achieves good performance on this dataset, with a rank-50 accuracy of 42.72\%. Benefiting from the maximum discriminability driven by both the adaptive sparse graphical representation and the spatial partition-based discriminant analysis, the proposed approach achieves a rank-50 accuracy of 55.28\%, which is superior to other methods.

\subsection{IIIT-D Semi-Forensic Sketch Database}
\label{subsection IV-D}

\begin{figure}[t]
\begin{center}
    \includegraphics[width=0.9\linewidth]{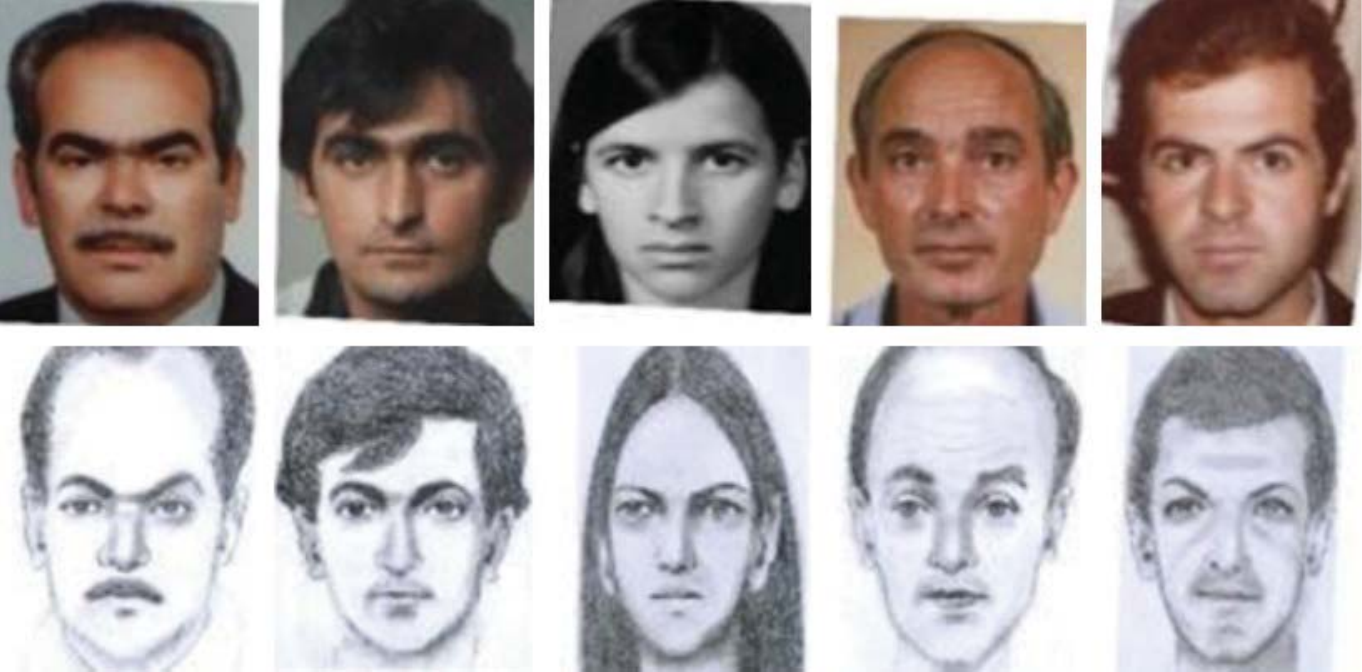}
\end{center}
   \caption{Example photos (first row) and semi-forensic sketches (second row) of the IIIT-D Semi-Forensic Sketch database.}
\label{Figure10}
\end{figure}

\begin{figure}[t]
\begin{center}
    \includegraphics[width=0.9\linewidth]{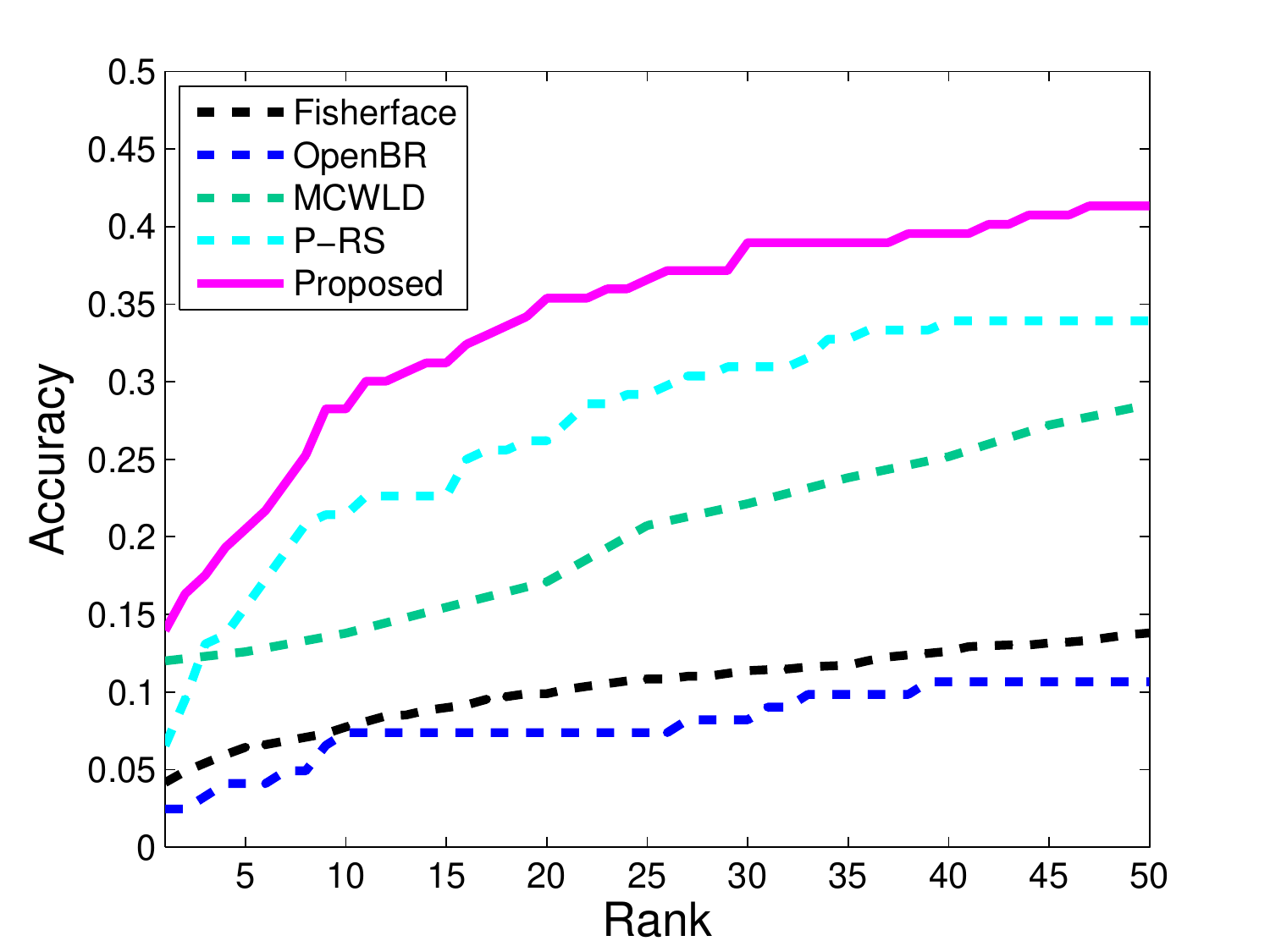}
\end{center}
   \caption{The cumulative match score results on the IIIT-D Semi-Forensic Sketch database.}
\label{Figure11}
\end{figure}
The IIIT-D Semi-Forensic Sketch database \cite{Ref38} is composed of 140 semi-forensic sketches and corresponding photos. The semi-forensic sketches are drawn based on the memory of the forensic artist after viewing the photos for a few minutes. Therefore, the semi-forensic sketches are less similar to photos than viewed sketches, which makes them practical to narrow the gap between viewed sketches and real-world forensic sketches. It is observed that classifiers trained on semi-forensic sketches can better fit forensic sketch recognition scenario \cite{Ref29}. In our experiment, 123 sketch-photo pairs in the CUHK AR database \cite{Ref13} are used to construct the representation dataset. The protocol in \cite{Ref29} is followed. The semi-forensic sketches are used for training classifiers and our collected forensic sketch database \cite{Ref4} containing 168 forensic sketch-photo pairs are used for testing. 10000 photos from LFW-a are used to extend the gallery. Some examples of the semi-forensic sketch-photo pairs are shown in Fig. \ref{Figure10}.

The cumulative match score results on the semi-forensic dataset is shown in Fig. \ref{Figure11}. The two baseline approaches, \textit{i.e.}, Fisherface and OpenBR, achieve similar performances on the semi-forensic sketches. The MCWLD method \cite{Ref29} utilized 6324 photos to extend the gallery and achieved a rank-50 accuracy of 28.52\%. The P-RS approach achieved a rank-50 accuracy of 33.93\% on this dataset. The proposed method utilized 10000 photos to extend the gallery and achieves a superior performance against the MCWLD method, with a larger size of gallery. A rank-50 accuracy of 41.33\% is achieved on this challenging dataset, which verifies the effectiveness of the proposed method.

\subsection{Forensic Sketch Database}
\label{subsection IV-E}

\begin{figure}[t]
\begin{center}
    \includegraphics[width=0.9\linewidth]{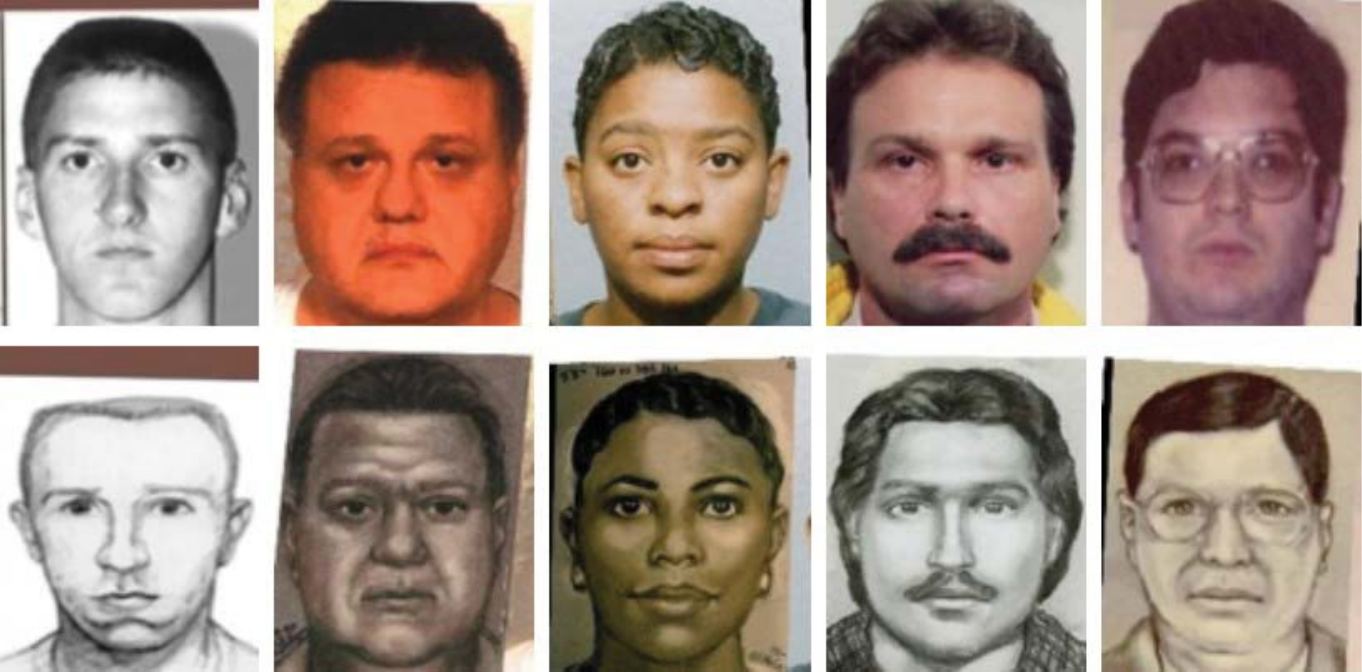}
\end{center}
   \caption{Example mug shot photos (first row) and forensic sketches (second row) of the forensic sketch database.}
\label{Figure12}
\end{figure}
The forensic sketch database \cite{Ref4} contains 168 real-world forensic sketches with corresponding mug shot photos. Examples of forensic sketch and mug shot photos are shown in Fig. \ref{Figure12}. Forensic sketches are drawn by the professional forensic artist based on the description of the eyewitnesses or victims. There are variant face perceptions between people and the eyewitnesses or victims cannot recall and describe all the details of the faces. These effects lead to large differences between the faces in forensic sketches and mug shot photos, which makes the forensic sketch recognition scenario more challenging than viewed sketch or semi-forensic sketch recognition. In our experiment, the CUHK AR database \cite{Ref13} including 123 sketch-photo pairs is taken as the representation dataset. The same train-test partition protocol in \cite{Ref3} is followed in this section. We randomly select 112 persons to form the training set and the rest 56 persons form the testing set. The gallery is extended by 10000 face images from LFW-a database.

\begin{figure}[t]
\begin{center}
    \includegraphics[width=0.9\linewidth]{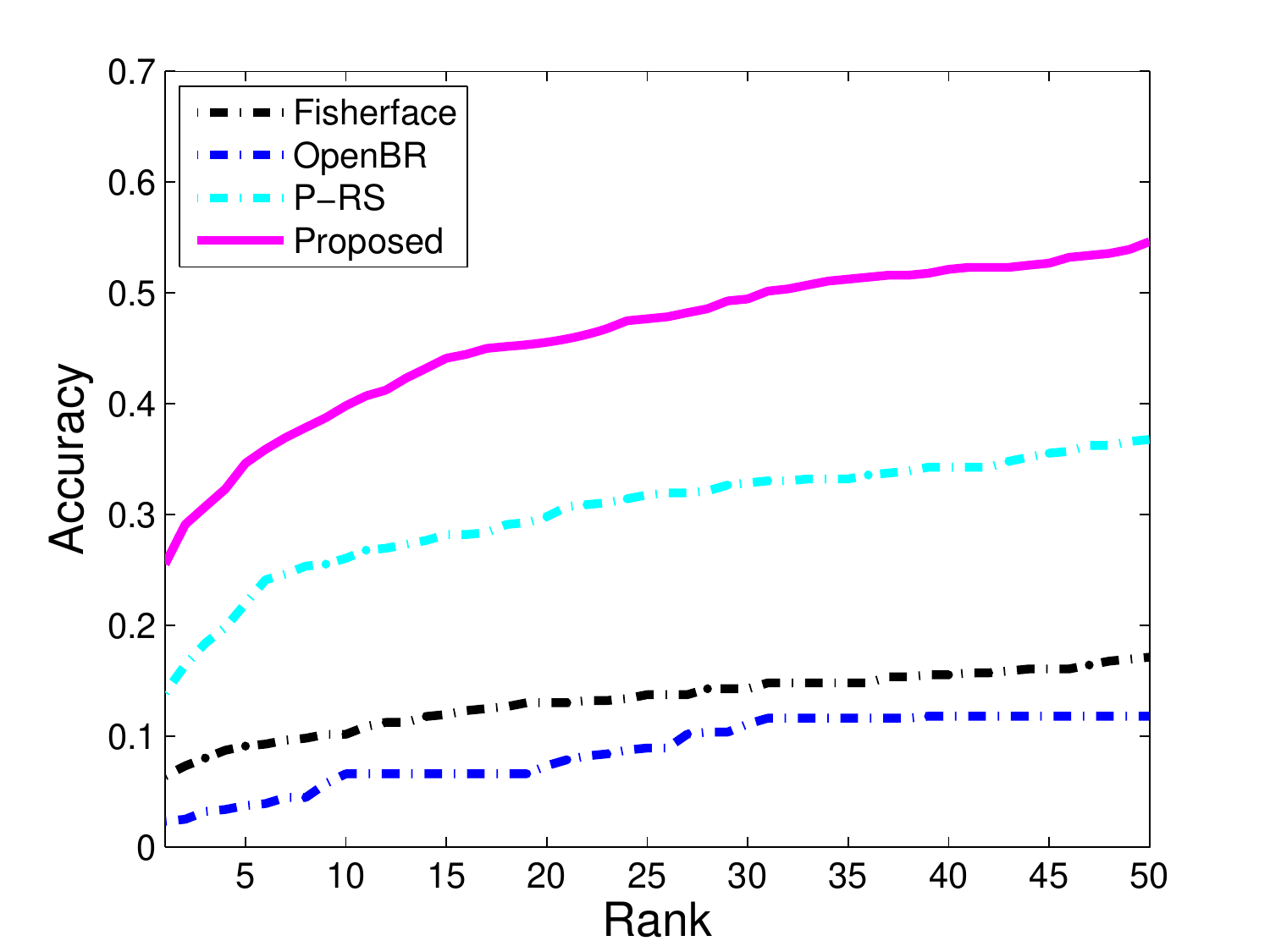}
\end{center}
   \caption{The cumulative match score results on the forensic sketch database.}
\label{Figure13}
\end{figure}

Fig. \ref{Figure13} shows the cumulative match scores of the proposed method and baseline methods on the forensic sketch dataset. The forensic sketches and mug shot photos in this dataset vary greatly from each other. Because they are collected from different sources, some of them are of poor quality. The two baseline methods (Fisherface and OpenBR) achieve limited performance, with accuracies increasing very slowly when the rank numbers become large. The P-RS method achieves a rank-50 accuracy of 36.79\% (higher than the reported 20.80\% in \cite{Ref3}). The proposed approach exploits the discriminative information in both forensic sketches and mug shot photos, and achieves superior performance with a rank-50 accuracy of 54.64\%.

\subsection{CASIA NIR-VIS 2.0 Face Database}
\label{subsection IV-F}

\begin{figure}[t]
\begin{center}
    \includegraphics[width=0.9\linewidth]{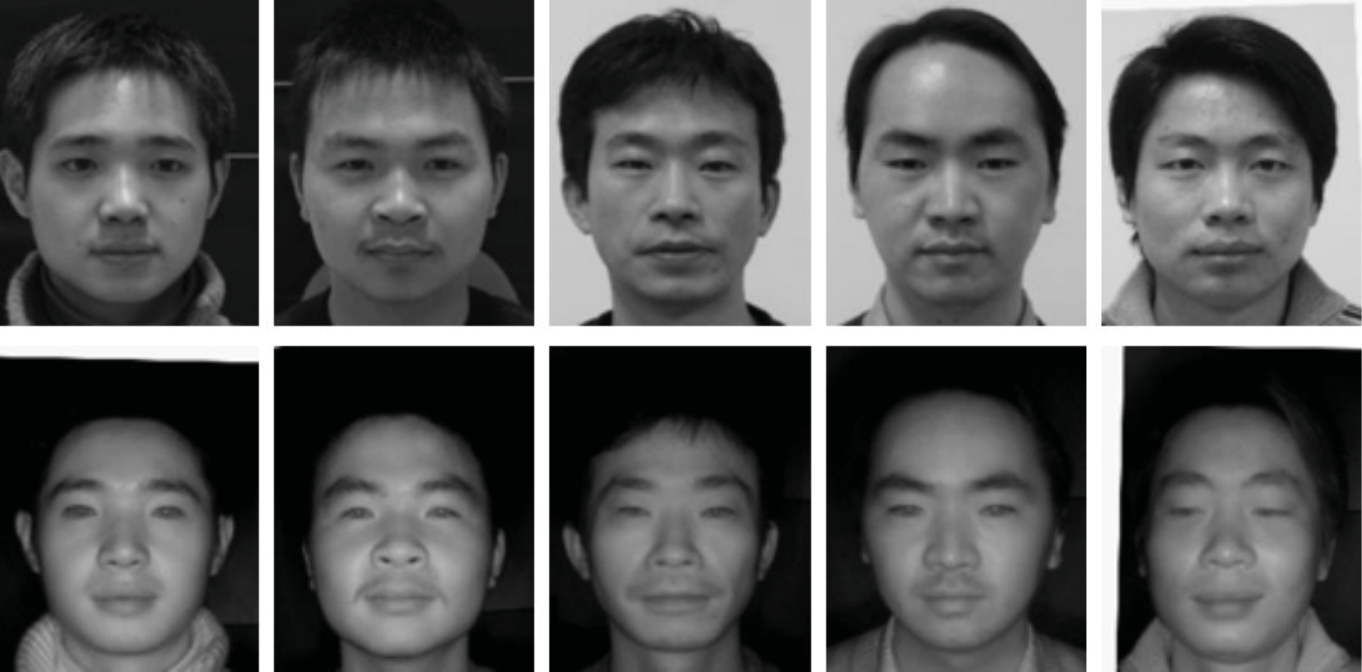}
\end{center}
   \caption{Example visible light photos (first row) and near infrared images (second row) of the CASIA NIR-VIS 2.0 face database.}
\label{Figure14}
\end{figure}

There are 725 subjects in the CASIA NIR-VIS 2.0 face database \cite{Ref39}. The subjects in this database are of different ages (from children to old people) and collected under different lighting conditions. Instead of using multiple images per subject, we randomly select one NIR-VIS pair per subject for training and testing. It is helpful to evaluate the performance of HFR methods and mimic real-world face recognition scenario with smaller training set and extended gallery \cite{Ref3}. Therefore, 10000 face images from the LFW-a dataset are used to extend the gallery in this section. Example NIR-VIS pairs are shown in Fig. \ref{Figure14}. We randomly choose 100 NIR-VIS pairs to form the representation dataset. In the rest 625 subjects, 417 subjects are used for training and the remaining 208 subjects form the testing set.

\begin{figure}[t]
\begin{center}
    \includegraphics[width=0.9\linewidth]{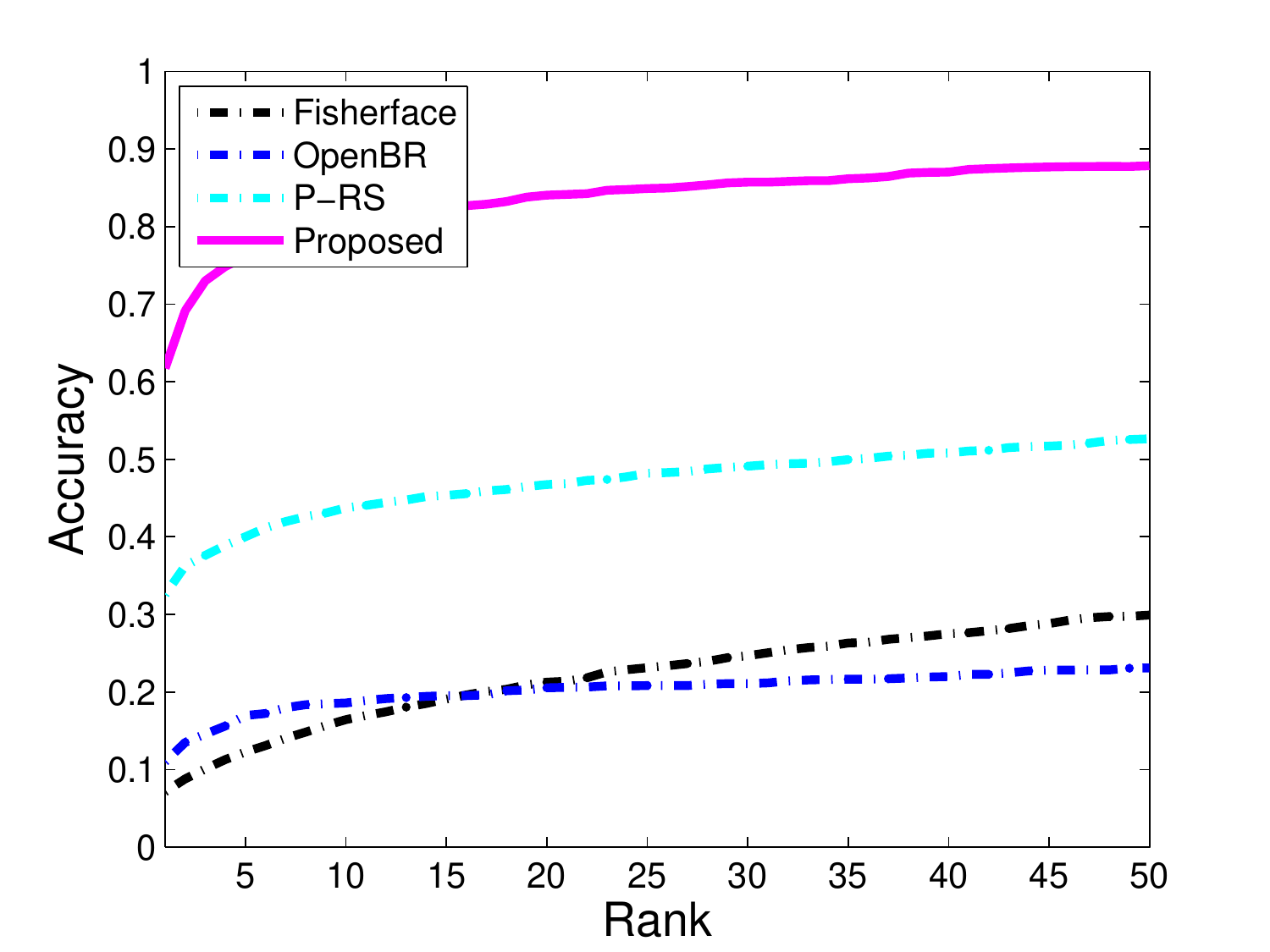}
\end{center}
   \caption{The cumulative match score results on the CASIA NIR-VIS 2.0 face database.}
\label{Figure15}
\end{figure}

We evaluate the proposed approach on the NIR-VIS dataset and the results are shown in Fig. \ref{Figure15}. The near infrared images are captured by NIR cameras to overcome illumination variation problem. Therefore there are great appearance differences between NIR and VIS images. The Fisherface and OpenBR methods cannot cope with the NIR scenario and only achieve rank-50 accuracies of 29.90\% and 23.09\% respectively. The P-RS \cite{Ref3} method is developed to deal with multiple HFR scenarios including NIR-VIS matching and leads to good matching accuracy with a rank-50 accuracy of 52.64\%. Our proposed method achieves a rank-50 accuracy of 87.84\% on this challenging database benefited from the most discriminative information by the adaptive sparse graphical representation and spatial partition based discriminant analysis framework.

\subsection{USTC-NVIE TIR-VIS Database}
\label{subsection IV-G}

\begin{figure}[t]
\begin{center}
    \includegraphics[width=0.9\linewidth]{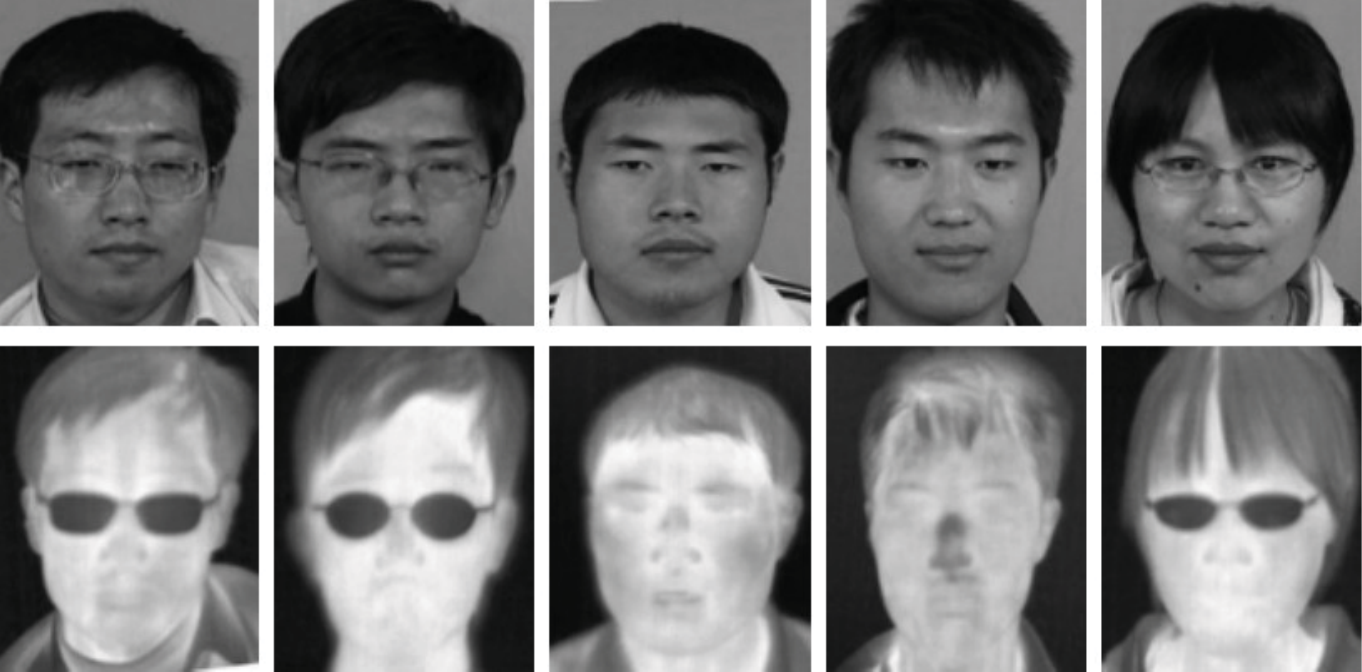}
\end{center}
   \caption{Example visible light photos (first row) and thermal infrared images (second row) of the USTC-NVIE TIR-VIS database.}
\label{Figure16}
\end{figure}

\begin{figure}[t]
\begin{center}
    \includegraphics[width=0.9\linewidth]{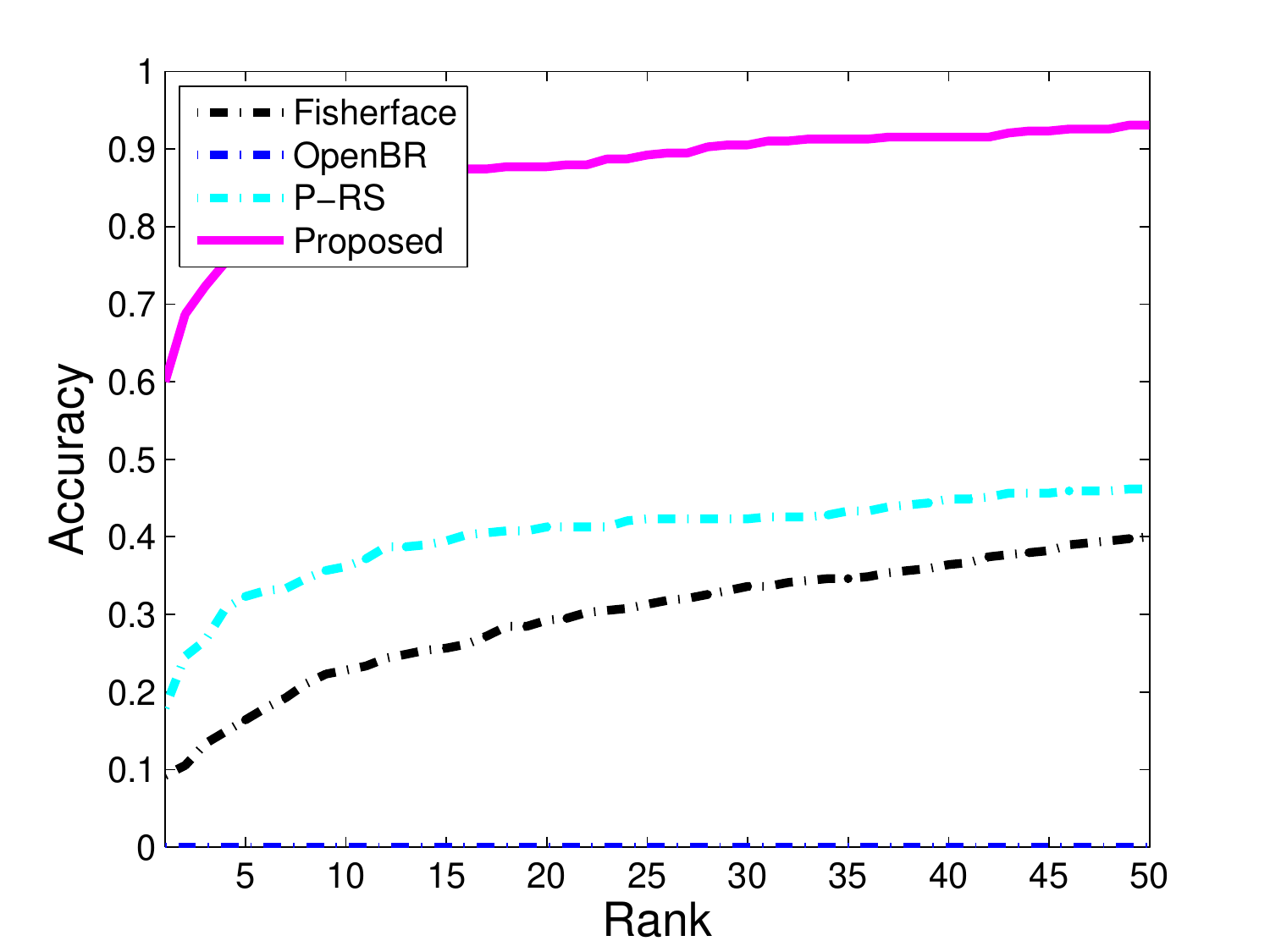}
\end{center}
   \caption{The cumulative match score results on the USTC-NVIE TIR-VIS database.}
\label{Figure17}
\end{figure}
The USTC-NVIE database \cite{Ref40} is composed of 215 subjects, with VIS and TIR images captured by the visible camera and infrared camera respectively. Due to the imaging principle of the thermal infrared camera, wearing glasses or not greatly affects the results of TIR imaging. When performing face recognition, it is important information whether wearing glasses or not. There are also lighting and expression variations in this database which increase the difficulty of matching. Fig. \ref{Figure16} shows some examples of TIR and VIS images. In our experiment, 129 subjects are used\footnote{Only 129 subjects are available in the USTC-NVIE database because of the loss of some TIR and VIS videos as reported in \cite{Ref40}.} with one TIR-VIS pair per subject. We further extend the gallery size by adding 10000 face images in LFW-a on this scenario. 60 TIR-VIS pairs are randomly selected as the representation dataset. We randomly select 30 TIR-VIS pairs to train the classifiers and the rest 39 pairs are used for testing.

Fig. \ref{Figure17} demonstrates the cumulative match score results on the TIR-VIS database. The thermal infrared cameras capture the thermal emission from the face, which makes the TIR images lack details of the faces. The open source face recognition algorithm OpenBR is invalid on this database. The Fisherface method achieves a rank-50 accuracy of 40\%. The P-RS method \cite{Ref3} can deal with TIR-VIS matching scenario to some extent, with a rank-50 accuracy of 46.15\%. The proposed method achieves excellent performance on this scenario, with a rank-50 accuracy of 93.08\%. This drastic improvement benefits from the maximum discriminability of our method driven by both the adaptive sparse graphical representation and the spatial partition based discriminant analysis.

\section{Conclusion}
\label{section V}

In this paper, a novel sparse graphical representation based discriminant analysis, denoted as SGR-DA, is proposed for multiple HFR scenarios. The most discriminative information is extracted through two aspects: the adaptive sparse graphical representation and spatial partition-based discriminant analysis. Firstly, the adaptive sparse property of our method maximizes the discriminability of different subjects. Secondly, the row-based, column-based, and learning-based spatial partition strategies are presented to refine the adaptive sparse vectors and further improve the discriminability. Extensive experiments were conducted on six commonly used heterogeneous face datasets. We achieved superior rank-1 accuracy (5\% higher) in comparison with state-of-the-art methods on the CUFSF viewed sketch dataset. We also outperformed existing methods on the composite sketch, semi-forensic sketch, and forensic sketch datasets under different protocols. It is further shown that the proposed approach has excellent generalization ability on both NIR-VIS dataset and TIR-VIS dataset. Our future work will focus on (1) further improve the accuracy of each HFR scenarios separately, and (2) investigate the relations between different HFR scenarios to help improve the performance of these scenarios together. We will also consider incorporating additional HFR scenarios, such as matching 2D photos with 3D range images and matching faces of different resolutions.

\section*{Appendix}
\begin{equation*}
\begin{aligned}
&\max_{\mathbf{w}}\ p(\mathbf{w}_{\mathbf{y}_1},\cdots,\mathbf{w}_{\mathbf{y}_N},\mathbf{y}_1,\cdots,\mathbf{y}_N)\\
\end{aligned}
\end{equation*}
\begin{equation*}
\begin{aligned}
=&\max_{\mathbf{w}}\ \prod_i\Phi(\mathbf{f}(\mathbf{y}_i),\mathbf{f}(\mathbf{w}_{\mathbf{y}_i}))\prod_{(i,j)\in\Xi}\Psi(\mathbf{w}_{\mathbf{y}_i},\mathbf{w}_{\mathbf{y}_j})\\
\end{aligned}
\end{equation*}
\begin{equation*}
\begin{aligned}
\propto&\max_{\mathbf{w}}\ \prod_i{\mbox{exp}\{-\|\mathbf{f}(\mathbf{y}_i)-\sum_{m=1}^M{w_{\mathbf{y}_{i,m}}\mathbf{f}(\mathbf{y}_{i,m})}\|^2/2\delta_\Phi^2\}}\\
\end{aligned}
\end{equation*}
\begin{equation*}
\begin{aligned}
&\prod_{(i,j)\in\Xi}{\mbox{exp}\{-\|{\sum_{m=1}^M{w_{\mathbf{y}_{i,m}}\mathbf{o}_{i,m}^j}}-{\sum_{m=1}^M{w_{\mathbf{y}_{j,m}}\mathbf{o}_{j,m}^i}}\|^2/2\delta_\Psi^2\}}\\
\end{aligned}
\end{equation*}
\begin{equation*}
\begin{aligned}
\propto&\max_{\mathbf{w}}\ \mbox{exp}\{-\sum_{i=1}^N\|\mathbf{f}(\mathbf{y}_i)-\sum_{m=1}^M{w_{\mathbf{y}_{i,m}}\mathbf{f}(\mathbf{y}_{i,m})}\|^2\\
\end{aligned}
\end{equation*}
\begin{equation*}
\begin{aligned}
-&\alpha\sum_{(i,j)\in\Xi}\|{\sum_{m=1}^M{w_{\mathbf{y}_{i,m}}\mathbf{o}_{i,m}^j}}-{\sum_{m=1}^M{w_{\mathbf{y}_{j,m}}\mathbf{o}_{j,m}^i}}\|^2\}\\
\end{aligned}
\end{equation*}
\begin{equation*}
\begin{aligned}
\Longleftrightarrow&\min_{\mathbf{w}}\ \sum_{i=1}^N\|\mathbf{f}(\mathbf{y}_i)-\sum_{m=1}^M{w_{\mathbf{y}_{i,m}}\mathbf{f}(\mathbf{y}_{i,m})}\|^2\\
\end{aligned}
\end{equation*}
\begin{equation*}
\begin{aligned}
+&\alpha\sum_{(i,j)\in\Xi}\|{\sum_{m=1}^M{w_{\mathbf{y}_{i,m}}\mathbf{o}_{i,m}^j}}-{\sum_{m=1}^M{w_{\mathbf{y}_{j,m}}\mathbf{o}_{j,m}^i}}\|^2\\
\end{aligned}
\end{equation*}
\begin{equation*}
\begin{aligned}
\Longleftrightarrow&\min_{\mathbf{w}}\ \mathbf{w}^T\mathbf{Q}\mathbf{w}+\mathbf{w}^T\mathbf{c}\\
\end{aligned}
\end{equation*}
\begin{equation*}
\begin{aligned}
s.t.\ &\sum_{m=1}^M{w_{\mathbf{y}_{i,m}}}=1,\ 0\leq{w_{\mathbf{y}_{i,m}}}\leq1,\\
&i=1,2,\cdots,N,\ m=1,2,\cdots,M
\end{aligned}
\end{equation*}
where $\mathbf{o}_{i,m}^j$ denotes the vector of intensity values extracted from overlapping area between $i$th patch and $j$th patch in the $m$th related neighbor. $\alpha$ is set to 0.25. The quadratic parameters $\mathbf{Q}$ and $\mathbf{c}$ are given below:
\begin{equation*}
\begin{aligned}
\mathbf{Q}=&\alpha\sum_{(i,j)\in\Xi}(\mathbf{O}_i^j-\mathbf{O}_j^i)^T(\mathbf{O}_i^j-\mathbf{O}_j^i)+\sum_{i=1}^N\mathbf{F}_i^T\mathbf{F}_i\\
\mathbf{c}=&-2\sum_{i=1}^N\mathbf{F}_i^T\mathbf{f}(\mathbf{y}_i)\\
\end{aligned}
\end{equation*}
where $\mathbf{F}_i$ and $\mathbf{O}_j^i$ are two matrices, with the $m$th column being $\mathbf{f}(\mathbf{y}_{i,m})$ and $\mathbf{o}_{i,m}^j$, respectively.

\bibliographystyle{IEEEtran}
\bibliography{DHFR}

\begin{thebibliography}{10}
\providecommand{\url}[1]{#1}
\csname url@samestyle\endcsname
\providecommand{\newblock}{\relax}
\providecommand{\bibinfo}[2]{#2}
\providecommand{\BIBentrySTDinterwordspacing}{\spaceskip=0pt\relax}
\providecommand{\BIBentryALTinterwordstretchfactor}{4}
\providecommand{\BIBentryALTinterwordspacing}{\spaceskip=\fontdimen2\font plus
\BIBentryALTinterwordstretchfactor\fontdimen3\font minus
  \fontdimen4\font\relax}
\providecommand{\BIBforeignlanguage}[2]{{%
\expandafter\ifx\csname l@#1\endcsname\relax
\typeout{** WARNING: IEEEtran.bst: No hyphenation pattern has been}%
\typeout{** loaded for the language `#1'. Using the pattern for}%
\typeout{** the default language instead.}%
\else
\language=\csname l@#1\endcsname
\fi
#2}}
\providecommand{\BIBdecl}{\relax}
\BIBdecl

\bibitem{Ref1}
S.~Klum, H.~Han, B.~Klare, and A.~K. Jain, ``The {F}ace{S}ketch{ID} system:
  Matching facial composites to mugshots,'' \emph{IEEE Transactions on
  Information Forensics and Security}, vol.~9, no.~12, pp. 2248--2263, 2014.

\bibitem{Ref2}
D.~Lin and X.~Tang, ``Inter-modality face recognition,'' in \emph{Proceedings
  of European Conference on Computer Vision}, 2006, pp. 13--26.

\bibitem{Ref3}
B.~Klare and A.~Jain, ``Heterogeneous face recognition using kernel prototype
  similarities,'' \emph{IEEE Transactions on Pattern Analysis and Machine
  Intelligence}, vol.~35, no.~6, pp. 1410--1422, 2013.

\bibitem{RefRR3}
N.~Wang, D.~Tao, X.~Gao, X.~Li, and J.~Li, ``A comprehensive survey to face
  hallucination,'' \emph{International Journal of Computer Vision}, vol.~31,
  no.~1, pp. 9--30, 2014.

\bibitem{Ref5}
H.~Zhou, Z.~Kuang, and K.~Wong, ``Markov weight fields for face sketch
  synthesis,'' in \emph{Proceedings of IEEE Conference on Computer Vision and
  Pattern Recognition}, 2012, pp. 1091--1097.

\bibitem{Ref6}
N.~Wang, D.~Tao, X.~Gao, X.~Li, and J.~Li, ``Transductive face sketch-photo
  synthesis,'' \emph{IEEE Transactions on Neural Networks and Learning System},
  vol.~24, no.~9, pp. 1364--1376, 2013.

\bibitem{Ref10}
X.~Tang and X.~Wang, ``Face sketch synthesis and recognition,'' in
  \emph{Proceedings of IEEE International Conference on Computer Vision}, 2003,
  pp. 687--694.

\bibitem{Ref11}
Q.~Liu, X.~Tang, H.~Jin, H.~Lu, and S.~Ma, ``A nonlinear approach for face
  sketch synthesis and recognition,'' in \emph{Proceedings of IEEE Conference
  on Computer Vision and Pattern Recognition}, 2005, pp. 1005--1010.

\bibitem{Ref12}
J.~Chen, D.~Yi, J.~Yang, G.~Zhao, S.~Li, and M.~Pietikainen, ``Learning
  mappings for face synthesis from near infrared to visual light images,'' in
  \emph{Proceedings of IEEE Conference on Computer Vision and Pattern
  Recognition}, 2009, pp. 156--163.

\bibitem{Ref13}
X.~Wang and X.~Tang, ``Face photo-sketch synthesis and recognition,''
  \emph{IEEE Transactions on Pattern Analysis and Machine Intelligence},
  vol.~31, no.~11, pp. 1955--1967, 2009.

\bibitem{Ref14}
J.~Li, P.~Hao, C.~Zhang, and M.~Dou, ``Hallucinating faces from thermal
  infrared images,'' in \emph{Proceeding of IEEE International Conference on
  Image Processing}, 2008, pp. 465--468.

\bibitem{Ref15}
Y.~Song, L.~Bao, Q.~Yang, and M.~Yang, ``Real-time exemplar-based face sketch
  synthesis,'' in \emph{Proceedings of European Conference on Computer Vision},
  2014, pp. 800--813.

\bibitem{Ref16}
L.~Zhang, L.~Lin, X.~Wu, S.~Ding, and L.~Zhang, ``End-to-end photo-sketch
  generation via fully convolutional representation learning,'' [Online].
  Available: http://arxiv.org/abs/1501.07180, 2015.

\bibitem{Ref17}
D.~Yi, R.~Liu, R.~Chu, Z.~Lei, and S.~Li, ``Face matching between near infrared
  and visible light images,'' in \emph{Proceedings of International Conference
  on Biometrics}, 2007, pp. 523--530.

\bibitem{Ref18}
Z.~Lei and S.~Li, ``Coupled spectral regression for matching heterogeneous
  faces,'' in \emph{Proceedings of IEEE Conference on Computer Vision and
  Pattern Recognition}, 2009, pp. 1123--1128.

\bibitem{Ref19}
Z.~Lei, C.~Zhou, D.~Yi, A.~Jain, and S.~Li, ``An improved coupled spectral
  regression for heterogeneous face recognition,'' in \emph{Proceedings of
  International Conference on Biometrics}, 2012, pp. 7--12.

\bibitem{Ref20}
A.~Sharma and D.~Jacobs, ``Bypass synthesis: {PLS} for face recognition with
  pose, low-resolution and sketch,'' in \emph{Proceedings of IEEE Conference on
  Computer Vision and Pattern Recognition}, 2011, pp. 593--600.

\bibitem{Ref21}
A.~Mignon and F.~Jurie, ``{CMML}: a new metric learning approach for cross
  modal matching,'' in \emph{Proceedings of Asian Conference on Computer
  Vision}, 2012, pp. 1--14.

\bibitem{Ref22}
M.~Kan, S.~Shan, H.~Zhang, S.~Lao, and X.~Chen, ``Multi-view discriminant
  analysis,'' in \emph{Proceedings of European Conference on Computer Vision},
  2012, pp. 808--821.

\bibitem{Ref23}
D.~Yi, Z.~Lei, and S.~Li, ``Shared representation learning for heterogeneous
  face recognition,'' in \emph{Proceedings of IEEE International Conference on
  Automatic Face and Gesture Recognition}, 2015.

\bibitem{Ref7}
B.~Klare, Z.~Li, and A.~Jain, ``Matching forensic sketches to mug shot
  photos,'' \emph{IEEE Transactions on Pattern Analysis and Machine
  Intelligence}, vol.~33, no.~3, pp. 639--646, 2011.

\bibitem{Ref8}
W.~Zhang, X.~Wang, and X.~Tang, ``Coupled information-theoretic encoding for
  face photo-sketch recognition,'' in \emph{Proceedings of IEEE Conference on
  Computer Vision and Pattern Recognition}, 2011, pp. 513--520.

\bibitem{Ref24}
S.~Liao, D.~Yi, Z.~Lei, R.~Qin, and S.~Li, ``Heterogeneous face recognition
  from local structures of normalized appearance,'' in \emph{Proceedings of
  IAPR International Conference on Biometrics}, 2009.

\bibitem{Ref27}
H.~Galoogahi and T.~Sim, ``Face sketch recognition by local radon binary
  pattern,'' in \emph{Proceedings of IEEE International Conference on Image
  Processing}, 2012, pp. 1837--1840.

\bibitem{Ref28}
A.~Alex, V.~Asari, and A.~Mathew, ``Local difference of gaussian binary
  pattern: robust features for face sketch recognition,'' in \emph{Proceedings
  of IEEE International Conference on Systems, Man, and Cybernetics}, 2013, pp.
  1211--1216.

\bibitem{Ref29}
H.~Bhatt, S.~Bharadwaj, R.~Singh, and M.~Vatsa, ``Memetically optimized {MCWLD}
  for matching sketches with digital face images,'' \emph{IEEE Transactions on
  Information Forensics and Security}, vol.~7, no.~5, pp. 1522--1535, 2012.

\bibitem{Ref30}
H.~Han, B.~Klare, K.~Bonnen, and A.~Jain, ``Matching composite sketches to face
  photos: a component-based approach,'' \emph{IEEE Transactions on Information
  Forensics and Security}, vol.~8, no.~1, pp. 191--204, 2013.

\bibitem{Ref32}
P.~Mittal, A.~Jain, G.~Goswami, R.~Singh, and M.~Vatsa, ``Recognizing composite
  sketches with digital face images via {SSD} dictionary,'' in
  \emph{Proceedings of IEEE International Conference on Biometrics}, 2014, pp.
  1--6.

\bibitem{Ref33}
P.~Mittal, M.~Vatsa, and R.~Singh, ``Composite sketch recognition via deep
  network-{A} transfer learning approach,'' in \emph{Proceedings of IAPR
  International Conference on Biometrics}, 2015.

\bibitem{Ref4}
C.~Peng, X.~Gao, N.~Wang, and J.~Li, ``Graphical representation for
  heterogeneous face recognition,'' \emph{IEEE Transactions on Pattern Analysis
  and Machine Intelligence}, 2016.

\bibitem{RefR1}
Z.~Cao, Q.~Yin, X.~Tang, and J.~Sun, ``Face recognition with learning-based
  descriptor,'' in \emph{Proceedings of IEEE Conference on Computer Vision and
  Pattern Recognition}, 2010, pp. 2707--2714.

\bibitem{RefR2}
Z.~Lei, M.~Pietik\"{a}inen, and S.~Li, ``Learning discriminant face
  descriptor,'' \emph{IEEE Transactions on Pattern Analysis and Machine
  Intelligence}, vol.~36, no.~2, pp. 289--302, 2014.

\bibitem{Ref9}
S.~Liu, D.~Yi, Z.~Lei, and S.~Li, ``Heterogeneous face image matching using
  multi-scale features,'' in \emph{Proceedings of IAPR International Conference
  on Biometrics}, 2012, pp. 79--84.

\bibitem{RefTC2}
Z.~Chai, H.~Mendez-Vazquez, R.~He, Z.~Sun, and T.~Tan, ``Semantic pixel sets
  based local binary patterns for face recognition,'' in \emph{Proceedings of
  Asian Conference on Computer Vision}, 2012, pp. 639--651.

\bibitem{RefTNN1}
C.~Peng, X.~Gao, N.~Wang, D.~Tao, X.~Li, and J.~Li, ``Multiple
  representations-based face sketch-photo synthesis,'' \emph{IEEE Transactions
  on Neural Networks and Learning Systems}, 2016.

\bibitem{Ref25}
D.~Lowe, ``Distinctive image features from scale-invariant key-points,''
  \emph{International Journal of Computer Vision}, vol.~60, no.~2, pp. 91--110,
  2004.

\bibitem{Ref26}
T.~Ojala, M.~Pietik\"{a}inen, and T.~M\"{a}enp\"{a}\"{a}, ``Multiresolution
  gray-scale and rotation invariant texture classification with local binary
  patterns,'' \emph{IEEE Transactions on Pattern Analysis and Machine
  Intelligence}, vol.~24, no.~7, pp. 971--987, 2002.

\bibitem{Ref34}
D.~Lee and H.~Seung, ``Learning the parts of objects with nonnegative matrix
  factorization,'' \emph{Nature}, vol. 401, pp. 788--791, 1999.

\bibitem{RefTC1}
B.~Cheng, J.~Yang, S.~Yan, Y.~Fu, and T.~Huang, ``Learning with $\ell^1$-graph
  for image analysis,'' \emph{IEEE Transactions on Image Processing}, vol.~19,
  no.~4, pp. 858--866, 2010.

\bibitem{Ref35}
I.~Jolliffe, \emph{Principal component analysis}.\hskip 1em plus 0.5em minus
  0.4em\relax New York: Springer, 2002.

\bibitem{Ref36}
P.~Belhumeur, J.~Hespanda, and D.~Kiregeman, ``Eigenfaces vs. fisherfaces:
  recognition using class specific linear projection,'' \emph{IEEE Transactions
  on Pattern Analysis and Machine Intelligence}, vol.~19, no.~7, pp. 711--720,
  1997.

\bibitem{Ref37}
J.~MacQueen, ``Some methods for classification and analysis of multivariate
  observations,'' in \emph{Proceedings of 5t Berkeley Symposium on Mathematical
  Statistics and Probability}, 1967, pp. 281--297.

\bibitem{Ref38}
H.~S. Bhatt, S.~Bharadwaj, R.~Singh, and M.~Vatsa, ``Memetic approach for
  matching sketches with digital face images,'' IIITD-TR-2011-006, Tech. Rep.,
  2011.

\bibitem{Ref39}
S.~Li, D.~Yi, Z.~Lei, and S.~Liao, ``The {CASIA} {NIR}-{VIS} 2.0 face
  database,'' in \emph{Proceedings of IEEE Conference on Computer Vision and
  Pattern Recognition Workshops}, 2013, pp. 348--353.

\bibitem{Ref40}
S.~Wang, Z.~Liu, S.~Lv, Y.~Lv, G.~Wu, P.~Peng, F.~Chen, and X.~Wang, ``A
  natural visible and infrared facial expression database for expression
  recognition and emotion inference,'' \emph{IEEE Transactions on Multimedia},
  vol.~12, no.~7, pp. 682--691, 2010.

\bibitem{Ref31}
J.~Klontz, B.~Klare, S.~Klum, A.~Jain, and M.~Burge, ``Open source biometric
  recognition,'' \emph{IEEE Biometrics: Theory, Applications, and Systems
  (under review)}, 2013.

\bibitem{Ref42}
H.~Bay, A.~Ess, T.~Tuytelaars, and L.~Gool, ``{SURF}: speeded up robust
  features,'' \emph{Computer Vision and Image Understanding}, vol. 110, no.~3,
  pp. 346--359, 2008.

\bibitem{Ref43}
N.~Dalal and B.~Triggs, ``Histograms of oriented gradients for human
  detection,'' in \emph{Proceedings of IEEE Conference on Computer Vision and
  Pattern Recognition}, 2005, pp. 886--893.

\bibitem{Ref41}
Y.~Sun, X.~Wang, and X.~Tang, ``Deep convolutional network cascade for facial
  point detection,'' in \emph{Proceedings of IEEE Conference on Computer Vision
  and Pattern Recognition}, 2013, pp. 3476--3483.

\bibitem{Ref46}
P.~Mittal, A.~Jain, R.~Singh, and M.~Vatsa, ``Boosting local descriptors for
  matching composite and digital face images,'' in \emph{Proceedings of IEEE
  Conference on Image Processing}, 2013, pp. 2797--2801.

\bibitem{Ref44}
X.~Wang and X.~Tang, ``Random sampling for subspace face recognition,''
  \emph{International Journal of Computer Vision}, vol.~70, no.~1, pp. 91--104,
  2006.

\bibitem{Ref45}
A.~Martinez and R.~Benavente, ``The {AR} face database,'' CVC Technical Report
  \#24, Tech. Rep., 1998.

\bibitem{Ref47}
L.~Wolf, T.~Hassner, and Y.~Taigman, ``Effective unconstrained face recognition
  by combining multiple descriptors and learned background statistics,''
  \emph{IEEE Transactions on Pattern Analysis and Machine Intelligence},
  vol.~33, no.~10, pp. 1978--1990, 2011.

\bibitem{RefTC3}
L.~Best-Bowden, H.~Han, C.~Otto, B.~Klare, and A.~Jain, ``Unconstrained face
  recognition: Identifying a person of interest from a media collection,''
  \emph{IEEE Transactions on Information Forensics and Security}, vol.~9,
  no.~2, pp. 2144--2157, 2014.

\end{thebibliography}
\end{document}